\documentclass[10pt,twocolumn,letterpaper]{article}

\usepackage[pagenumbers]{iccv} 

\usepackage{times}
\usepackage{float}
\usepackage{epsfig}
\usepackage{graphicx}
\usepackage{amsmath}
\usepackage{bm}
\usepackage{amssymb}
\usepackage{booktabs}
\usepackage{subcaption}
\usepackage{makecell}
\usepackage{rotating}
\usepackage{multirow}
\usepackage{xcolor}
\usepackage{cuted}
\usepackage{colortbl}
\usepackage{tcolorbox}
\usepackage{array}
\usepackage{enumitem}
\usepackage{multicol}
\usepackage[skip=2pt]{caption}
\usepackage[accsupp]{axessibility}

\newcommand{\mAP}{\mathrm{mAP}}

\definecolor{dilcolor}{HTML}{d6466d}
\definecolor{ecdilcolor}{HTML}{5900b1}
\definecolor{cilcolor}{HTML}{00b159}
\definecolor{classcolor}{HTML}{00aedb}
\definecolor{objdetcolor}{HTML}{f37735}
\definecolor{ecrico}{HTML}{c5e0b4}
\definecolor{drico}{HTML}{bdd7ee}

\newcommand{\dilbox}[1]{\tcbox[colback=dilcolor!30, colframe=white, boxrule=0pt, rounded corners=all, on line, boxsep=0.5pt, left=0.5pt, right=0.5pt, top=0.5pt, bottom=0.5pt]{#1}}
\newcommand{\ecdilbox}[1]{\tcbox[colback=ecdilcolor!30, colframe=white, boxrule=0pt, rounded corners=all, on line, boxsep=0.5pt, left=0.5pt, right=0.5pt, top=0.5pt, bottom=0.5pt]{#1}}
\newcommand{\cilbox}[1]{\tcbox[colback=cilcolor!30, colframe=white, boxrule=0pt, rounded corners=all, on line, boxsep=0.5pt, left=0.5pt, right=0.5pt, top=0.5pt, bottom=0.5pt]{#1}}
\newcommand{\classbox}[1]{\tcbox[colback=classcolor!30, colframe=white, boxrule=0pt, rounded corners=all, on line, boxsep=0.5pt, left=0.5pt, right=0.5pt, top=0.5pt, bottom=0.5pt]{#1}}
\newcommand{\objdetbox}[1]{\tcbox[colback=objdetcolor!30, colframe=white, boxrule=0pt, rounded corners=all, on line, boxsep=0.5pt, left=0.5pt, right=0.5pt, top=0.1pt, bottom=0.1pt]{#1}}
\newcommand{\ecricobox}[1]{%
  \tcbox[colback=ecrico!70, colframe=white, boxrule=0pt, rounded corners=all, on line, 
         boxsep=0.5pt, left=1pt, right=1pt, top=0.2pt, bottom=0.2pt]{\strut #1}%
}
\newcommand{\dricobox}[1]{%
  \tcbox[colback=drico!70, colframe=white, boxrule=0pt, rounded corners=all, on line, 
         boxsep=0.5pt, left=1pt, right=1pt, top=0.1pt, bottom=0.1pt]{\strut #1}%
}

\newcommand{\veryshortarrow}[1][3pt]{%
  \mspace{3mu}%
  \hbox{\rule[\dimexpr\fontdimen22\textfont2-.2pt\relax]{#1}{.4pt}}%
  \mkern-4mu%
  \hbox{\usefont{U}{lasy}{m}{n}\symbol{41}}%
  \mspace{3mu}%
}

\definecolor{iccvblue}{rgb}{0.21,0.49,0.74}
\usepackage[pagebackref,breaklinks,colorlinks,allcolors=iccvblue]{hyperref}

\def\paperID{6}
\def\confName{ICCV}
\def\confYear{2025}

\title{RICO: Two Realistic Benchmarks and an In-Depth Analysis\\ for Incremental Learning in Object Detection}

\author{%
  Matthias~Neuwirth\textendash Trapp$^{1, 2}$\quad
  Maarten~Bieshaar$^{2}$\quad
  Danda~Pani~Paudel$^{3}$\quad
  Luc~Van~Gool$^{3}$\\
  {\small\texttt{mneuwirth@ethz.ch}}\\[0.6em]
  {\small$^{1}$ETH Zürich\quad}
  {\small$^{2}$Bosch Research\quad}
  {\small$^{3}$INSAIT, Sofia University “St.~Kliment Ohridski”}
}

\begin{document}
\maketitle

\begin{strip}
\vspace{-1.5cm}
    \centering
    \includegraphics[width=\textwidth]{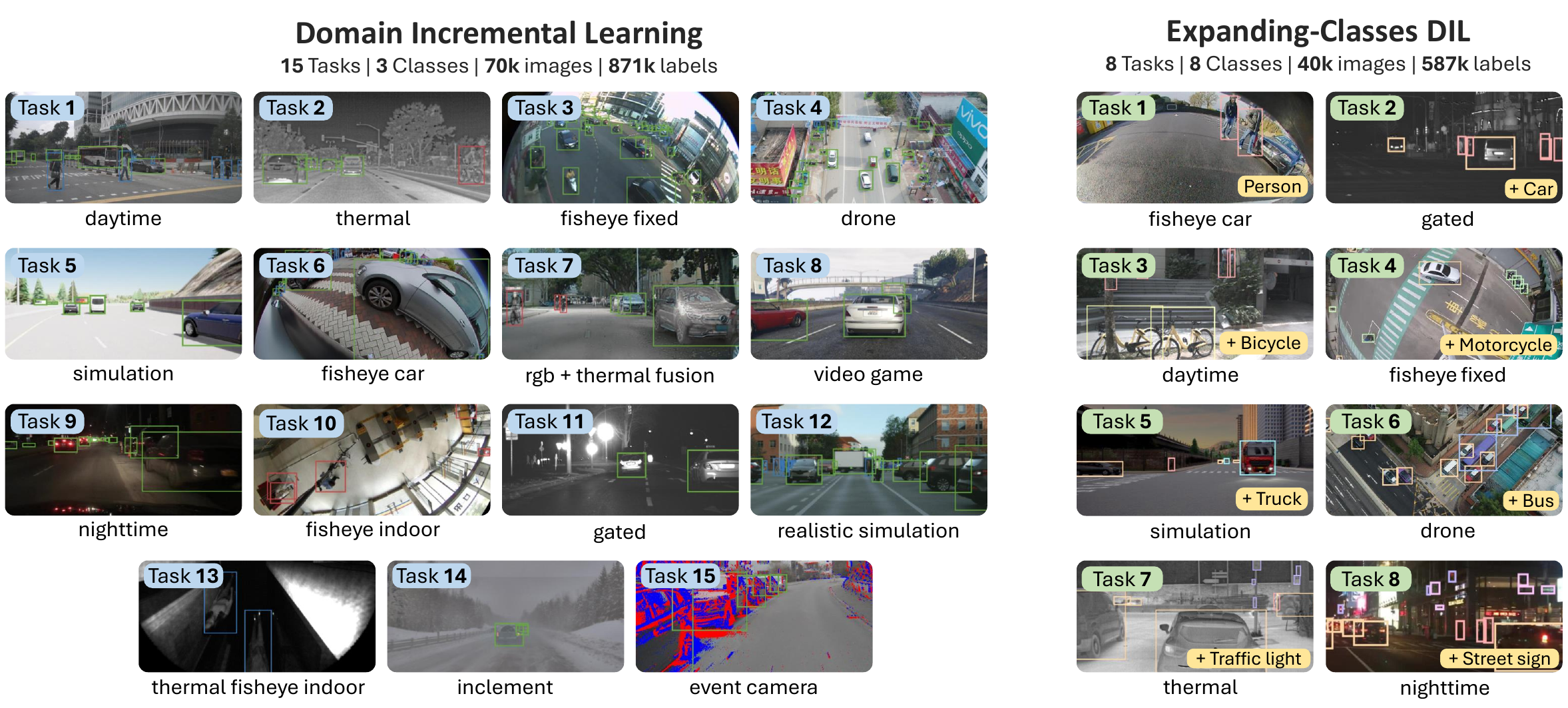}
    \captionof{figure}{
        We introduce \textbf{two novel benchmarks} built from 14 datasets: standard and expanding-classes domain incremental learning (DIL). The DIL benchmark has 15 tasks with diverse camera types, synthetic/real data, varying daytime, weather, and labeling policies. The expanding-classes DIL benchmark has 8 tasks, each adding a new class \& domain while maintaining task diversity.
    }
    \label{fig:fig1_tasks}
\end{strip}

\begin{abstract}
Incremental Learning (IL) trains models sequentially on new data without full retraining, offering privacy, efficiency, and scalability. IL must balance adaptability to new data with retention of old knowledge. However, evaluations often rely on synthetic, simplified benchmarks, obscuring real-world IL performance. To address this, we introduce two \textbf{Realistic Incremental Object Detection Benchmarks} (RICO): \textbf{Domain RICO} (D-RICO) features domain shifts with a fixed class set, and \textbf{Expanding-Classes RICO} (EC-RICO) integrates new domains and classes per IL step. Built from 14 diverse datasets covering real and synthetic domains, varying conditions (\eg, weather, time of day), camera sensors, perspectives, and labeling policies, both benchmarks capture challenges absent in existing evaluations. Our experiments show that all IL methods underperform in adaptability and retention, while replaying a small amount of previous data already outperforms all methods. However, individual training on the data remains superior. We heuristically attribute this gap to weak teachers in distillation, single models’ inability to manage diverse tasks, and insufficient plasticity. Our code is publicly available at \href{https://github.com/boschresearch/ricobenchmark}{https://github.com/boschresearch/ricobenchmark}.
\end{abstract}
    
\section{Introduction}

\begin{figure*}[t]
    \includegraphics[width=1\linewidth]{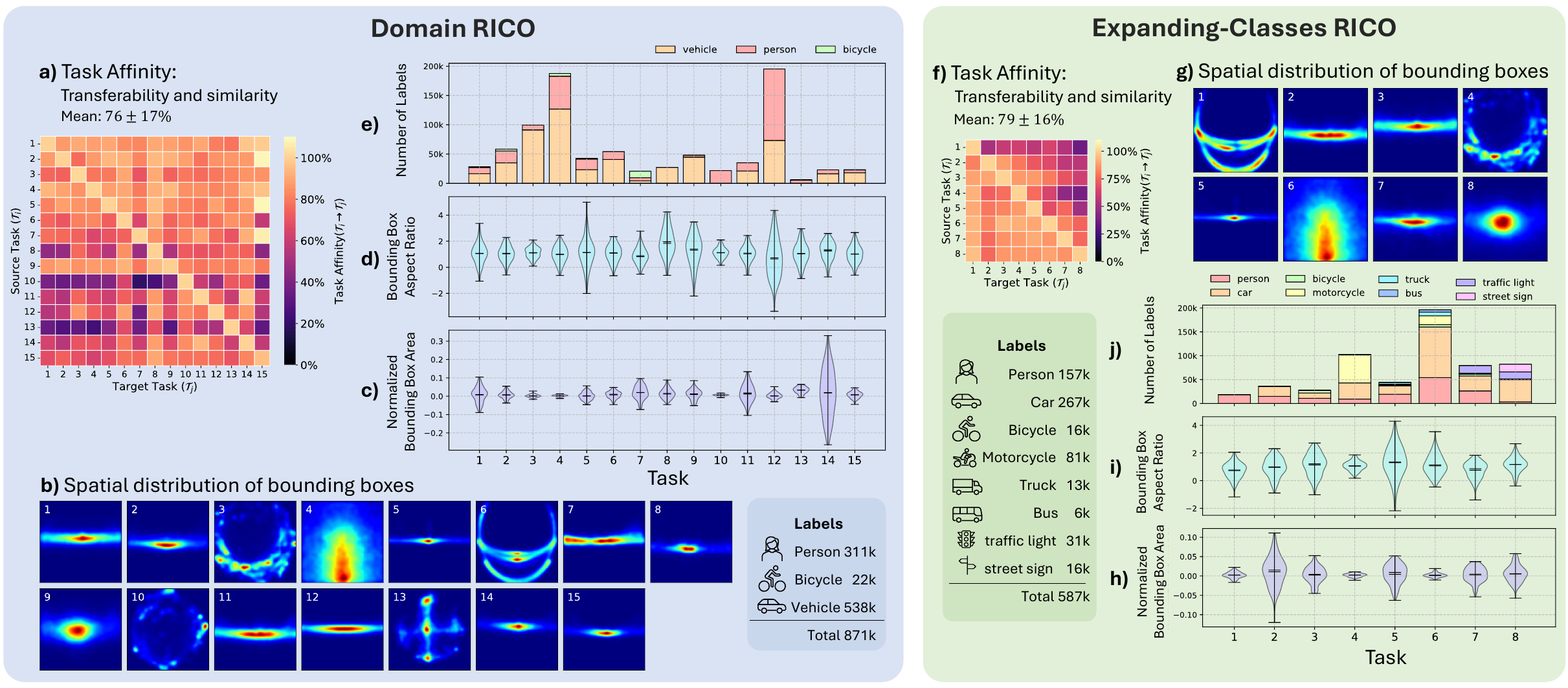}
    \caption{Overview of the domain-RICO and expanding-classes RICO benchmark tasks. See Sections~\ref{sec:benchmark:d-rico} and~\ref{sec:benchmark:ec-rico} for details.}
    \label{fig:fig2}
\end{figure*}

Incremental learning enables artificial neural networks (ANN) to acquire knowledge sequentially, offering benefits like improved data privacy, computational efficiency, and scalability~\cite{zhou_continual_2024, zheng_towards_2024, wang_comprehensive_2024}. 
It contributes to progress towards \textit{artificial general intelligence}~\cite{wang_comprehensive_2024}.
However, ANN face the fundamental challenge of balancing \textit{stability} (maintaining previously acquired knowledge to prevent \textit{catastrophic forgetting}) and \textit{plasticity} (the capacity to incorporate new information efficiently)~\cite{sha_forgetting_2024, aljundi_memory_2018, wang_comprehensive_2024}.

This work addresses domain-incremental learning (DIL) for object detection, where new domains (tasks) are learned sequentially. This setting is crucial for applications like autonomous driving~\cite{verwimp_clad_2023, shieh_continual_2020, su_object_2024, witte_severity_2023}, robotics~\cite{hajizada_continual_2024, lesort_continual_2020-1}, and surveillance~\cite{zhang_prompting_2024, khan_continual_2024}, where many different kinds of distribution shifts can occur. Existing benchmarks lack realistic diversity, using either a single dataset with uniform characteristics~\cite{pasti_tiny_2024-1, menezes_continual_2022, wang_wanderlust_2021, verwimp_clad_2023} or multiple datasets with extreme variations~\cite{song_non-exemplar_2024}. Whereas DIL benchmarks exist for classification, their insights do not transfer to object detection due to the additional spatial localization challenges in object detection and structural model differences~\cite{wang_comprehensive_2024, menezes_continual_2022}. Therefore, incremental object detection complements classification to achieve general methods~\cite{mitchell_continual_2025}.

Instead of relying on a single dataset for benchmarking, we propose constructing an IL benchmark by integrating multiple datasets.
This way, compared to other benchmarks~\cite{pasti_tiny_2024-1, wang_wanderlust_2021, verwimp_clad_2023, song_non-exemplar_2024}  (see Table~\ref{tab:comp-different-benchmarks}), we achieve:
\begin{itemize} \setlength{\itemindent}{0.5cm}
    \item greater diversity in image characteristics,
    \item increased variation in class distribution,
    \item more variation in bounding box statistics, and
    \item multiple different labeling policies and qualities.
\end{itemize}
We select, curate, and process the datasets, resulting in a natural task diversity as in real-world IL applications.

We introduce two novel DIL benchmarks: \textbf{Domain Realistic Incremental Object Detection (D-RICO)}, with 15 tasks from 14 distinct datasets maintaining fixed classes across domains, and \textbf{Expanding-Classes RICO (EC-RICO)}, comprising 8 tasks from 8 datasets, each introducing a new class and domain.

Figure~\ref{fig:fig1_tasks} illustrates task diversity in image space. The heuristic task affinity (Figure~\ref{fig:fig2} a) and f)) reveals order-dependent transferability and similarity (see Section~\ref{subsec:bench:task-affinity}). Figures~\ref{fig:fig2} b)–e), g)–j) show variations in class distributions and bounding box statistics. Table~\ref{tab:comp-different-benchmarks} compares D-RICO and EC-RICO to other benchmarks, highlighting our superior dataset scope, task variety, and domain coverage for IL in object detection. Further quantitative experiments highlight this even more (see Suppl. Mat. Section D1).

The two benchmarks enable studying IL methods for object detection in a more diverse, challenging, and realistic scenario.
\textit{Realistic} here refers to a sensor/application-agnostic model deployment, where different factors can vary from one task to the next.
This allows for further exploration of the stability-plasticity dilemma, as they require models to retain diverse capabilities over longer sequences than ever before in IL for object detection while adapting to various distribution shifts.
Also, the curated and aligned datasets can be used for studies on \eg domain generalization~\cite{zhou_domain_2022}, domain adaptation~\cite{stahlbock_brief_2021}, multi-task learning~\cite{zhang_survey_2022}, and pre-training~\cite{chen_vlp_2023}.

We provide the following key contributions and insights using our novel D-RICO and EC-RICO benchmarks: 

\begin{itemize}
\item \textbf{Limits of Existing Approaches:} We demonstrate that \textbf{state-of-the-art IL methods struggle} on our novel diverse benchmarks, with widely used techniques such as distillation showing limited effectiveness.
\item \textbf{Replay:} We show that \textbf{a simple low-rate replay approach is a strong baseline} to mitigate forgetting.
\item \textbf{Stability-Plasticity Trade-off:} We empirically show that \textbf{stability and plasticity can counteract each other} in IL, highlighting that the focus should not only be put on mitigating forgetting. 
\end{itemize}

\begin{table}[t]
\centering
\scriptsize
\setlength{\tabcolsep}{3pt}
\caption{Comparison of the proposed benchmarks to a representative set of existing benchmarks and evaluation datasets.}
\label{tab:comp-different-benchmarks}
\begin{tabular}{@{}lcccccccc@{}}
\toprule
\textbf{Benchmark} & \textbf{Scen.} & \textbf{Obj.} & \textbf{Cls.} & \textbf{Tsks.} & \textbf{Dsets.} & \textbf{Lbls.}  & \textbf{Imgs.}
\\
\midrule

CORe50~\cite{lomonaco_core50_2017}
& \dilbox{DIL}      & \classbox{class.}     & 50    & 11    & 1     & 165k  & 165k
\\

DomainNet~\cite{peng_moment_2019}
& \dilbox{DIL}      & \classbox{class.}     & 345   & 6     & n/a   & 600k  & 600k
\\

CDDB~\cite{li_continual_2023}
& \dilbox{DIL}      & \classbox{class.}     & 12    & 12    & 5     & 842k  & 842k 
\\
\midrule

ImageNet~\cite{deng_imagenet_2009-1}
& \cilbox{CIL}      & \classbox{class.}     & 1000  & 10    & 1     & 1.3M  & 1.3M
\\

CIFAR~\cite{krizhevsky_learning_2009}
& \cilbox{CIL}      & \classbox{class.}     & 100   & 10    & 1     & 60k   & 60k
\\

Oxford Flowers~\cite{nilsback_automated_2008}
& \cilbox{CIL}      & \classbox{class.}     & 102   & 10    & 1     & 8k    & 8k
\\

Standford Cars~\cite{krause_3d_2013}
& \cilbox{CIL}      & \classbox{class.}     & 196   & 9     & 1     & 16k   & 16k
\\
\midrule

TiROD~\cite{pasti_tiny_2024-1}
& \dilbox{DIL}      & \objdetbox{obj. det.} & 13    & 10    & 1     & 18k   & 7k
\\

OAK~\cite{wang_wanderlust_2021}
& \dilbox{DIL}      & \objdetbox{obj. det.} & 105   & n/a   & 1     & 326k  & 27k
\\

CLAD-D~\cite{verwimp_clad_2023}
& \dilbox{DIL}      & \objdetbox{obj. det.} & 6     & 4     & 1     & 136k  & 19k
\\ 

VOC Series~\cite{song_non-exemplar_2024}
& \dilbox{DIL}      & \objdetbox{obj. det.} & 6     & 4     & 4     & 30k   & 12k
\\
\midrule

VOC 2007~\cite{everingham_pascal_2010}
& \cilbox{CIL}      & \objdetbox{obj. det.} & 20    & 10    & 1     & 25k   & 10k
\\

COCO 2017~\cite{fleet_microsoft_2014}
& \cilbox{CIL}      & \objdetbox{obj. det.} & 80    & 2     & 1     & 1.8M  & 164k
\\ \midrule

\rowcolor{gray!20} 
\rule{0pt}{3ex} D-RICO~(ours) & \dilbox{DIL} & \objdetbox{obj. det.} & 3 & 15 & 14 & 871k & 69k \\

\rowcolor{gray!20} 
\rule[-1.5ex]{0pt}{3.5ex} EC-RICO~(ours) & \ecdilbox{EC-DIL} & \objdetbox{obj. det.} & 8 & 8 & 8 & 578k & 40k \\

\bottomrule
\end{tabular}
\vspace{1mm} 
\caption*{\scriptsize Abbreviations: Scen. = IL Scenario, Obj. = Objective, Cls. = Classes, Tsk. = Number of tasks, Dsets. = Datasets, Lbls. = Number of labels, Img. = Number of images, class. = classification, obj. det. = object detection, DIL = domain incremental learning, CIL = class incremental learning, EC = expanding classes}
\end{table}
\section{Related Work}~\label{sec:related-work}

\textbf{Object Detection.}
Object detection models comprise two main categories: single-stage models~\cite{hussain_yolov1_2024}, prioritizing speed, and two-stage models~\cite{ren_faster_2017, cai_cascade_2021, fang_eva-02_2023, du_overview_2020}, emphasizing accuracy.
The Detection Transformer (DETR)~\cite{carion_end--end_2020} represents an advancement employing attention mechanisms for bounding box selection in single-stage models.

\noindent\textbf{Incremental Learning for Object Detection.}
Most IL for obejct detection research focuses on two-stage detectors~\cite{leonardis_bridge_2025, wagner_forgetting_2024, song_non-exemplar_2024, menezes_continual_2022}, although studies on IL for single-stage models~\cite{mo_multi-level_2025, shieh_continual_2020}, including DETR~\cite{dong_incremental-detr_2023, liu_continual_2023} also exist.

The two most popular approaches to mitigate catastrophic forgetting in IL for object detection are distillation~\cite{leonardis_bridge_2025, mo_multi-level_2025, dong_towards_2024, kang_alleviating_2023, dong_incremental-detr_2023, dong_class-incremental_2023, qian_contrastive_2022, gang_predictive_2022}, which is a form of regularization~\cite{wang_comprehensive_2024}, and rehearsal-based techniques that replay previous data~\cite{kim_sddgr_2024, monte_replay_2024, yang_one-shot_2023, liu_augmented_2023, shieh_continual_2020}.

Other approaches include representation-based methods~\cite{mo_multi-level_2025, lu_few-shot_2024}, optimization-based methods~\cite{joseph_incremental_2022, li_raise_2022, wang_opennet_2023}, and various hybrid or novel methods~\cite{yang_continual_2022, liu_continual_2023, ibrahim_node_2024}.
Compared to classification, which dominates IL research, significantly fewer approaches have been explored~\cite{wang_comprehensive_2024}.

\noindent\textbf{Datasets and Benchmarks.}
Currently, four dedicated DIL benchmarks exist for IL in object detection: TiROD~\cite{pasti_tiny_2024-1}, OAK~\cite{wang_wanderlust_2021}, CLAD-D~\cite{verwimp_clad_2023}, and Pascal Series~\cite{song_non-exemplar_2024}.
TiROD captures diverse environments using a dual-camera robot, whereas OAK focuses on online IL from a student's daily life.
CLAD-D reorganizes a subset of SODA10M~\cite{han_soda10m_2021} according to location, weather, and time, while Pascal Series extends Pascal VOC 2007~\cite{everingham_pascal_2010} to include domain shifts.
However, these benchmarks lack variations in sensors, bounding box statistics, annotation policies, and sequence lengths, and they exhibit limited diversity in geographic, perspective, and class distribution shifts.

Additionally, some works split datasets, such as BDD100k~\cite{yu_bdd100k_2020}, Cityscapes~\cite{cordts_cityscapes_2016}, and others~\cite{sheeny_radiate_2021, geiger_vision_2013, inoue_cross-domain_2018, georgakis_multiview_2016}, into individual tasks~\cite{bidaki_online_2025, witte_severity_2023, kalb_continual_2021, su_object_2024, song_non-exemplar_2024, liu_multi-task_2020-1}.
For class-incremental learning (CIL) in object detection, where each task introduces a new class, most studies split the Pascal VOC 2007~\cite{everingham_pascal_2010} or MS COCO~\cite{fleet_microsoft_2014} datasets by classes~\cite{wang_comprehensive_2024, menezes_continual_2022, liu_continual_2023, dong_incremental-detr_2023, song_non-exemplar_2024, dong_towards_2024, yang_one-shot_2023, shieh_continual_2020-1, dong_class-incremental_2023, zheng_contrast_2021, zhang_few-shot_2025}.
The diversity in these IL settings is even more limited.

Our proposed D-RICO and EC-RICO benchmarks differ from others in several ways: (1) they include more tasks, (2) the tasks are more diverse, (3) they are closer to real-world scenarios, and (4) they are more challenging.
\renewcommand{\arraystretch}{1.3}

\begin{table*}[htbp]
\scriptsize
\centering
\caption{Details on D-RICO (D) and EC-RICO (EC) benchmark. \textit{All three} means person, bicycle, and vehicle.}
\label{tab:data_table}
\begin{tabular}{cclllll}
\toprule
\multicolumn{2}{c}{\textbf{Taskorder}}  &&& \multicolumn{2}{c}{\textbf{Classes}} &\\
\cmidrule(lr){1-2} \cmidrule(lr){5-6} 
\multicolumn{1}{c}{\textbf{D}} & \multicolumn{1}{c}{\textbf{EC}} & \textbf{Task Name}       & \textbf{Dataset}                                               & \multicolumn{1}{c}{\textbf{D}} & \multicolumn{1}{c}{\textbf{EC}}               & \textbf{Short Description}     \\ \midrule
\dricobox{1}   & \ecricobox{3}     & \textit{daytime}               & nuImages~\cite{caesar_nuscenes_2020}                  & \dricobox{all three }        & \ecricobox{+ bicycle }            &
urban, daylight, real-world, vehicle-mounted, Singapore
\\

\dricobox{2}   & \ecricobox{7}     & \textit{thermal}               & Teledyne FLIR~\cite{teledyne_flir_free_2024}        & \dricobox{all three}         & \ecricobox{+ traffic light}       &
thermal, urban, varying lighting, weather conditions
\\

\dricobox{3}   & \ecricobox{4}     & \textit{fisheye fix}           & FishEye8K~\cite{gochoo_fisheye8k_2023}                & \dricobox{person, vehicle}   & \ecricobox{+ motorcycle}          &
fisheye, daytime, urban traffic, Taiwan, wide-angle, multi-camera        
\\

\dricobox{4}   & \ecricobox{6}     & \textit{drone}                 & VisDrone~\cite{zhu_detection_2022}                    & \dricobox{all three}         & \ecricobox{+ bus}                 &
drone, urban and rural, variable density, different lighting, 14 cities    
\\

\dricobox{5}   & \ecricobox{5}     & \textit{simulation}            & SHIFT~\cite{sun_shift_2022}                           & \dricobox{all three}         & \ecricobox{+ truck }              &
synthetic, urban driving, CARLA, daytime, clear weather       
\\

\dricobox{6}   & \ecricobox{1}     & \textit{fisheye car}           & WoodScape~\cite{yogamani_woodscape_2021}              & \dricobox{person, vehicle}   & \ecricobox{person}                &
fisheye, vehicle-mounted, driving perspectives, multiple positions    
\\

\dricobox{7}   & -     & \textit{RGB + thermal fusion}        & SMOD~\cite{chen_amfd_2024}                            & \dricobox{all three}        & -                     &
RGB-thermal fusion using IFCNN~\cite{zhang_ifcnn_2020}
\\

\dricobox{8}   & -     & \textit{video game}            & Sim10k~\cite{johnson-roberson_driving_2017}           & \dricobox{vehicle}           & -                     &
synthetic, urban, GTA V, diverse driving scenarios      
\\

\dricobox{9}   & \ecricobox{8}     & \textit{nighttime}             & BDD100K~\cite{yu_bdd100k_2020}                        & \dricobox{all three}        & \ecricobox{+ street sign}         &
urban, nighttime, perception challenge, street lighting
\\

\dricobox{10}  & -     & \textit{fisheye indoor}        & LOAF~\cite{yang_large-scale_2023}                     & \dricobox{person}            & -                     &
fisheye, indoor, overhead, 360° view, surveillance
\\

\dricobox{11}  & \ecricobox{2}     & \textit{gated}                 & DENSE~\cite{bijelic_seeing_2020}                      & \dricobox{person, vehicle}   & \ecricobox{+ car}                 &
gated, urban, various conditions, depth-enhanced imaging      
\\

\dricobox{12}  & -     & \textit{photoreal. simulation}      & Synscapes~\cite{wrenninge_synscapes_2018}             & \dricobox{person, vehicle}   & -                     &
photorealistic, synthetic, urban, physically based rendering     
\\

\dricobox{13}  & -     & \textit{thermal fisheye indoor}          & TIMo~\cite{schneider_timodataset_2022}                & \dricobox{person}            & -                     &
thermal fisheye, indoor, human actions, multiple perspectives
\\

\dricobox{14}  & -     & \textit{inclement}             & DENSE~\cite{bijelic_seeing_2020}                      & \dricobox{person, vehicle}   & -                     &
fog, snow, rain, adverse weather        
\\

\dricobox{15}  & -     & \textit{event camera}          & DSEC~\cite{gehrig_low-latency_2024, gehrig_dsec_2021} & \dricobox{all three}         & -                     &
event-based, driving, varied lighting, RGB overlay

\\\bottomrule
\end{tabular}
\end{table*}

\section{Benchmark}\label{sec:benchmark-rico}

This section defines the DIL problem and introduces the \textbf{D-RICO} and \textbf{EC-RICO} benchmarks, analyzes task affinity, and establishes metrics to measure IL performance.

\subsection{Preliminary}

We consider DIL for object detection, where a model sequentially encounters tasks with distinct domains but a shared label space.
We aim to learn the $t$-th task $\mathcal{T}_t$ using the data $\mathcal{D}_t= (\mathcal{X}_t, \mathcal{Y}_t)$, where $\mathcal{X}_t = \{\mathbf{x}_i^t\}_{i=1}^{n_t}$ consists of $n_t$ images with $\mathbf{x}_i^t \in \mathbb{R}^{H_i^t \times W_i^t \times C_i^t}$, where $H_i^t$, $W_i^t$ and $C_i^t$ are the image height, width, and the number of channels.
The corresponding labels are given by $\mathcal{Y}_t = \{\mathbf{y}_i^t\}_{i=1}^{n_t}$.

Each label $\mathbf{y}_i^t$ is a set of object annotations, expressed as $\mathbf{y}_i^t = \{(c_{i,j}^t, \mathbf{b}_{i,j}^t)\}_{j=1}^{m_i^t}$.
Here, $c_{i,j}^t \in \mathcal{C}_t$ denotes the class label of the $j$-th object in the i-th image $\mathbf{x}_i^t$ and $\mathcal{C}_t$ are all the classes of task $\mathcal{T}_t$, while $\mathbf{b}_{i,j}^t \in \mathbb{R}^4$ represents the respective bounding box coordinates. 
The number of objects per image varies and is denoted by $m_i^t$. The task ID is known during training but is not available to the model at inference time.
A model trained for task $\mathcal{T}_t$ on data $\mathcal{D}_t$ is denoted $\mathcal{M}_t$.

\subsection{Dataset Selection and Alignment}\label{sec:benchmark:dataset}

We choose autonomous driving and surveillance as our data sources because they are primary applications for object detection, and numerous datasets are available in these fields.

We select datasets based on criteria: (1) containing $\ge$~5,000 labeled images, (2) differ from other datasets in at least one structural or semantic aspect, (3) include annotations for common object classes in autonomous driving and surveillance, and (4) freely available for research.
Table~\ref{tab:data_table} provides an overview of all datasets we selected.

We process all datasets to enhance consistency while preserving dataset-specific characteristics. This processing involves:
(1) modifying images (\eg~merging modalities, rescaling);
(2) selecting image subsets to increase task distinction and improve data quality (\eg~retaining night images, removing unwanted objects, filtering similar frames); and
(3) aligning annotations (\eg~merging boxes, relabeling, removing annotations, unifying format).
Although (3) mitigates labeling policy contradictions, further alignment would necessitate manual labeling.
The 14 datasets are the basis for the 15 D-RICO and 8 EC-RICO tasks, with DENSE contributing two tasks.


\subsection{Domain RICO}\label{sec:benchmark:d-rico}

D-RICO has 15 tasks and 3 classes: person, bicycle, and vehicle.
'Persons' include individuals not in vehicles or on bicycles, while 'vehicles' cover all non-bicycle vehicles.
'Bicycles' include all bicycles, with or without riders.
Some tasks use a subset of the 3 classes, but all objects may still appear, adding complexity.
This resembles generalized-CIL~\cite{mi_generalized_2020}, with varying class counts per task, while focusing on object detection and domain IL.
Each task has 2,782 training, 423 validation, and 1,399 testing images.

The task order is randomized under the following constraints:
(1) \textit{daytime} is selected as the initial task, as it represents the most fundamental scenario,
(2) \textit{gated} and \textit{inclement} are not back-to-back, as they originate from the same dataset, and
(3) tasks with similar camera sensors are not placed consecutively.
This ensures a logical sequence starting from the most common scenario, minimizing dataset-specific biases, and maintains diversity in sensor characteristics to prevent sequential redundancy.

Example images for each task are shown in Figure~\ref{fig:fig1_tasks}, highlighting their visual distinctions.
Table~\ref{tab:data_table} lists all the labeled classes per task and Figure~\ref{fig:fig2} e) illustrates the corresponding variations.
The differences in spatial distribution, area, and aspect ratio for the bounding boxes are shown in Figure~\ref{fig:fig2} b) - d).
Furthermore, with a sequence length of 15 tasks, D-RICO is longer than all existing benchmarks and evaluation settings (see Table~\ref{tab:comp-different-benchmarks}).
We conduct quantitative comparisons to two existing benchmarks and demonstrate that D-RICO is more challenging and more diverse (see Suppl. Mat. Section D1).

\subsection{Expanding-Classes RICO}\label{sec:benchmark:ec-rico}

As a second benchmark we propose EC-RICO, as in many applications, the number of required classes increases over time due to changing requirements~\cite{zhou_deep_2023}.
It consists of 8 tasks, with the labels for person, car, bicycle, motorcycle, truck, bus, traffic light, and street sign.

For each task, the label space expands by introducing one additional class, \ie $|\mathcal{C}_{t+1}| = |\mathcal{C}_{t}| + 1$, with $|\mathcal{C}_t|$ being the number of classes in $\mathcal{C}_t$.
The current task has labels for the current and all previous classes.
Objects of future classes, while unlabeled, may still be present in the images.

We adopt this expanding-classes setting instead of CIL to keep the focus on DIL and to avoid requiring the model to re-learn previous classes in new domains without direct supervision.
The task order is determined based on the availability of class labels in the underlying datasets.

Figure~\ref{fig:fig1_tasks} shows example images for each task. 
Figure~\ref{fig:fig2} g) - j) illustrates the class and bounding box statistics, which remain as diverse as in D-RICO.
Each task has 3,040 training, 511 validation, and 1,417 testing images. 

\subsection{Assessing Task Transferability and Similarity}\label{subsec:bench:task-affinity}
We adapt the task affinity (TA) metric from~\cite{zamir_taskonomy_2018} to study the task's transferability and similarity influence on the IL performance in Section~\ref{subsec:res:task-order}.
Following~\cite{zamir_taskonomy_2018}, we measure the TA from task $\mathcal{T}_i$ to $\mathcal{T}_j$ by fine-tuning model $\mathcal{M}_i$ on $\mathcal{D}_j$ (see Section~\ref{subsec:exp:setup} for details).
We adapt only the output layers, \ie a small portion of the weights, to quantify how well the representations learned from $\mathcal{D}_i$ can be used for $\mathcal{T}_j$. Formally, TA from task $\mathcal{T}_i$ to $\mathcal{T}_j$ is defined as $\text{TA}(i\rightarrow j) = \frac{\hat{a}_{i\rightarrow j}}{\tilde{a}_{j}}$, where $\hat{a}_{i\rightarrow j}$ is the performance (see Section~\ref{sec:benchmark:metrics}) of $\mathcal{M}_i$ after finetuning on $\mathcal{D}_j$, and $\tilde{a}_{j}$ is the performance of $\mathcal{M}_j$ trained from scratch.

Figure~\ref{fig:fig2} a) presents the TA matrix for D-RICO and f) for EC-RICO.
The TA matrix reveals asymmetry, highlighting that transferability is direction-dependent.
Moreover, the variation in TA scores suggests that some tasks exhibit more similarity than others, underscoring the diversity and challenges posed by D-RICO and EC-RICO.

\subsection{Evaluation Metrics}\label{sec:benchmark:metrics}
We evaluate IL performance using metrics from \cite{wang_comprehensive_2024, chaudhry_riemannian_2018}, with mean Average Precision $\mAP$~\cite{menezes_continual_2022} as our base performance measure.
Three aspects are assessed: 

\begin{enumerate}
    \item \textbf{Overall performance.} 
    We quantify this via average $\mAP$, denoted as $\overline{\mAP}$. Given $\mAP_{k,j}$ as the $\mAP$ on $\mathcal{D}_j$ test set after learning $\mathcal{T}_k$ ($j \leq k$), $\overline{\mAP}$ after the $k$-th tasks is
    \begin{equation}
        \overline{\mAP}_k = \frac{1}{k} \sum_{j=1}^{k} \mAP_{k,j},
    \end{equation}
    with higher values indicating better performance.
    
    \item \textbf{Memory stability.}
    The \textit{forgetting measure} ($\mathrm{FM}$) quantifies the stability by computing the difference between the best previous and current performance of $\mathcal{T}_j$. After training of $\mathcal{M}_k$, $\mathrm{FM}_k$ is given as
    \begin{equation}
        \mathrm{FM}_k = \frac{1}{k-1} \sum_{j=1}^{k-1} \max_{1\le l\le k-1} \left(\mAP_{l,j} - \mAP_{k,j} \right).
    \end{equation}
    Higher $\mathrm{FM}$ values indicate greater forgetting, while negative values suggest task improvement.

    \item \textbf{Learning plasticity.} We quantify a model's task adaptation capability using two metrics:
    \begin{enumerate}
        \item \textit{Forward transfer} ($\mathrm{FWT}$) compares $\mAP$ to reference models trained individually, measuring how prior learning influences new tasks.
        \begin{equation}
            \mathrm{FWT}_k = \frac{1}{k-1} \sum_{j=2}^{k} \left(\mAP_{j,j} - \mAP'_{j}\right),
        \end{equation}
        where $\mAP'_{j}$ is the isolated reference model's $\mAP$ of $\mathcal{T}_j$. Positive values indicate superior adaptation to new tasks.
        
        \item \textit{Intransigence measure} ($\mathrm{IM}$) evaluates new task learning difficulty by comparing to the joint training $\mAP$
        \begin{equation}
            \mathrm{IM}_k = \frac{1}{k} \sum_{j=1}^{k} \left( \mAP_{j,j} - \mAP^*_j\right),
        \end{equation}
        where $\mAP^*_j$ is the $\mAP$ of the $j$-th task test set on a reference model jointly trained on $\cup_{j=1}^{T} \mathcal{D}_j$. Positive values show greater plasticity.
    \end{enumerate}
\end{enumerate}

The metrics after learning all $T$ total tasks are denoted by $\overline{\mAP}=\overline{\mAP}_T$, $\mathrm{FM}=\mathrm{FM}_T$, $\mathrm{FWT}=\mathrm{FWT}_T$, and $\mathrm{IM}=\mathrm{IM}_T$.

The desired goal is that IL methods exceed individual and joint model performance via cross-task knowledge transfer, \ie $\overline{\mAP}>\frac{1}{T} \sum_{j=1}^{T} \mAP_{j}'$.
This requires approaches with high plasticity and low forgetting rate.

\subsection{How to Access and Utilize RICO}
To support research on incremental object detection challenges, we release code and tools for working with our D-RICO and EC-RICO benchmarks. Our release includes:

\begin{enumerate}[leftmargin=1cm]
    \item scripts and instructions for downloading and setting up the D-RICO and EC-RICO benchmark datasets; and
    \item a customized, IL-optimized version of Detectron2~\cite{wu_detectron2_2019}, enabling the development and evaluation of new models, methods, and benchmarks.
\end{enumerate}

\noindent More details on D-RICO and EC-RICO can be found in the Supplementary Material.

\begin{table*}[htbp]
\centering
\footnotesize
\caption{Results for domain RICO and expanding-classes RICO benchmark. Best in IL methods in bold.}
\label{tabel:main-results}
\begin{tabular}{lllllllll} 
\toprule
 & \multicolumn{4}{c}{\textbf{Domain RICO}} 
 & \multicolumn{4}{c}{\textbf{Expanding-Classes RICO}} \\ 
\cmidrule(lr){2-5} \cmidrule(lr){6-9}
  \textbf{Method}  & \multicolumn{1}{c}{\textbf{$\overline{\bm{\mAP}}$
  \textuparrow}} & \multicolumn{1}{c}{\textbf{$\mathbf{FM}$  \textdownarrow}}  & \multicolumn{1}{c}{\textbf{$\mathbf{FWT}$ 
 \textuparrow}} & \multicolumn{1}{c}{\textbf{$\mathbf{IM}$ \textuparrow}} &\multicolumn{1}{c}{\textbf{$\overline{\bm{\mAP}}$ \textuparrow}} & \multicolumn{1}{c}{\textbf{$\mathbf{FM}$  \textdownarrow}} & \multicolumn{1}{c}{\textbf{$\mathbf{FWT}$ \textuparrow}} &\multicolumn{1}{c}{\textbf{$\mathbf{IM}$ \textuparrow}} \\ 
\midrule
Joint Training   
                    & $43.75{\scriptstyle \pm0.03}$     &                                   &                                  &   
                    & $38.46{\scriptstyle \pm0.06}$     &                                   &                                   &   \\

Individual Training 
                    & $49.37{\scriptstyle \pm0.13}$     &                                   &                                   &   
                    & $45.54{\scriptstyle \pm0.01}$     &                                   &                                   &   \\

\midrule

Naïve FT    
                    & $30.60{\scriptstyle \pm0.35}$     & $19.43{\scriptstyle \pm0.30}$     & $\mathbf{-0.63}{\scriptstyle \pm0.08}$     & $\mathbf{4.95} {\scriptstyle\pm 0.07}$ 
                    & $37.51{\scriptstyle \pm0.1}$      & $8.62{\scriptstyle \pm0.05}$      & $\mathbf{-0.23}{\scriptstyle \pm0.1}$      & $\mathbf{6.52} {\scriptstyle\pm 0.06}$\\

\midrule

Replay 1\%          
                    & $40.20{\scriptstyle \pm0.19}$     & $8.91{\scriptstyle \pm0.26}$      & $-0.79{\scriptstyle \pm0.15}$     & $4.74 {\scriptstyle\pm 0.13}$ 
                    & $\mathbf{38.09}{\scriptstyle \pm0.06}$     & $5.83{\scriptstyle \pm0.23}$      & $-2.17{\scriptstyle \pm0.16}$     & $4.66 {\scriptstyle\pm 0.16}$ \\

Replay 10\%
                    & $43.06{\scriptstyle \pm0.14}$     & $4.40{\scriptstyle \pm0.10}$      & $-2.23{\scriptstyle \pm0.07}$     & $3.39 {\scriptstyle\pm 0.07}$ 
                    & $37.2{\scriptstyle \pm0.1}$       & $3.82{\scriptstyle \pm0.12}$      & $-5.24{\scriptstyle \pm0.02}$     & $2.02 {\scriptstyle\pm 0.01}$ \\

Replay 25\%         
                    & $\mathbf{43.43}{\scriptstyle \pm0.58}$     & $3.35{\scriptstyle \pm0.64}$      & $-2.88{\scriptstyle \pm0.06}$     & $2.77 {\scriptstyle\pm 0.05}$ 
                    & $37.53{\scriptstyle \pm0.35}$     & $2.21{\scriptstyle \pm0.3}$       & $-6.44{\scriptstyle \pm0.14}$     & $0.93 {\scriptstyle\pm 0.10}$ \\

\midrule

ABR~\cite{liu_augmented_2023} 
                    & $32.81 {\scriptstyle\pm 0.66}$     & $15.26 {\scriptstyle\pm 0.71}$    & $-2.46 {\scriptstyle\pm 0.05}$   & $3.28 {\scriptstyle\pm 0.04}$ 
                    & $38.04 {\scriptstyle\pm 0.11}$    & $7.05 {\scriptstyle\pm 0.22}$    & $-1.21 {\scriptstyle\pm 0.08}$    & $5.68 {\scriptstyle\pm 0.08}$\\

Meta-ILOD~\cite{joseph_incremental_2022} 
                    & $38.51 {\scriptstyle\pm 0.17}$   & $9.26 {\scriptstyle\pm 0.25}$   & $-3.97 {\scriptstyle\pm 0.50}$   & $1.76 {\scriptstyle\pm 0.49}$ 
                    & $36.99 {\scriptstyle\pm 0.16}$    & $6.42 {\scriptstyle\pm 0.25}$     & $-2.79 {\scriptstyle\pm 0.09}$    & $4.08 {\scriptstyle\pm 0.06}$\\

BPF~\cite{leonardis_bridge_2025} 
                    & $30.74 {\scriptstyle\pm 1.00}$    & $19.13 {\scriptstyle\pm 1.04}$    & $-0.76 {\scriptstyle\pm 0.03}$    & $4.82 {\scriptstyle\pm 0.04}$
                    & $37.32 {\scriptstyle\pm 0.13}$   & $8.07 {\scriptstyle\pm 0.12}$    & $-0.85 {\scriptstyle\pm 0.07}$   & $5.86 {\scriptstyle\pm 0.04}$\\

LDB~\cite{song_non-exemplar_2024}
                    & $42.49 {\scriptstyle\pm 0.17}$   & $\mathbf{1.21} {\scriptstyle\pm 0.13}$       & $-6.17 {\scriptstyle\pm 0.06}$   & $-0.15 {\scriptstyle\pm 0.06}$ 
                    & $28.88 {\scriptstyle\pm 0.25}$  & $\mathbf{2.16}{\scriptstyle\pm 0.14}$
                    & $-16.6 {\scriptstyle\pm 0.24}$  & $-7.76 {\scriptstyle\pm 0.22}$\\

\bottomrule
\end{tabular}
\end{table*}

\section{Experiments}
This section shows simple baselines outperform SOTA IL methods on D-RICO and EC-RICO, with analysis revealing factors behind this and insights for improving IL methods.

\subsection{Experimental Setup}\label{subsec:exp:setup}

We employ a two-stage object detection setup following~\cite{liu_augmented_2023, joseph_incremental_2022, leonardis_bridge_2025, song_non-exemplar_2024}. 
Similar to~\cite{song_non-exemplar_2024}, we use a ViTDet~\cite{li_exploring_2022}-based approach.
We adopt the EVA-02-L backbone~\cite{fang_eva-02_2023}, which has shown strong performance on MS COCO~\cite{fleet_microsoft_2014, fang_eva-02_2023}.
Following~\cite{fang_eva-02_2023}, we use the Cascade Faster R-CNN detection head~\cite{cai_cascade_2017}.
We initialize the backbone with weights finetuned on Objects365~\cite{shao_objects365_2019} and MS COCO~\cite{fleet_microsoft_2014}, while randomly initializing the detection head. 
We freeze the backbone during training; experiments with an unfrozen backbone are available in the Supplementary Material.

For loss computation, we mask loss terms for absent classes~\cite{dollar_pedestrian_2012, ren_faster_2017, braun_eurocity_2019}. For EC-RICO, we begin with an 8-class head but optimize only the current classes. Each task is trained for 700 iterations with a batch size of 20. Images are resized to 1536$\times$1536 with rescaling, crop, brightness, contrast, and color jittering. We use AdamW~\cite{loshchilov_decoupled_2019-1} with warm-up~\cite{kalra_why_2024} and cosine scheduling~\cite{loshchilov_sgdr_2017}. Due to task diversity, task-specific normalization is applied. A small hyperparameter search is conducted, with final settings evaluated over three runs using different random initializations (mean $\pm$ standard deviation reported).

\subsection{Methods}\label{sec:exp:methods}
We conduct joint and individual task training as references, use naïve finetuning as a lower bound, and use three replay settings as simple baselines.
From the literature, we select Meta-ILOD~\cite{joseph_incremental_2022} as a strong approach in CIL, BPF~\cite{leonardis_bridge_2025} as SOTA on the common Pascal VOC 2007~\cite{everingham_pascal_2010} IL setting for rehearsal-free CIL, ABR~\cite{liu_augmented_2023} as SOTA for rehearsal-based CIL, and LDB~\cite{song_non-exemplar_2024} as SOTA for DIL.
We include CIL methods into the analysis to assess their suitability in DIL.

\begin{enumerate}
    \item \textbf{Joint Training} trains all tasks $\cup_{j=1}^{T} \mathcal{D}_j$ simultaneously.
    Testing is performed on individual task test sets.
    
    \item \textbf{Individual Training} trains and evaluates each task independently using separately trained models.
    
    \item \textbf{Naïve Finetuning (FT)} sequentially trains tasks without employing any forgetting mitigation techniques.
    
    \item \textbf{Replay} rehearses 1\%, 10\%, or 25\% of randomly selected images from previous tasks in the current task. The 1\% matches the small replay size in~\cite{ostapenko_continual_2022}, while 10\% and 25\% exceed medium and large sizes but are less efficient due to random sampling.
    The replay buffer grows indefinitely, expanding with the selected percentage of images and labels.
    
    \item \textbf{ABR}~\cite{liu_augmented_2023} prevents catastrophic forgetting by replaying objects from previous tasks using mixup~\cite{zhang_mixup_2018} and mosaic~\cite{bochkovskiy_yolov4_2020}.
    It incorporates two distillation losses: one for region-of-interest features and another for class logits.
    
    \item \textbf{Meta-ILOD}~\cite{joseph_incremental_2022} learns gradient preconditioning matrices that guide the optimization toward a minimum that benefits previous tasks.
    These matrices are represented by layers in the detection head and are optimized independently using data from previous tasks.
        
    \item \textbf{BPF}~\cite{leonardis_bridge_2025} employs pseudo-labels generated by the previous task model.
    It utilizes a distillation loss that incorporates knowledge from past and expert models.

    \item \textbf{LDB}~\cite{song_non-exemplar_2024} trains a base model initially and adapts it for subsequent tasks by adding task-specific output layers and bias terms to the backbone.
    The bias terms and output layer are selected using a nearest-mean classifier (NMC) at inference.
\end{enumerate}

\subsection{D-RICO Results}

D-RICO poses a significant challenge due to its long sequence of diverse tasks. As shown in Table~\ref{tabel:main-results}, naïve FT yields the lowest $\overline{\mAP}$, falling 13\% and 19\% below joint and individual training, respectively. Replaying only 1\% of the data improves $\overline{\mAP}$ by 10\%, with further gains observed at higher replay levels. LDB almost reaches the performance of 10\% replay, while Meta-ILOD performs comparably to 1\% replay, and ABR and BPF offer only marginal improvements over naïve FT.

\begin{figure}
    \centering
    \includegraphics[width=1\linewidth]{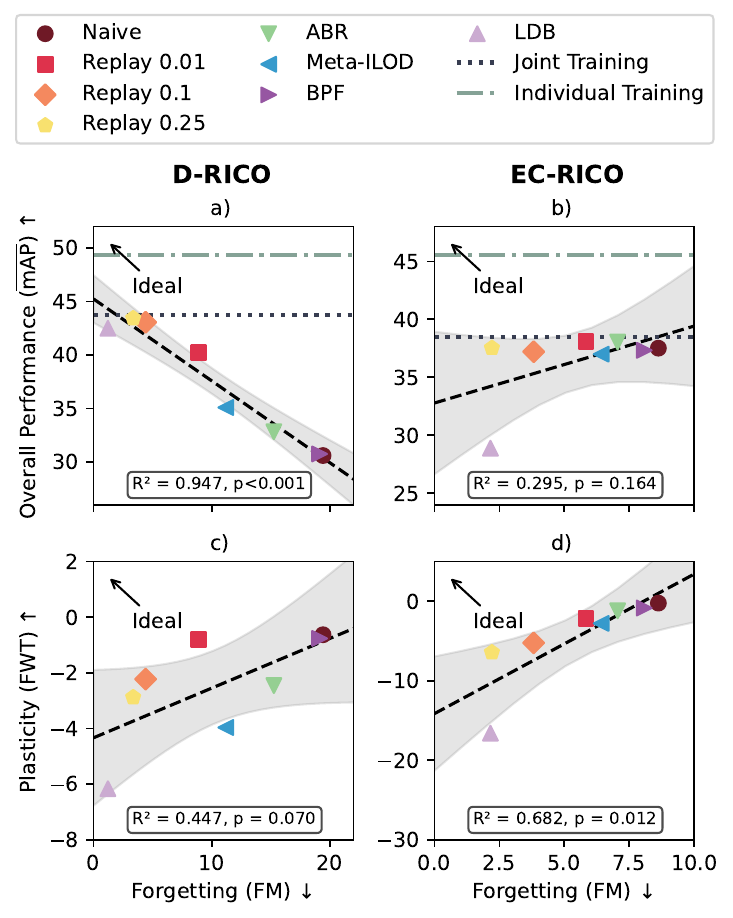}
    \caption{Correlations between Forward Transfer, Forgetting, and Performance across experiments, with trend lines and 95\% confidence intervals. Current methods remain far from the ideal of high performance \& plasticity with low forgetting. All units are $\mAP$.}
    \label{fig:exp:corr-experiments}
\end{figure}

Regarding $\mathrm{FM}$, naïve FT exhibits the greatest forgetting, whereas replay strategies steadily forget less.
LDB achieves the lowest $\mathrm{FM}$ value, even below that of 25\% replay.
Figure~\ref{fig:exp:corr-experiments}~a) shows a linear correlation between lower $\mathrm{FM}$ and higher $\overline{\mAP}$.
However, even zero forgetting would not match individual training performance under the current trend.
Figure~\ref{fig:exp:corr-experiments}~c) further illustrates this, as no experiment achieves both high plasticity and low forgetting.

The 6\% gap between joint and individual training $\overline{\mAP}$ shows that a single model cannot fully capture all tasks.
IL experiments, except for LDB, use a single model. LDB learns task-specific layers, but since these comprise a small fraction of total parameters, it exhibits low plasticity (see Section~\ref{sec:exp:model-expension}). Low standard deviations in all experiments confirm robustness regarding initializations.

\subsection{EC-RICO Results}

The EC-RICO benchmark raises complexity by increasing the number of classes per task (see Section~\ref{sec:benchmark:ec-rico}). 
Table~\ref{tabel:main-results} and Figure~\ref{fig:exp:corr-experiments} b) show that all IL experiments, except LDB, achieve $\overline{\mAP}$ comparable to joint training.
The gap between joint and individual training highlights again that a single model cannot effectively integrate all tasks.

Although $\overline{\mAP}$ remains similar, stability, and plasticity vary.
Naïve FT exhibits the highest forgetting, while replay mitigates it.
ABR, Meta-ILOD, and BPF yield $\mathrm{FM}$ values between naïve FT and 1\% replay, with LDB offering the highest stability, similar to 25\% replay.

Regarding plasticity, naïve FT adapts best to new tasks, while replay lowers plasticity.
ABR, Meta-ILOD, and BPF achieve $\mathrm{FWT}$ and $\mathrm{IM}$ values between naïve FT and 1\% replay again.
Like in D-RICO, LDB shows a drastically lower plasticity than all other methods (see Section~\cref{sec:exp:model-expension}).

All methods exhibit low forgetting or high plasticity, but not both, leaving a gap in the upper left area (Figure~\ref{fig:exp:corr-experiments}~b) and~d)). 
Unlike D-RICO and other benchmarks, EC-RICO is unique as almost all methods achieve similar AA, yet differ significantly in stability and plasticity.

\subsection{Model Extension}\label{sec:exp:model-expension}
Except for LDB, all IL experiments share their weights across all tasks.
LDB trains certain parts of the model independently for each task and chooses them based on the NMC during inference.
Evaluating the NMC shows over 90\% accuracy for both benchmarks in task selection.

However, LDB freezes all parameters after the first task except for the domain biases and output layers.
As shown in Section~\ref{subsec:bench:task-affinity} and Figure~\ref{fig:fig2} a) and f), adapting only the output layer is insufficient—particularly for EC-RICO, where transferring from the first task leads to low performance in the following.
This means that LDB is limited by model plasticity, as seen in $\mathrm{FWT}$ and $\mathrm{IM}$ for EC-RICO (Table~\ref{tabel:main-results}).

\subsection{Distillation}\label{sec:exp:distill}
The main results in Table~\ref{tabel:main-results} show ABR, Meta-ILOD and BPF underperform.
In addition to other mechanisms, all three methods rely on distilling knowledge from the previous task model into the current one—a process we recognize as a form of knowledge transfer through regularization. 
This assumes that the previous task model can make meaningful predictions on the current task data, \ie is a good teacher.
This, however, is not guaranteed in a diverse task setup.
To show that the previous model is a poor teacher for the current task, we measure the next task $\overline{\mAP}$ as $\overline{\mAP}_{\mathrm{nt}} = \frac{1}{T-1} \sum_{j=2}^{T} a_{j-1,j}^{C_{j-1}}$,
where $a_{k,j}^{C_i}$ is the accuracy of the test set of task $j$ for the classes $C_i$ of task $i$ after learning task $k$.
Table~\ref{tabel:distill_next_task_transfer} shows that the previous model performs poorly on the current task, showing that distillation cannot transfer substantial knowledge.

\begin{table}[htbp]
\centering
\footnotesize

\caption{Comparing performance on the next task, demonstrating that the previous model is a weak teacher in distillation. Best in bold.}
\label{tabel:distill_next_task_transfer}
\begin{tabular}{lll} 
\toprule

 & \multicolumn{2}{c}{\textbf{$\overline{\bm{\mAP}}_{\mathrm{nt}}$\textuparrow}} 
\\ 
\cmidrule(lr){2-3} 
  \textbf{Method}  & \multicolumn{1}{c}{\textbf{D-RICO}} & \multicolumn{1}{c}{\textbf{EC-RICO}} \\ 
\midrule

Naïve FT    
                    & $27.48 {\scriptstyle\pm 0.08}$    & $23.46 {\scriptstyle\pm 0.05}$ \\

Replay 10\%
                    & $28.11 {\scriptstyle\pm 0.23}$    & $22.73 {\scriptstyle\pm 0.16}$ \\

ABR~\cite{liu_augmented_2023} 
                    & $25.15 {\scriptstyle\pm 0.05}$    & $\mathbf{23.94} {\scriptstyle\pm 0.02}$ \\

Meta-ILOD~\cite{liu_augmented_2023} 
                    & $\mathbf{28.70} {\scriptstyle\pm 0.11}$    & $23.41 {\scriptstyle\pm 0.04}$ \\

BPF~\cite{leonardis_bridge_2025} 
                    & $27.79 {\scriptstyle\pm 0.14}$   & $23.58 {\scriptstyle\pm 0.01}$ \\
\bottomrule
\end{tabular}
\end{table}

\subsection{Task Order}\label{subsec:res:task-order}
The task affinity matrix in Figure~\ref{fig:fig2} suggests that the order in which tasks are learned significantly impacts performance.
Similar to~\cite{de_lange_continual_2021}, we evaluate D-RICO under five different task orders to analyze this effect in more detail:

\begin{enumerate}[leftmargin=0.6cm, labelsep=0.2cm, label=\textbf{\Alph*)}]
    \item default task sequence (as in Figure~\ref{fig:fig1_tasks});
    \item reverse of order \textbf{A)};
    \item \textit{daytime} task $\rightarrow$ next by highest task affinity;
    \item highest test $\mAP$  $\rightarrow$ next by highest task affinity;
    \item lowest test $\mAP$ $\rightarrow$ next by lowest task affinity;
\end{enumerate}

Table~\ref{tab:results:task-order} compares naïve FT and replay 10\% across the five task orders.
For naïve FT, the reverse order \textbf{B)} achieves the highest performance in terms of $\overline{\mAP}$ and $\mathrm{FM}$, followed by \textbf{E)}.
The remaining three orders yield similar results.
For replay, $\overline{\mAP}$, $\mathrm{FM}$, and $\mathrm{IM}$ remain comparable across task orders, but $\mathrm{FWT}$ varies significantly.
The results indicate no correlation to the task affinity.

\begin{table}[htbp]
\centering
\setlength{\tabcolsep}{3pt}
\scriptsize
\caption{Results on the D-RICO for 5 task orders. Best in bold.}
\label{tab:results:task-order}
\begin{tabular}{lccccc}
\toprule
                \textbf{Method}          & \textbf{Order}            & \textbf{$\overline{\bm{\mAP}}$ \textuparrow} & \textbf{$\mathbf{FM}$  \textdownarrow} & \textbf{$\mathbf{FWT}$ \textuparrow} & \textbf{$\mathbf{IM}$ \textuparrow} \\ \midrule

            & A)            & $30.60$ & $19.43$ & $-0.63$ & $4.95$  \\
            & B)            & $\mathbf{38.29}$ & $\mathbf{11.31}$ &  $1.77$ & $4.53$  \\
Naïve FT       & C)            & $31.73$ & $18.41$ & $-0.67$ & $4.90$  \\
  & D)            & $31.68$ & $18.49$ & $-2.05$ & $\mathbf{5.16}$  \\
            & E)            & $33.93$ & $16.14$ &  $\mathbf{1.96}$ & $4.78$  \\
            \midrule
            
            & A)            & $43.05$ & $4.40$ & $-2.23$ & $\mathbf{3.39}   $ \\
            & B)            & $\mathbf{43.30}$ & $\mathbf{3.74}$ & $-0.52$ & $2.32$ \\
Replay 10\% & C)            & $42.40$ & $3.98$ & $-3.68$ & $2.05$ \\
            & D)            & $42.49$ & $4.27$ & $-4.09$ & $2.54$ \\
            & E)            & $42.18$ & $5.46$ & $\mathbf{-0.01}$ & $2.82$ \\
            \bottomrule
\end{tabular}%
\end{table}

\noindent See Supplementary Material for additional experiments and details on the setup.
\section{Discussion}

\noindent\textbf{Regarding D-RICO and EC-RICO as a Benchmark.}
D-RICO represents a long sequence of diverse tasks; EC-RICO also introduces a new class for each task.
Compared to other benchmarks, they introduce many different distribution shifts in the image and annotation space.
We showed that all methods fail to reach individual training performance on D-RICO and EC-RICO (see Table~\ref{tabel:main-results}), demonstrating their unique challenge for IL.
\textit{$\blacktriangleright$~D-RICO and EC-RICO are two diverse and challenging benchmarks for IL.}

\medskip 
\noindent\textbf{Regarding the Stability-Plasticity Dilemma.}
The $\overline{\mAP}$ and $\mathrm{FM}$ metrics are commonly reported in IL; we also analyze $\mathrm{FWT}$ and $\mathrm{IM}$ to measure plasticity.
The results (see Figure~\ref{fig:exp:corr-experiments}) show a trade-off between stability ($\mathrm{FM}$) and plasticity ($\mathrm{FWT}$), especially in EC-RICO, where their balance stabilizes $\overline{\mAP}$.
As also shown in~\cite{kim_stability-plasticity_2023}, most methods prioritize stability, but enhancing plasticity is vital to achieving individual training performance and advancing IL.
\textit{$\blacktriangleright$~Future research should give more attention to model plasticity.}

\medskip
\noindent\textbf{Regarding Distillation.}
Knowledge distillation, a common anti-forgetting strategy in CIL, underperforms on D-RICO and EC-RICO due to weak teachers, as shown by the \textit{next task} $\mAP$ (see Table~\ref{tabel:distill_next_task_transfer}) and task affinity matrix (see Figure~\ref{fig:fig2} a) and f)). Similar limitations for DIL are noted in~\cite{kalb_continual_2022}.
\textit{$\blacktriangleright$~Distillation is suboptimal for mitigating forgetting in challenging DIL.}

\medskip
\noindent\textbf{Regarding a Single Model.}
All experiments fail to achieve the $\overline{\mAP}$ of individual training (Table~\ref{tabel:main-results}), indicating that a single model struggles with task diversity and contradictions.
This limitation is also noted in~\cite{doan_continual_2023}.
Task-specific weights might be beneficial, but LDB (Section~\ref{sec:exp:model-expension}) and task affinity (Section~\ref{subsec:bench:task-affinity}) results indicate that significant model adjustments are necessary.
However, efficiency and knowledge transfer challenge task-specific weights. 
\textit{$\blacktriangleright$~Standard model architectures are inadequate in challenging DIL.}

\medskip 
\noindent\textbf{Regarding Task Order and Task Affinity.}
We show that weak IL methods (\eg, naïve FT) yield different results for varying task orders, like in~\cite{witte_severity_2023, bell_effect_2022}; however, task affinity does not predict this order (see Section~\ref{subsec:res:task-order}). 
Strong replay-based methods remain largely unaffected, consistent with the results in~\cite{de_lange_continual_2021}.
\textit{$\blacktriangleright$~Task order has limited relevance for strong IL methods.}

\medskip 
\noindent\textbf{Regarding Replay.}
Replay is a simple yet highly effective approach for mitigating forgetting, outperforming other methods (Table~\ref{tabel:main-results}).
While this study used random sampling, smarter selection strategies could improve efficiency.
These findings align with~\cite{van_de_ven_three_2022, zhang_energy_2024, smith_adaptive_2024, wang_random_2024}.
However, increasing replay size only converges to joint training performance and can not reach individual training performance.
\textit{$\blacktriangleright$~Replay serves as a strong baseline but is limited in plasticity.}

\medskip 
\noindent\textbf{Regarding the Limitations.}
We introduce D-RICO and EC-RICO as novel DIL benchmarks and show that SOTA methods struggle in these settings.
Our results suggest that existing approaches such as distillation, replay, and model extensions fail to balance plasticity and stability, preventing them from matching individual training performance while mitigating forgetting over long task sequences.
However, these findings are limited to the presented DIL benchmarks. 
Building on D-RICO and EC-RICO, a natural next step would be to explore additional benchmarks, including online, few-shot, and class IL.
\textit{$\blacktriangleright$~D-RICO and EC-RICO lay the foundation for future challenging IL benchmarks.}

\section{Conclusion}
We introduce two novel benchmarks, D-RICO and EC-RICO, for more realistic incremental object detection, addressing limitations in current IL benchmarks. Unlike prior work, they are built from multiple diverse object detection datasets spanning automotive driving and surveillance, ensuring diversity in image characteristics, bounding box and class distributions, and labeling policies and quality. D-RICO includes 15 tasks with a fixed class set, while EC-RICO has 8 tasks, each introducing a new domain and classes. Our evaluation of SOTA IL methods shows all underperform compared to a simple replay strategy, with distillation particularly failing. Plasticity proves more crucial than previously indicated, while task order has limited impact. The difference between joint and individual task training highlights the challenges of continual learning within a single model, indicating the need for new approaches in complex scenarios. We anticipate D-RICO and EC-RICO will advance IL methods for real-world challenges.

{
    \small
    \bibliographystyle{ieeenat_fullname}
    \bibliography{first-paper}
}

\clearpage
\setcounter{page}{1}
\appendix
\renewcommand{\thefigure}{S.\arabic{figure}}
\renewcommand{\thetable}{S.\arabic{table}}
\setcounter{figure}{0}
\setcounter{table}{0}

\maketitlesupplementary

\section{Using the RICO Benchmark}\label{supp:sec:using-benchmark}
Domain RICO (D-RICO) and Extending-Classes RICO (EC-RICO) are developed as a platform to advance research in Incremental Learning (IL) within more realistic settings.
We propose this benchmark to serve dual purposes: (1) as a challenging evaluation framework for future IL methods and (2) as a resource for deriving novel insights into IL mechanisms.
We have made the benchmark set up and evaluation framework openly accessible to facilitate broader research engagement.
This section provides an overview of the benchmark utilization process, with additional technical specifications available in the GitHub repository.
Details on using our training and evaluation framework are given in Section~\ref{supp:sec:il-detectron2}.

Since we do not hold rights to the constituent datasets, we cannot distribute the complete benchmark as a unified package.
Utilizing the benchmark therefore requires the following steps:
\begin{enumerate}[leftmargin=1cm]
    \item Downloading the original datasets from their respective sources;
    \item Processing the images and annotations;
    \item Combining the processed images and annotations into D-RICO and EC-RICO.
\end{enumerate}

The following subsections elaborate on each of these steps.

\noindent \textbf{1. Downloading the Datasets}
The initial step requires acquiring all constituent datasets from their original sources.

\begin{table}[H]
\centering
\begin{tabular}{p{0.4\linewidth}p{0.4\linewidth}}
\textbullet \quad nuImages \cite{caesar_nuscenes_2020} & 
\textbullet \quad SHIFT \cite{sun_shift_2022} \\ 
\textbullet \quad BDD100K \cite{yu_bdd100k_2020} &
\textbullet \quad Teledyne FLIR \cite{teledyne_flir_free_2024} \\
\textbullet \quad WoodScape \cite{yogamani_woodscape_2021} & 
\textbullet \quad LOAF \cite{yang_large-scale_2023} \\
\textbullet \quad FishEye8K \cite{gochoo_fisheye8k_2023} & 
\textbullet \quad SMOD \cite{chen_amfd_2024} \\
\textbullet \quad DENSE \cite{bijelic_seeing_2020} &
\textbullet \quad VisDrone \cite{zhu_detection_2022} \\
\textbullet \quad Sim10k \cite{johnson-roberson_driving_2017} & 
\textbullet \quad Synscapes \cite{wrenninge_synscapes_2018} \\
\textbullet \quad TIMo \cite{schneider_timodataset_2022} &
\textbullet \quad DSEC \cite{gehrig_low-latency_2024, gehrig_dsec_2021} \\
\end{tabular}
\end{table}

Due to potential storage location changes, we do not provide direct download links. However, all information can be accessed through the cited papers, respective dataset homepages, and associated GitHub repositories.
Some datasets require access requests, which are typically granted for research purposes.
While the complete raw data across all datasets comprises several terabytes, the benchmark-specific subset is substantially smaller.

\noindent \textbf{2. Processing the images and annotations}
Section~\ref{sec:supp:benchmark-description} provides detailed descriptions of all datasets and their required processing protocols. 
To get the raw annotations and images into the correct format required for D-RICO and EC-RICO, we provide a script for each dataset. For some datasets, it is enough to just merge the classes in the RICO classes; for other datasets, extensive calculations are required.

\noindent \textbf{3. Combining the processed images and annotations into D-RICO and EC-RICO}
We provide a template file for the final D-RICO and EC-RICO annotation file. These files miss the actual annotations but hold all other relevant information like file name and image size. The missing annotations can either be filled in with a custom script or be reproduced using one of the provided scripts.

The D-RICO and EC-RICO benchmark can then be used.
We provide our training and evaluation framework that is based on Detectron2~\cite{wu_detectron2_2019}.
See Section~\ref{supp:sec:il-detectron2} for further details.

\section{More Details on the D-RICO and EC-RICO}\label{sec:supp:benchmark-description}

Section~3 introduces the Domain-RICO and Expanding-Classes RICO benchmarks. In this section, we present additional statistics and comparisons regarding the tasks, as well as a comprehensive explanation of the construction of each task and the data processing and preparation.

\subsection{Additional Statistics}

\subsubsection{Detailed Class Distribution}
Tables~\ref{tab:supp:ann:class-percentages-total-d} and~\ref{tab:supp:ann:class-percentages-total-ec} show the relative values and total label counts for each task, with visual distributions in Figure~2 e) and j).
Intra-task differences vary significantly, and annotations are unevenly distributed. For D-RICO, \textit{thermal fisheye indoor} holds less than 1\% of labels, while \textit{photorealistic simulation} and \textit{drone} each contribute 22\%. In EC-RICO, \textit{fisheye car} has 3\%, whereas \textit{drone} accounts for 33\%.
This variation enhances task diversity, as class distributions and object counts differ across tasks.

\begin{table}
\centering
\scriptsize
\caption{Annotation counts per task and class as percentages, including total dataset contribution. A dash (-) indicates that there are no available annotations for that class in the respective dataset.}
\label{tab:supp:ann:class-percentages-total-d}
\begin{tabular}{l c c c c c}
\toprule
          \textbf{Task} & \textbf{person} & \textbf{vehicle} & \textbf{bicycle} &    \textbf{Total} &  \textbf{Total in \%} \\
\midrule
      daytime  &   37.8 &   57.86 &    4.34 &  28255 &     3.24 \\
          thermal &  35.14 &   59.76 &     5.1 &  58337 &     6.70 \\
     fisheye fix &   8.58 &   91.42 &       - &  99688 &    11.45 \\
      drone &  29.91 &   67.66 &    2.42 & 187425 &    21.52 \\
         simulation &  42.48 &   54.55 &    2.97 &  42698 &     4.90 \\
     fisheye car &  25.15 &   74.85 &       - &  54298 &     6.24 \\
          RGB + thermal fusion &  26.14 &   20.12 &   53.73 &  20686 &     2.38 \\
        video game &      - &   100.0 &       - &  27038 &     3.10 \\
       nighttime &   6.52 &   92.91 &    0.58 &  48139 &     5.53 \\
          fisheye indoor &  100.0 &       - &       - &  21752 &     2.50 \\
    gated &  39.56 &   60.44 &       - &  35050 &     4.02 \\
     photoreal. simulation &   62.5 &    37.5 &       - & 195263 &    22.42 \\
          thermal fisheye indoor &  100.0 &       - &       - &   5747 &     0.66 \\
inclement &  29.34 &   70.66 &       - &  23022 &     2.64 \\
          event camera  &   20.2 &   77.33 &    2.48 &  23418 &     2.69 \\
\bottomrule
\end{tabular}
\end{table}

\begin{table}
\centering
\scriptsize
\setlength{\tabcolsep}{1pt}
\caption{Annotation counts per task and class as percentages, including total dataset contribution. A dash (-) indicates that there are no available annotations for that class in the respective dataset.}
\label{tab:supp:ann:class-percentages-total-ec}
\begin{tabular}{l c c c c c c c c c c}
\toprule
      \textbf{Task} &  \textbf{person} &    \textbf{car} & \textbf{bicycle} & \textbf{\makecell{motor-\\cycle}} & \textbf{truck} &   \textbf{bus} & \textbf{\makecell{traffic\\light}} & \textbf{\makecell{traffic\\sign}} &    \textbf{Total} &  \textbf{Total in \%} \\
\midrule
 fisheye car &  100.00 &      - &       - &          - &     - &     - &             - &            - &  18777 &     3.20 \\
gated &   41.95 &  58.05 &       - &          - &     - &     - &             - &            - &  35963 &     6.13 \\
  daytime &   39.62 &  40.74 &   19.64 &          - &     - &     - &             - &            - &  27461 &     4.68 \\
 fisheye fix  &    8.91 &  33.11 &       - &      57.97 &     - &     - &             - &            - & 102594 &    17.48 \\
     simulation &   43.95 &  39.82 &     3.1 &       4.31 &  8.82 &     - &             - &            - &  44100 &     7.51 \\
  drone &   27.65 &  53.93 &    2.47 &       9.47 &  4.01 &  2.47 &             - &            - & 196355 &    33.45 \\
      thermal &   33.15 &  39.41 &    4.52 &       0.87 &  0.54 &  1.24 &         20.28 &            - &  79521 &    13.55 \\
   nighttime &    4.31 &  56.22 &    0.28 &       0.15 &  1.18 &  0.47 &         17.82 &        19.57 &  82285 &    14.02 \\
\bottomrule
\end{tabular}
\end{table}

\subsubsection{t-SNE and Nearest Mean Classifier of Image Features}
We project extracted image features into a two-dimensional space using t-SNE to analyze the differences between tasks in image space. The features are obtained from a model trained following the setup in Section~4.1.

\begin{figure}[t]
    \centering
    \begin{subfigure}{1\linewidth}
        \centering
        \includegraphics[width=\linewidth]{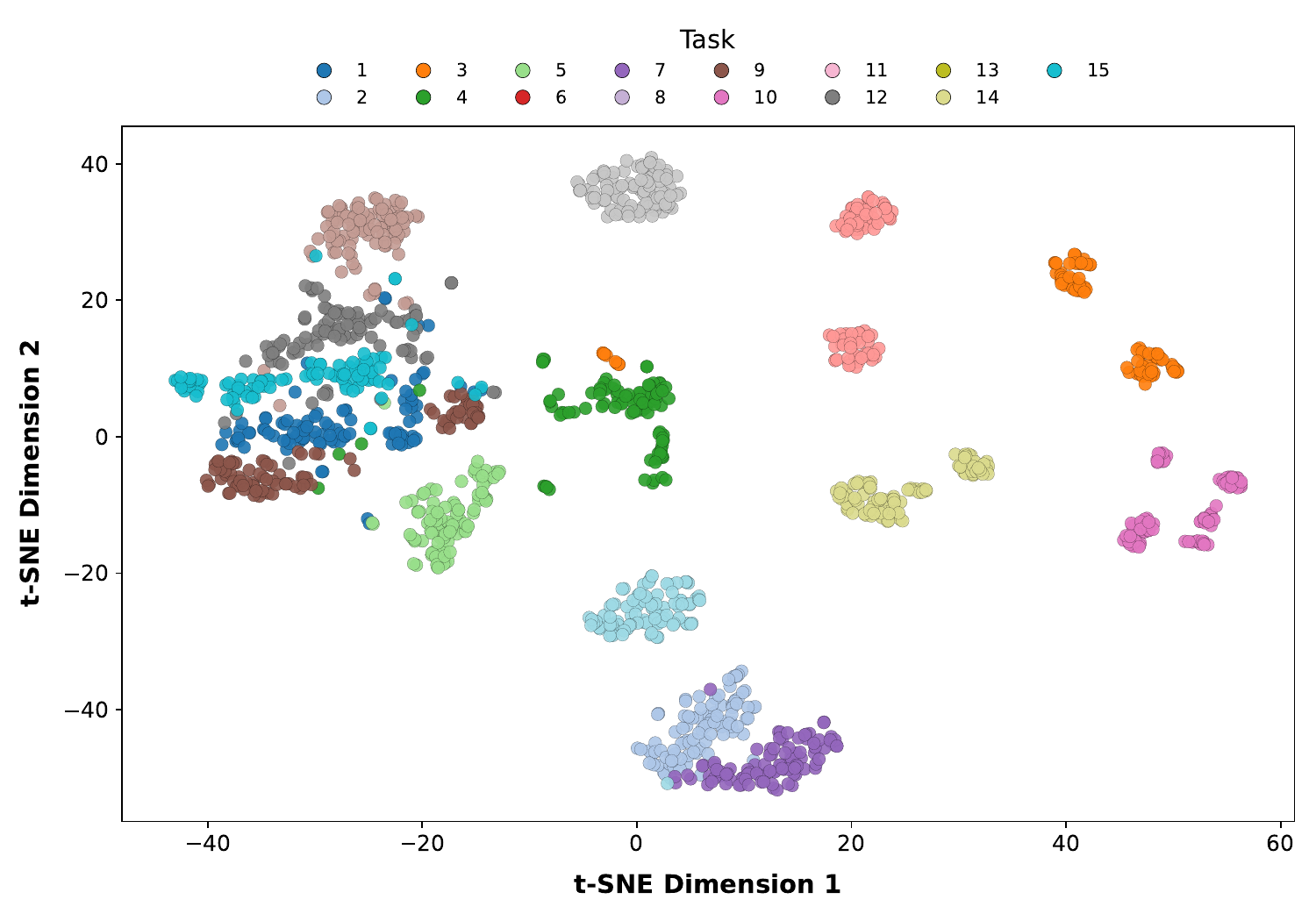}
        \caption{D-RICO}
        \label{fig:tsne_dcl}
    \end{subfigure}
    \begin{subfigure}{1\linewidth}
        \centering
        \includegraphics[width=\linewidth]{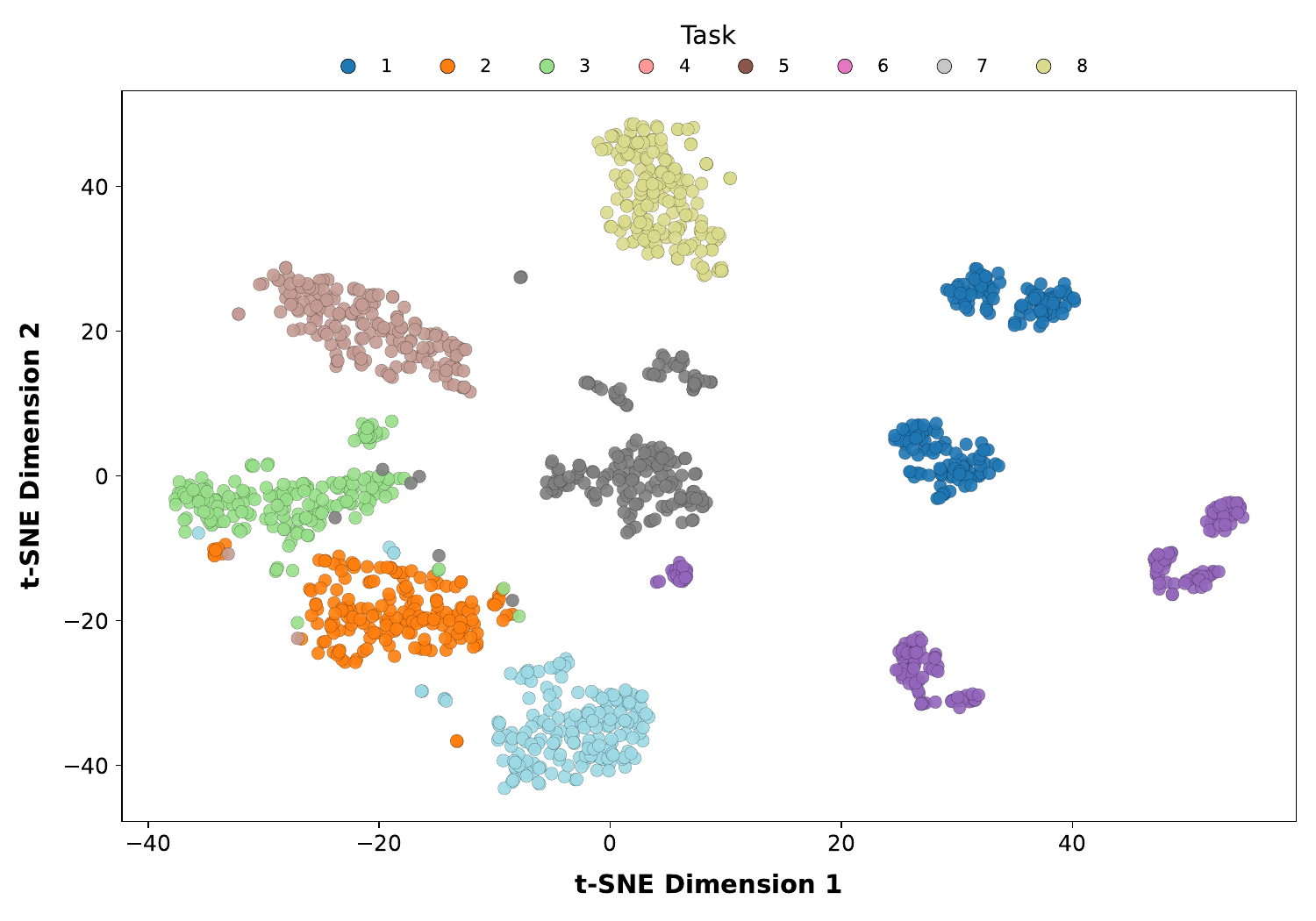}
        \caption{EC-RICO}
        \label{fig:tsne_ccl}
    \end{subfigure}
    \caption{t-SNE features}
    \label{fig:supp:tsne_combined}
\end{figure}

Figure~\ref{fig:supp:tsne_combined} presents the t-SNE visualization. The image features are well-separated for EC-RICO, whereas in D-RICO, some tasks are closer while remaining distinguishable. This indicates that tasks differ in terms of image representations.

\begin{figure}[t]
    \centering
    \includegraphics[width=\linewidth]{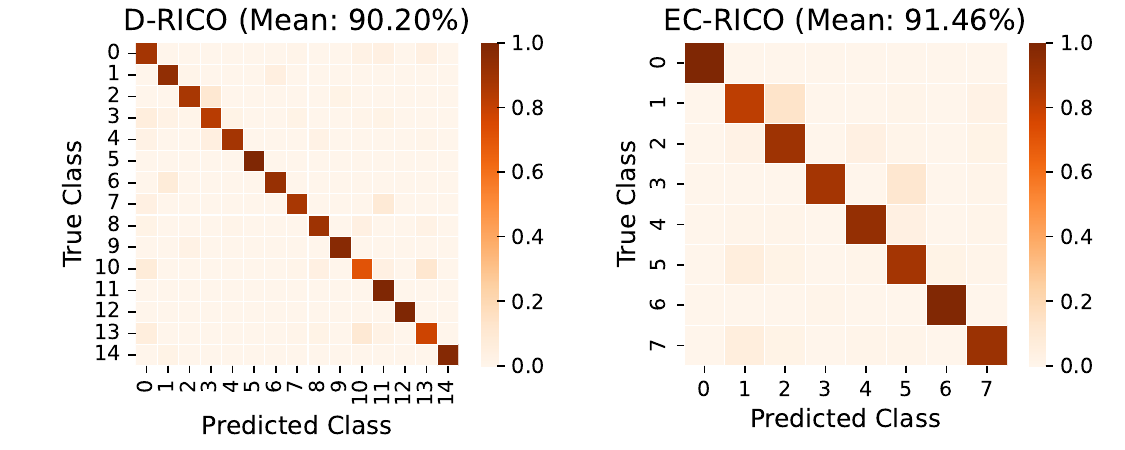}
    \caption{Confusion Matrix of the Nearest Mean Classifier based on image features.}
    \label{fig:supp:confusion:nmc}
\end{figure}

Furthermore, a simple nearest mean classifier (as used in LDB~\cite{song_non-exemplar_2024}) can already achieve good task classification based on these features. Figure~\ref{fig:supp:confusion:nmc} shows the corresponding confusion matrix, reinforcing the distinctiveness of tasks in feature space.

\subsubsection{Image sizes}
The input to the network is always $1536 \times 1536 \times 3$, while the original dataset images vary in size, as shown in Table~\ref{tab:supp:image-sizes}. Except for \textit{fisheye fix} and \textit{drone}, all tasks maintain a constant intra-task resolution. The \textit{thermal}, \textit{gated}, and \textit{thermal fisheye indoor} tasks contain grayscale images, whereas all others are RGB.

Since none of the original image sizes match $1536 \times 1536$, all images are padded, resized, or cropped accordingly. Grayscale images are duplicated across channels to match the required three-channel format. This highlights the benchmark's diversity, in contrast to others that typically maintain uniform image resolutions.

\begin{table}
\centering
\scriptsize
\caption{Image sizes for each task (mean $\pm$ standard deviation).}
\label{tab:supp:image-sizes}
\begin{tabular}{l c c }
\toprule
          \textbf{Task} & \textbf{Width $\times$ Height}  \\
\midrule
      daytime  &   $1600{\scriptstyle\pm 0} \times 900{\scriptstyle\pm 0} \times 3$ \\
          thermal &  $640{\scriptstyle\pm 0} \times 512{\scriptstyle\pm 0} \times 1$ \\
     fisheye fix &   $1404{\scriptstyle\pm 312} \times 1232{\scriptstyle\pm 310} \times 3$  \\
      drone &  $1484{\scriptstyle\pm 229} \times 951{\scriptstyle\pm 208} \times 3$ \\
         simulation &  $1280{\scriptstyle\pm 0} \times 800{\scriptstyle\pm 0} \times 3$\\
     fisheye car &  $1280{\scriptstyle\pm 0} \times 966{\scriptstyle\pm 0} \times 3$\\
          RGB + thermal fusion & $640{\scriptstyle\pm 0} \times 512{\scriptstyle\pm 0} \times 3$ \\
        video game &  $1914{\scriptstyle\pm 0} \times 1052{\scriptstyle\pm 0} \times 3$  \\
       nighttime &   $1280{\scriptstyle\pm 0} \times 720{\scriptstyle\pm 0} \times 3$\\
          fisheye indoor &  $1024{\scriptstyle\pm 0} \times 1024{\scriptstyle\pm 0} \times 3$  \\
    gated &  $1280{\scriptstyle\pm 0} \times 720{\scriptstyle\pm 0} \times 1$  \\
     photoreal. simulation &  $1440{\scriptstyle\pm 0} \times 720{\scriptstyle\pm 0} \times 3$ \\
          thermal fisheye indoor &  $512{\scriptstyle\pm 0} \times 512{\scriptstyle\pm 0} \times 1$ \\
inclement &  $1920{\scriptstyle\pm 0} \times 1024{\scriptstyle\pm 0} \times 3$ \\
          event camera  &   $640{\scriptstyle\pm 0} \times 480{\scriptstyle\pm 0} \times 3$ \\
\bottomrule
\end{tabular}
\end{table}

\subsubsection{Labeling Policy}
Each dataset follows a specific labeling policy that defines how objects in an image are labeled.
In Section~\ref{supp:sec:detailed-task-description} the labeling policies and how we processed them are stated in more detail. Additional details can be found in the respective publications.

\begin{figure}
    \centering
    \includegraphics[width=1\linewidth]{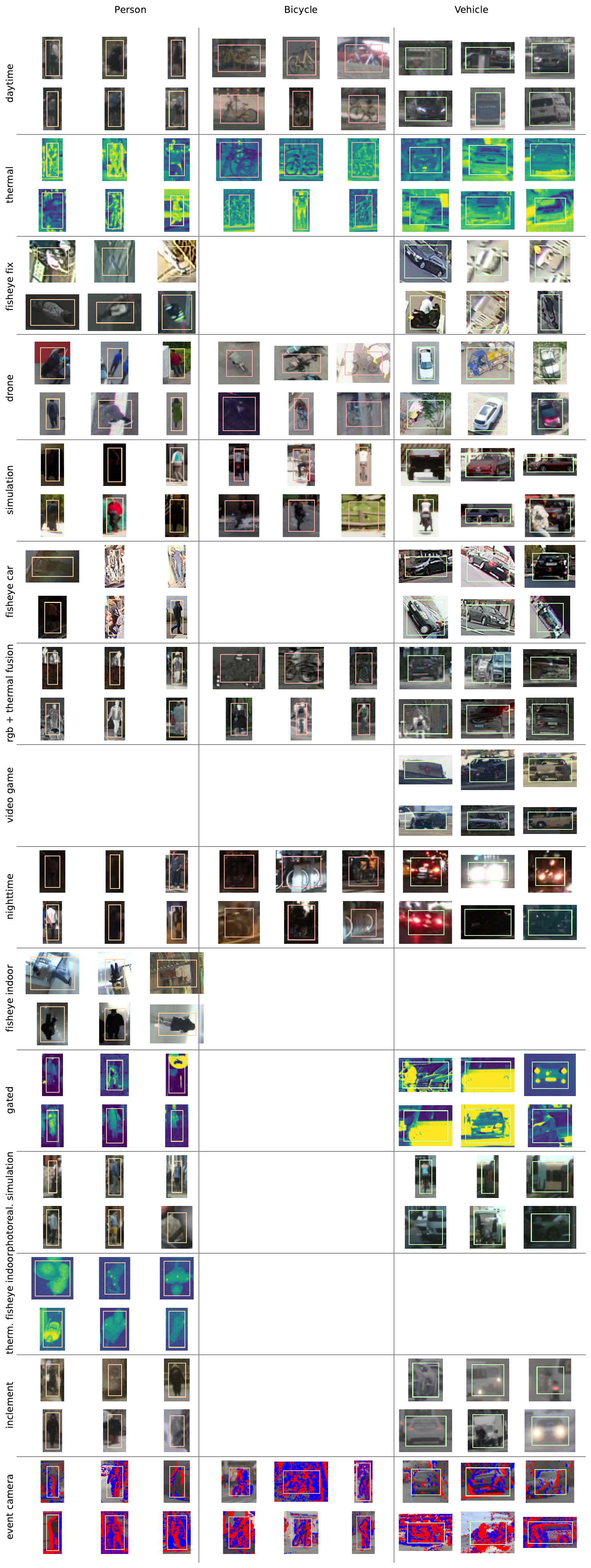}
    \caption{Random example objects to visualize the D-RICO benchmark's different annotation policies and qualities.}
    \label{fig:supp:label-policy-d-rico}
\end{figure}

\begin{figure*}
    \centering
    \includegraphics[width=1\linewidth]{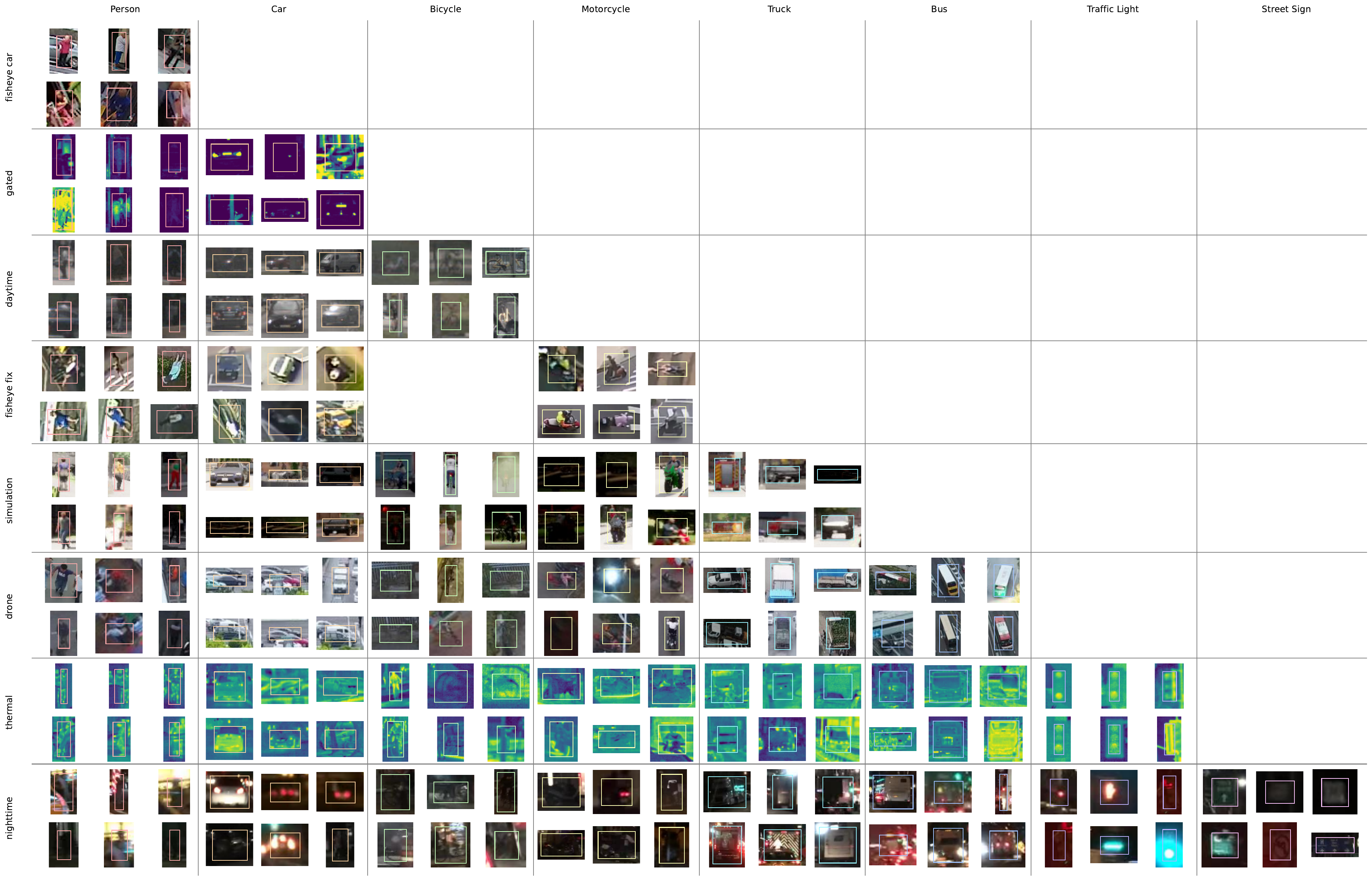}
    \caption{Random examples of objects to visualize the various annotation policies and qualities for the EC-RICO benchmark.}
    \label{fig:supp:label-policy-ec-rico}
\end{figure*}

Figures~\ref{fig:supp:label-policy-d-rico} and~\ref{fig:supp:label-policy-ec-rico} illustrate randomly sampled objects for each class of D-RICO and EC-RICO.
The visualizations highlight substantial differences in labeling policies and annotation quality.
The differences concern tight vs. loose bounding boxes, amodal vs. visible bounding boxes, background objects, small objects, groups of objects, and class definitions.
In addition to these differences, there are also differences in terms of objects that are not being labeled.

\subsection{Detailed Task and Dataset Description}\label{supp:sec:detailed-task-description}

\subsubsection{\texorpdfstring{Daytime (nuImages~\cite{caesar_nuscenes_2020})}{Daytime (nuImages)}}\label{supp:tasks:nuimages}
NuImages~\cite{caesar_nuscenes_2020} is a large-scale dataset derived from nuScenes that contains 93,000 annotated images captured in Boston and Singapore under various weather conditions and times of day. The dataset encompasses diverse urban driving scenarios, emphasizing autonomous perception tasks such as object detection and scene understanding.

\begin{figure}[t]
  \centering
  \begin{tabular}{cc}
    \includegraphics[width=0.45\linewidth]{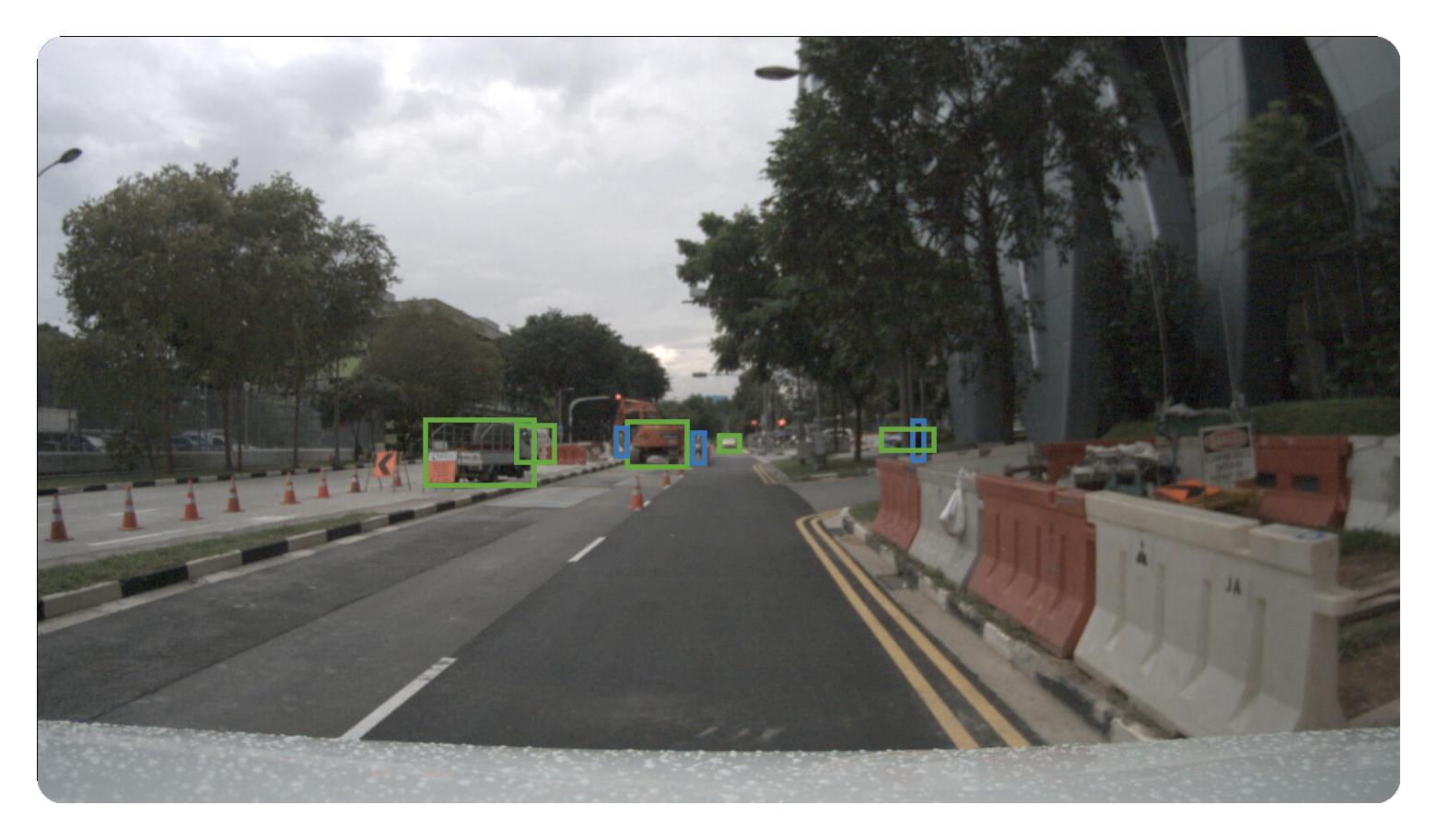} &
    \includegraphics[width=0.45\linewidth]{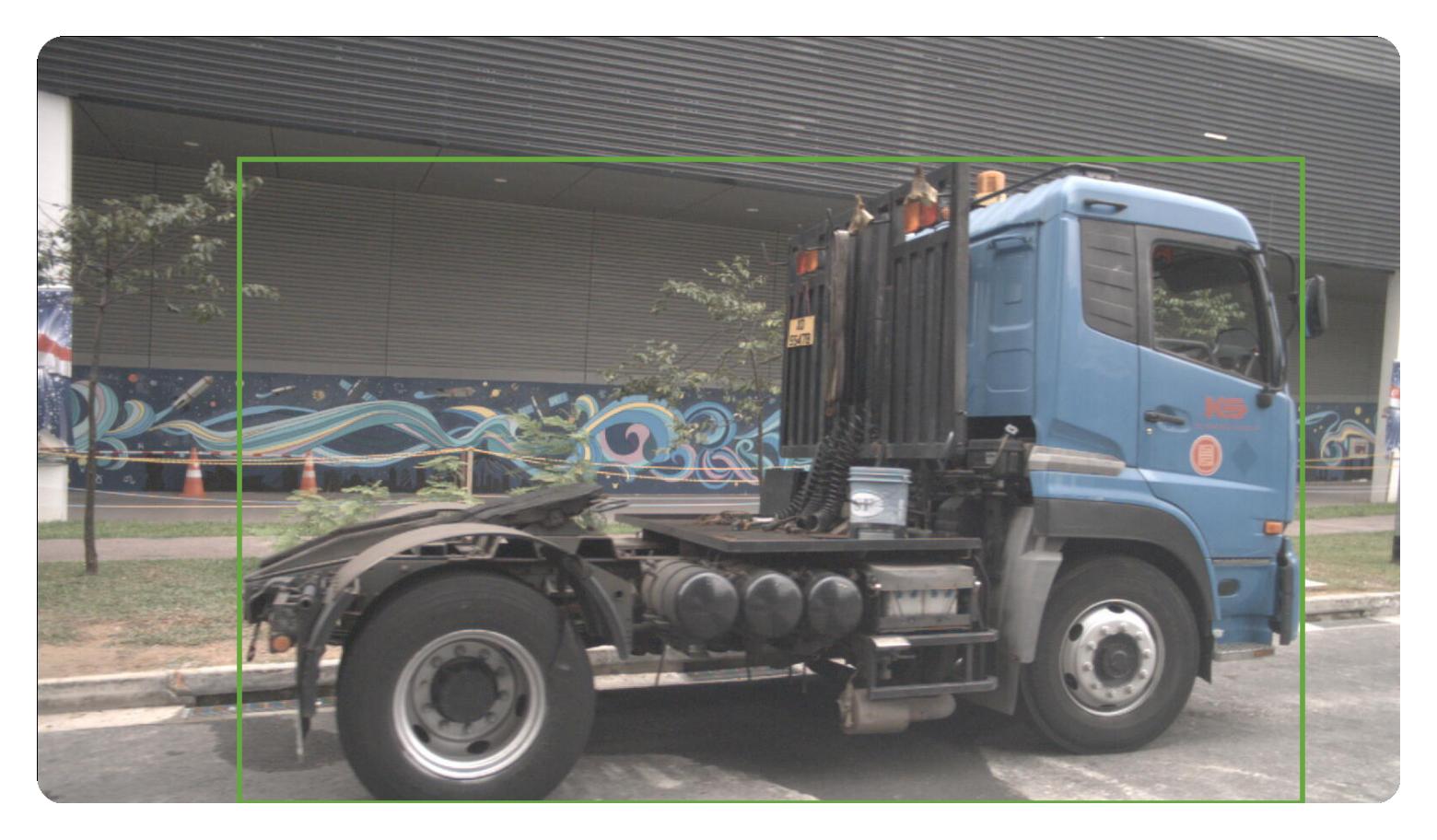} \\
    \includegraphics[width=0.45\linewidth]{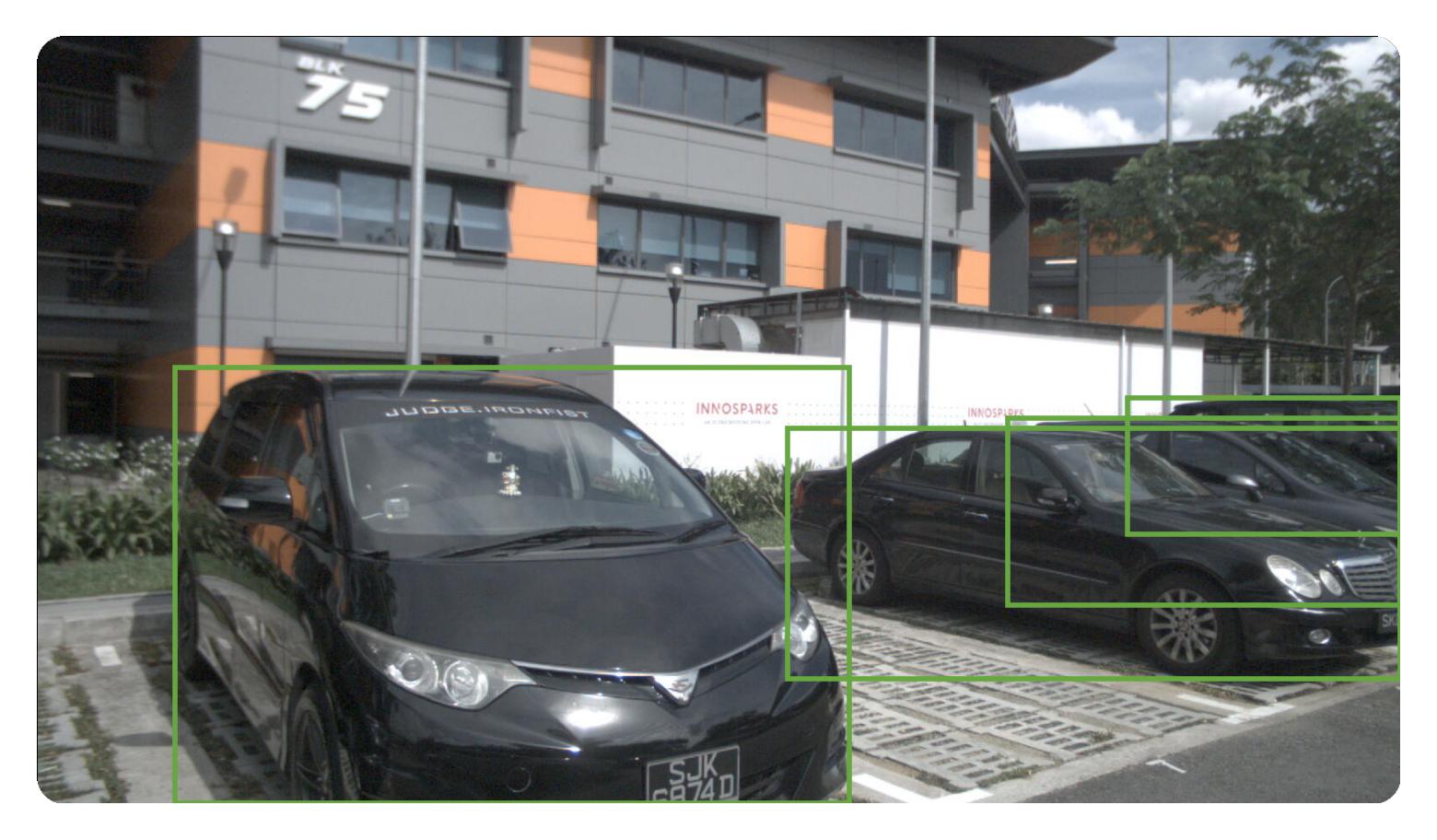} &
    \includegraphics[width=0.45\linewidth]{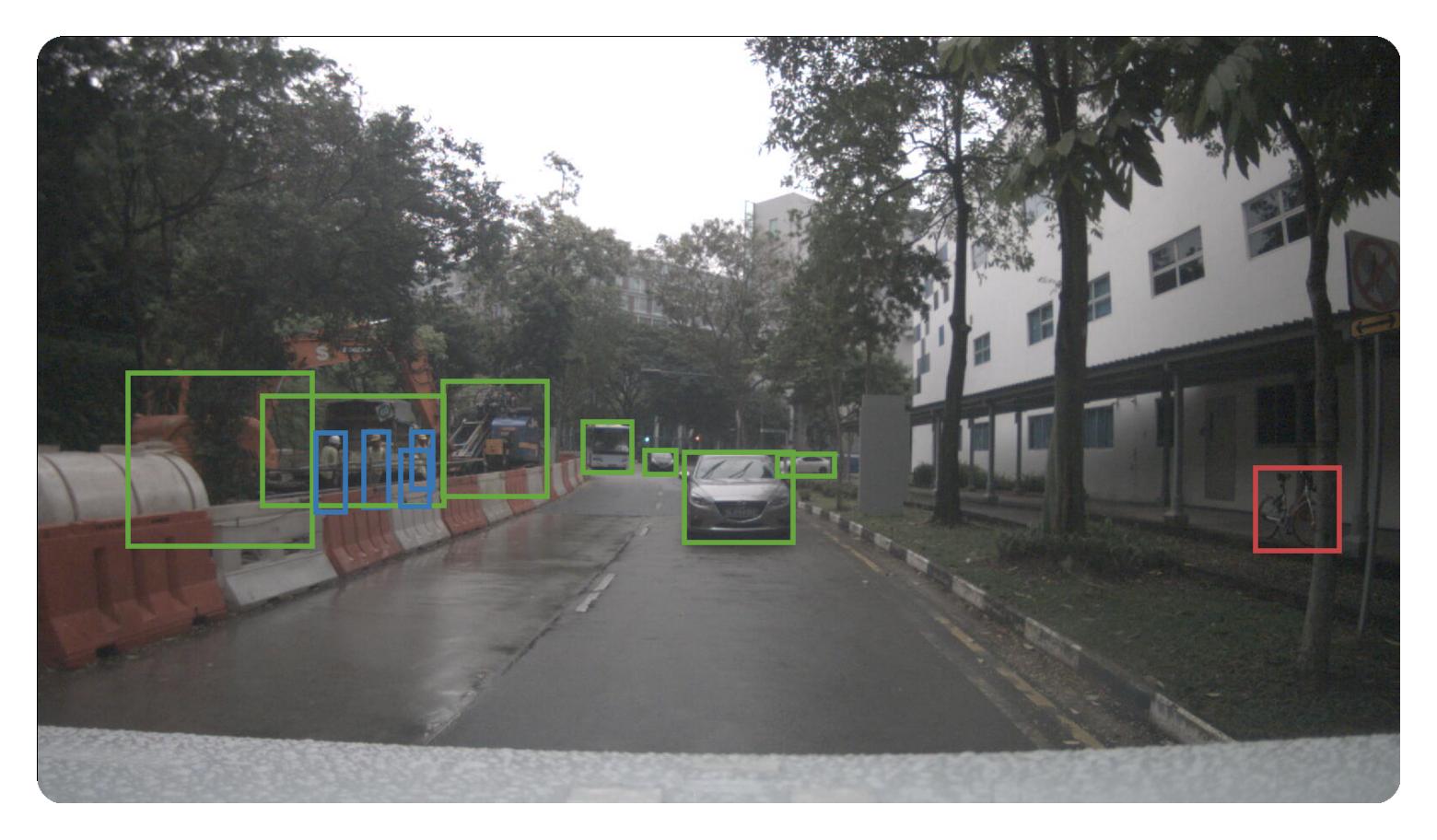} \\
  \end{tabular}
  \caption{Example images of the daytime task with labels for D-RICO.}
  \label{fig:supp:daytime-task}
\end{figure}

\noindent\textbf{Dataset Processing.}
We chose daytime images exclusively from the Singapore One-North district to make the task more distinct.
We exclude images with \textit{vehicle.bus.bendy} instances due to inconsistent annotation protocols wherein bus segments are individually labeled, contrary to standard buses' unified labeling.
Merging these segmented annotations poses difficulties, especially in multi-bus scenes.
Additionally, we exclude images featuring the class \textit{static\_object.bicycle\_rack} as it contradicts with the \textit{vehicle.bicycle} class, as the bicycles in the rack are not labeled to the bicycle class.
Examples are given in Figure~\ref{fig:supp:daytime-task}.

\noindent\textbf{Train, Val, and Test Split.}
To prevent data leakage, we enforce scene-level integrity by ensuring all images from the same scene ID remain in a single partition. We employ a stochastic optimization approach to achieve the target distribution of 60\% train, 10\% validation, and 30\% test. We repeatedly generate random scene-to-partition assignments and evaluate how closely they match the desired proportions. Among these, we select the configuration that minimizes deviation while preserving scene consistency. This Monte Carlo-based strategy ensures a balanced partitioning while maintaining contextual coherence within each split.

\noindent\textbf{Labeling Policy}
We use this dataset's labeling policy as a reference for the other datasets and compare them to this one. Objects are labeled tightly, and their bounding boxes cover only the visible parts. Bicycles and motorcycles include the rider within the bounding box. Small visible objects in the background are labeled.

\noindent\textbf{D-RICO Classes.}

\begin{itemize} \setlength{\itemindent}{0.5cm}
    \item \textbf{person:}
        \begin{itemize} \setlength{\itemindent}{1cm}
            \item \textit{human.pedestrian.adult}
            \item \textit{human.pedestrian.child}
            \item \textit{human.pedestrian.construction\_worker}
            \item \textit{human.pedestrian.personal\_mobility}
            \item \textit{human.pedestrian.police\_officer}
        \end{itemize}

    \item \textbf{bicycle:} \textit{vehicle.bicycle}

    \item \textbf{vehicle:}
        \begin{itemize} \setlength{\itemindent}{1cm}
            \item \textit{vehicle.bus.rigid}
            \item \textit{vehicle.car}
            \item \textit{vehicle.construction}
            \item \textit{vehicle.emergency.ambulance}
            \item \textit{vehicle.emergency.police}
            \item \textit{vehicle.motorcycle}
            \item \textit{vehicle.trailer}
            \item \textit{vehicle.truck}
        \end{itemize}
\end{itemize}

\noindent\textbf{EC-RICO Classes.}
\begin{itemize}\setlength{\itemindent}{0.5cm}

    \item \textbf{person:}
        \begin{itemize} \setlength{\itemindent}{1cm}
            \item \textit{human.pedestrian.adult}
            \item \textit{human.pedestrian.child}
            \item \textit{human.pedestrian.construction\_worker}
            \item \textit{human.pedestrian.personal\_mobility}
            \item \textit{human.pedestrian.police\_officer}
        \end{itemize}
        
    \item \textbf{Car:} \textit{vehicle.car}
        
    \item \textbf{Bicycle:}  \textit{vehicle.bicycle}

\end{itemize}


\subsubsection{\texorpdfstring{Thermal (Teledyne FLIR~\cite{teledyne_flir_free_2024})}{Thermal (Teledyne FLIR)}}\label{sec:supp:thermal-task}
The Teledyne FLIR~\cite{teledyne_flir_free_2024} dataset consists of 26,000 thermal images collected in various urban environments, covering a range of lighting and weather conditions. The dataset is used for thermal-based object detection and multispectral perception in autonomous driving applications.

\begin{figure}[t]
  \centering
  \begin{tabular}{cc}
    \includegraphics[width=0.45\linewidth]{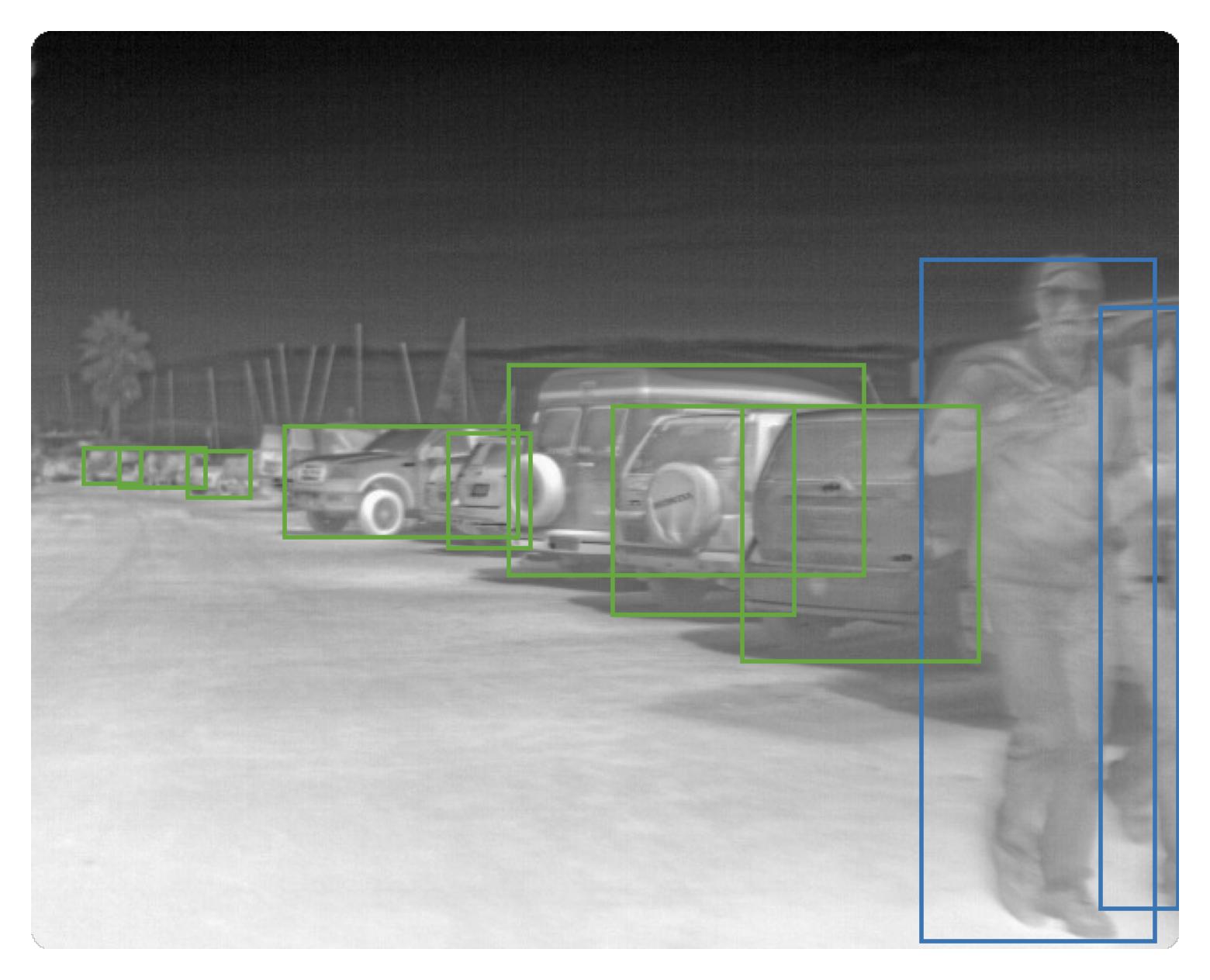} &
    \includegraphics[width=0.45\linewidth]{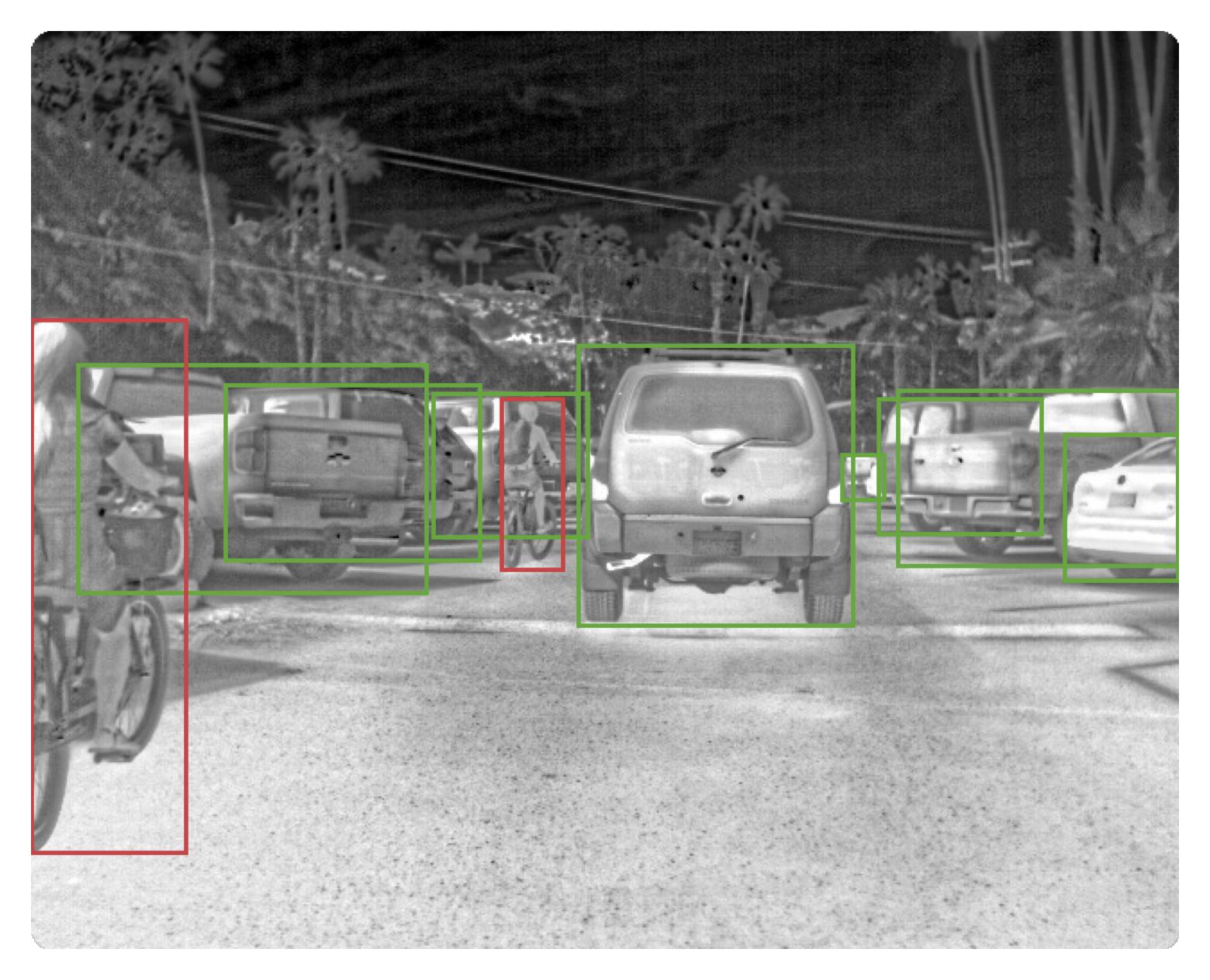} \\
    \includegraphics[width=0.45\linewidth]{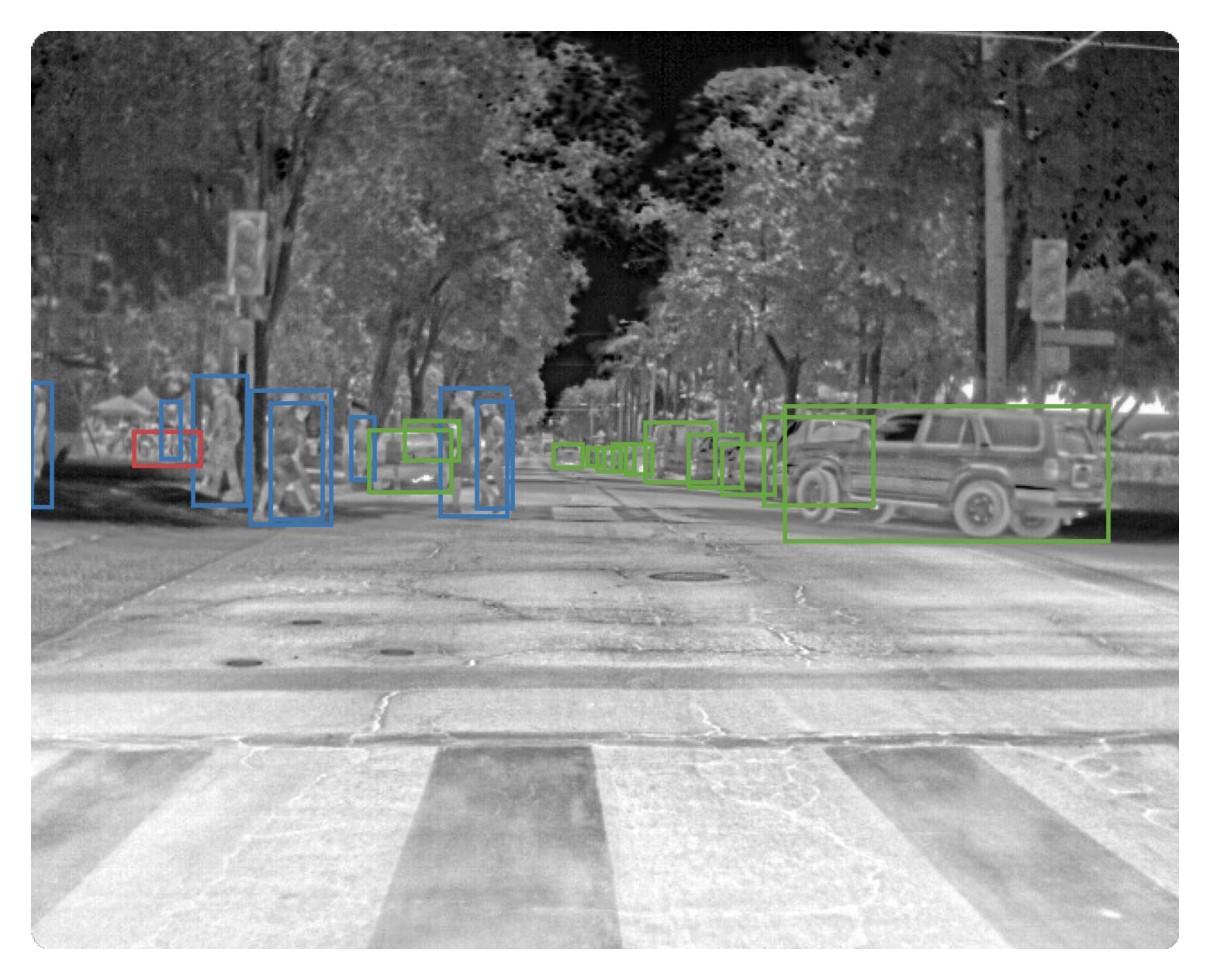} &
    \includegraphics[width=0.45\linewidth]{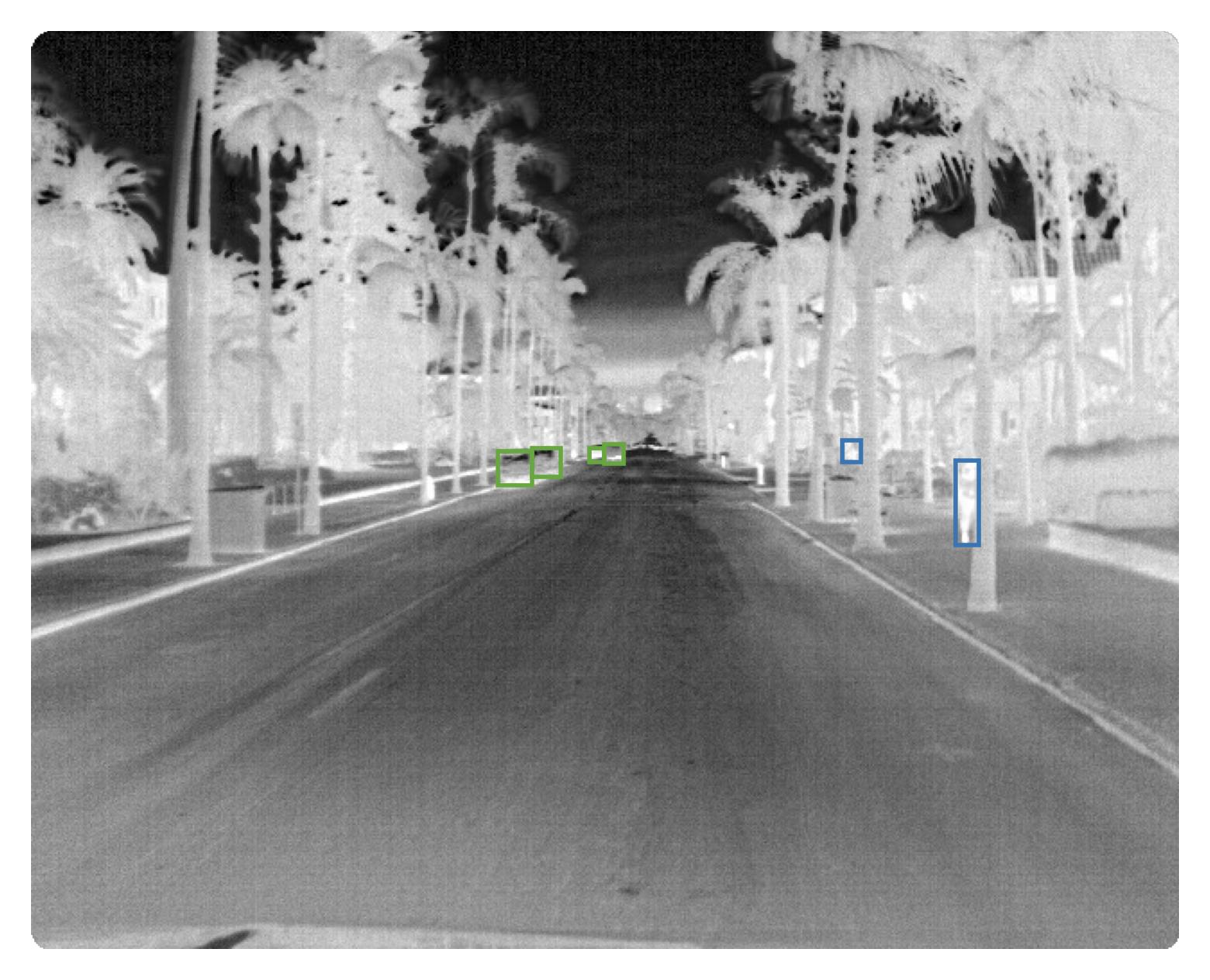} \\
  \end{tabular}
  \caption{Example images of the thermal task.}
  \label{fig:thermal-task}
\end{figure}

\noindent\textbf{Dataset Processing.}
The dataset labels the rider and the bicycle separately, as we define the bicycle class as the bicycle with the rider, we need to merge the two.
The rider is labeled as \textit{person}, so we calculate the intersection over union (IoU) of all person and bicycle bounding boxes. If the IoU exceeds 25\%, the two boxes are merged and labeled bicycle.
This is not always perfect; it also merges if someone is standing next to a bicycle.
See Figure~\ref{fig:thermal-task} for four example images.

\noindent\textbf{Train, Val, and Test Split.}
The dataset provides a scene ID from which the train, validation, and test split is created like described in Section~\ref{supp:tasks:nuimages}.

\noindent\textbf{Labeling Policy}
The labeling policy is close to that of nuImages (see Section~\ref{supp:tasks:nuimages})

\noindent\textbf{D-RICO Classes.}

\begin{itemize} \setlength{\itemindent}{0.5cm}
    \item \textbf{person:} \textit{person}
    \item \textbf{bicycle:} \textit{bicycle}
    \item \textbf{vehicle:} \textit{car}, \textit{truck}, \textit{bus}
\end{itemize}

\noindent\textbf{EC-RICO Classes.}
The naming of the classes in the dataset matches the naming of EC-RICO, hence, we select for the EC-RICO classes the following labels: \textit{person}, \textit{car}, \textit{bicycle}, \textit{motorcycle}, \textit{truck}, \textit{bus}, and \textit{traffic light}.
All other available labels are not used.


\subsubsection{\texorpdfstring{Fisheye Fix (FishEye8K~\cite{gochoo_fisheye8k_2023})}{Fisheye Fix (FishEye8K)}}
FishEye8K~\cite{gochoo_fisheye8k_2023} is a dataset comprising 8,000 fisheye images collected from 18 surveillance cameras at road intersections in Taiwan. It features over 157,000 annotated objects across five categories, capturing a range of lighting conditions and object scales to support fisheye-based perception research.

\begin{figure}[t]
  \centering
  \begin{tabular}{cc}
    \includegraphics[width=0.45\linewidth]{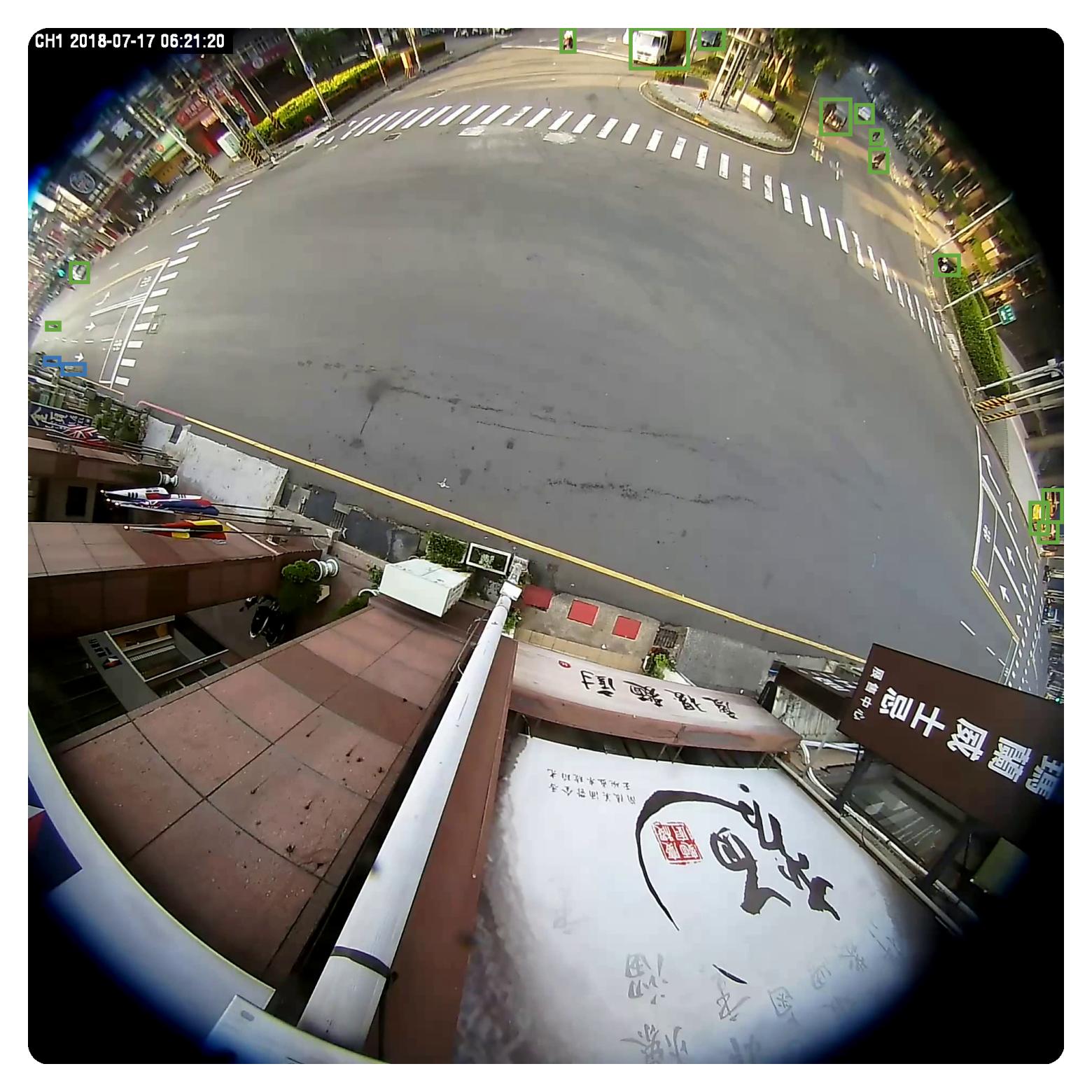} &
    \includegraphics[width=0.45\linewidth]{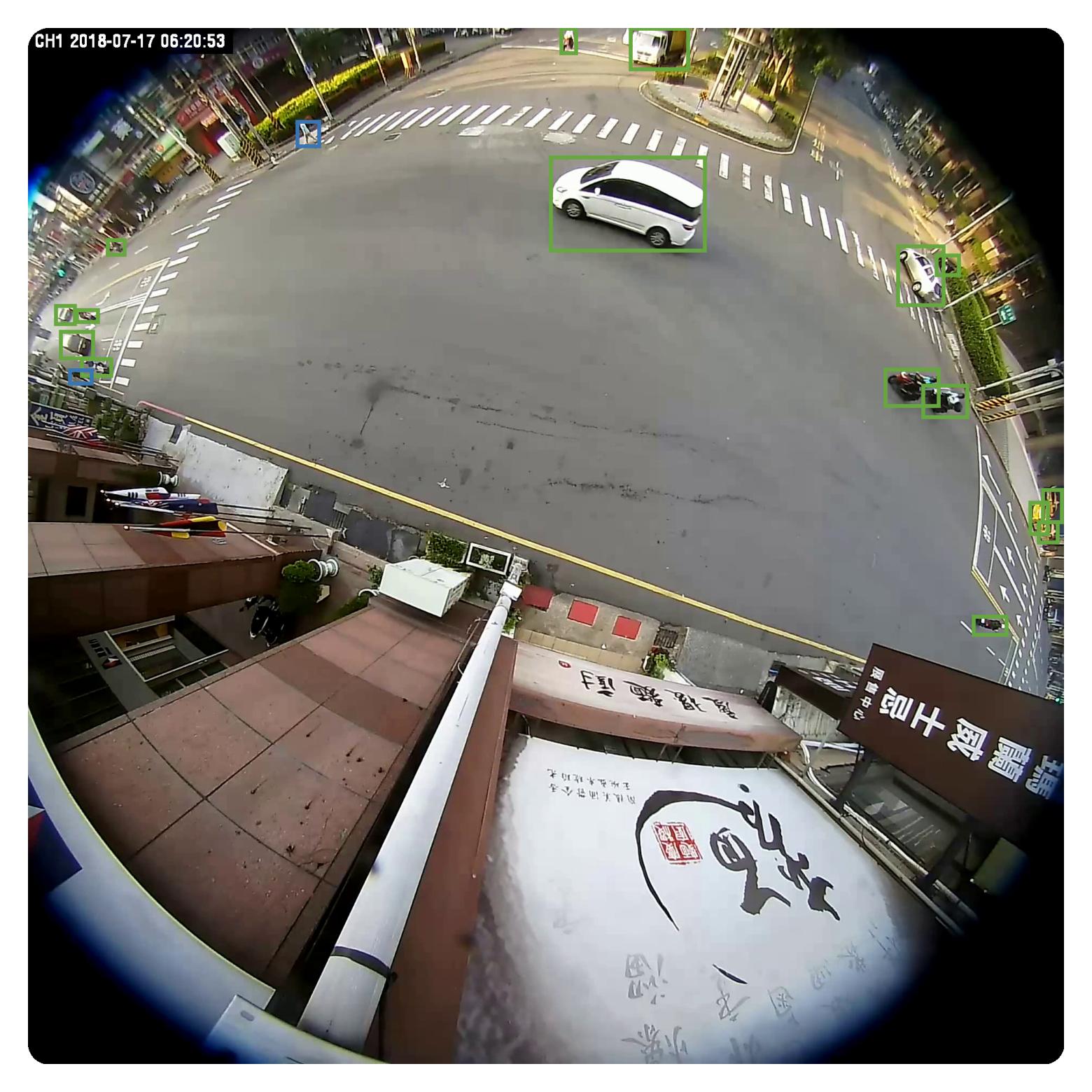} \\
    \includegraphics[width=0.45\linewidth]{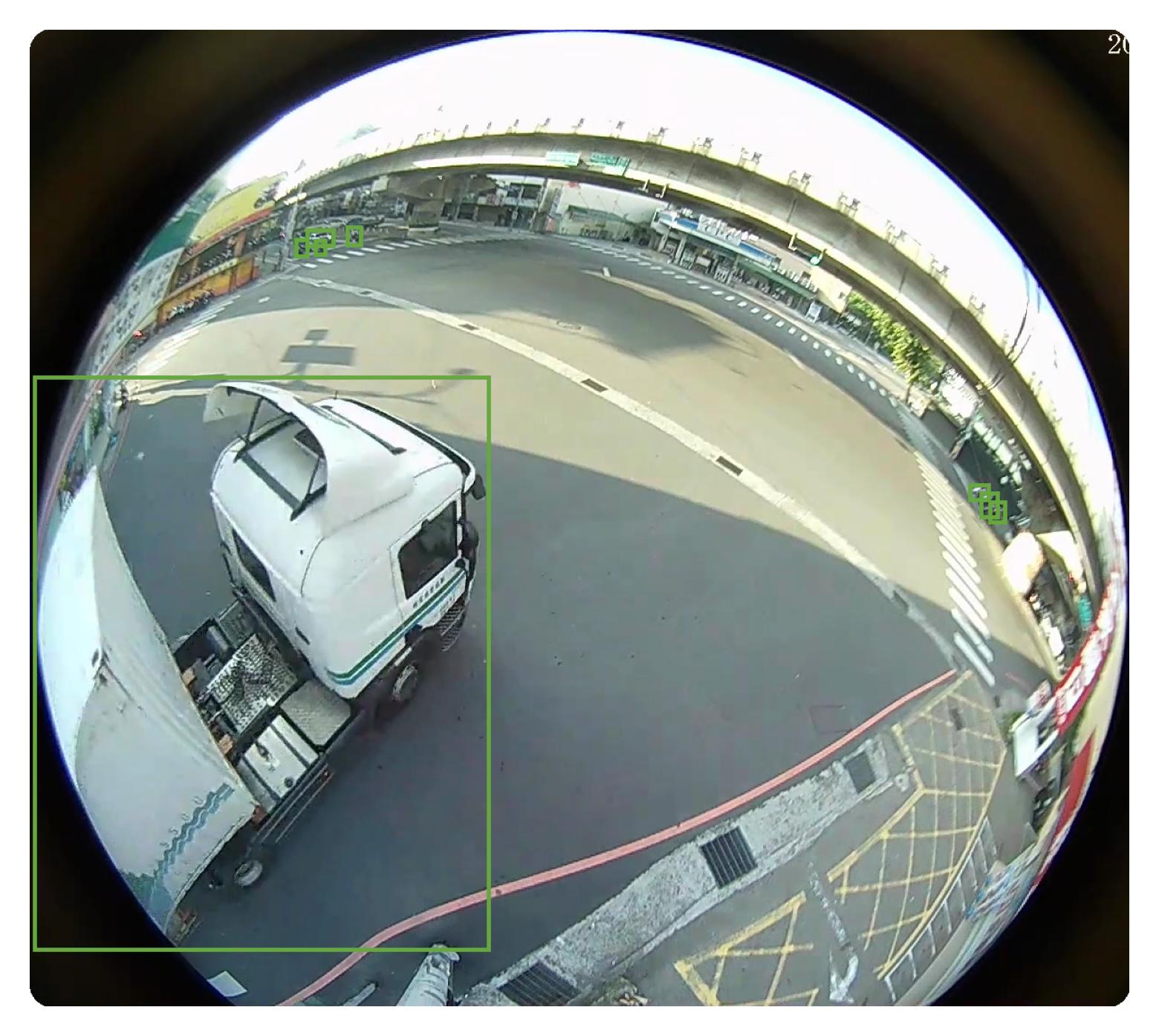} &
    \includegraphics[width=0.45\linewidth]{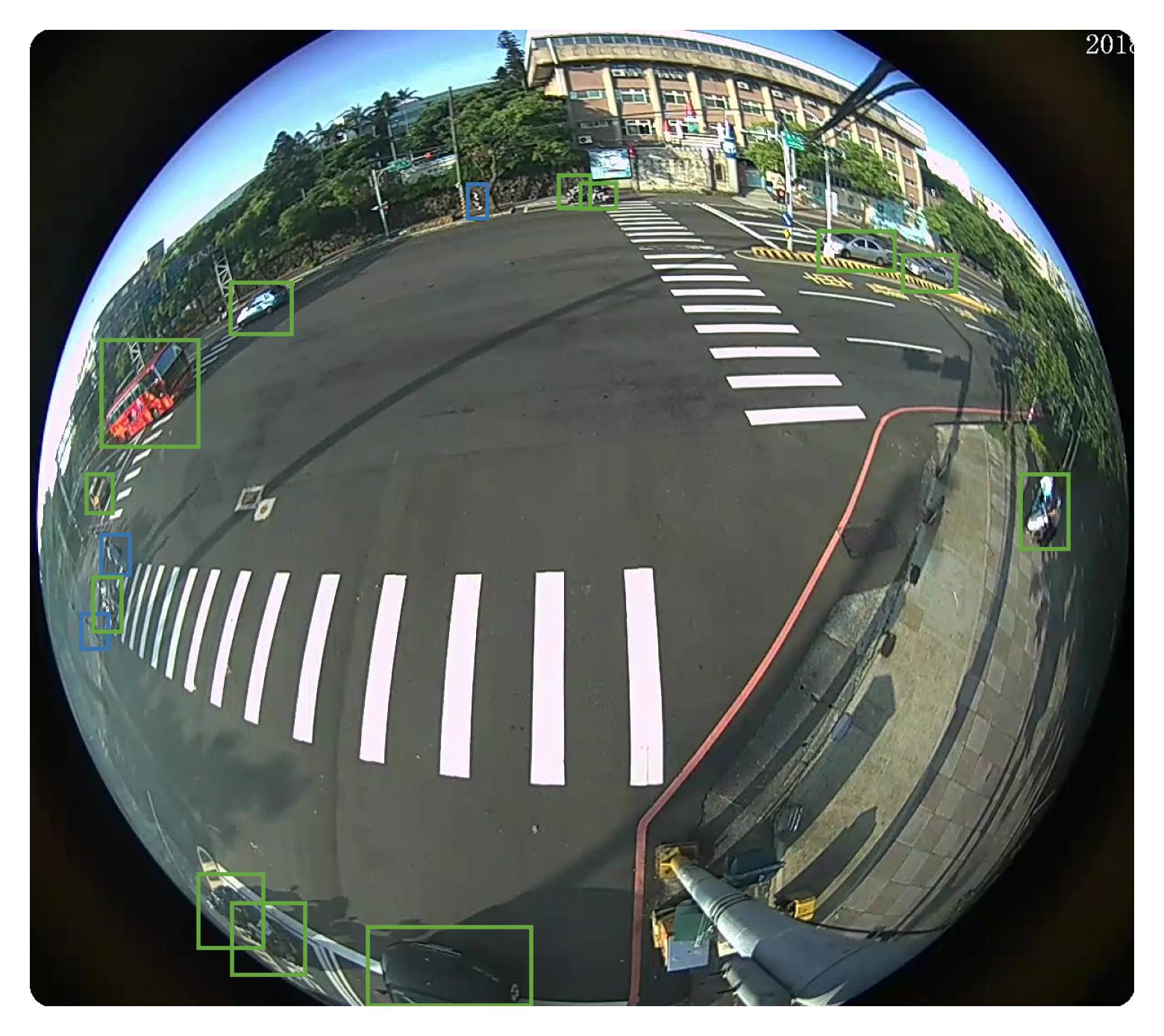} \\
  \end{tabular}
  \caption{Example images of the fisheye fix task.}
  \label{fig:fisheye-fix-task}
\end{figure}

\noindent\textbf{Dataset Processing.}
No further processing is required for this dataset.
Examples are shown in Figure~\ref{fig:fisheye-fix-task}.

\noindent\textbf{Train, Val, and Test Split.}
The dataset provides a scene ID from which the train, validation, and test split is created like described in Section~\ref{supp:tasks:nuimages}.

\noindent\textbf{Labeling Policy}
Compared to the reference labeling policy (see Section~\ref{supp:tasks:nuimages}), the \textit{bike} class consists of motorcycles and bicycles.
However, as the number of motorcycles in this dataset is much greater than that of bicycles, we decided to label all as vehicles.

\noindent\textbf{D-RICO Classes.}

\begin{itemize} \setlength{\itemindent}{0.5cm}
    \item \textbf{person:} \textit{Pedestrian}

    \item \textbf{vehicle:} \textit{Car}, \textit{Bus}, \textit{Truck}, \textit{Bike}
\end{itemize}

\noindent\textbf{EC-RICO Classes.}

\begin{itemize} \setlength{\itemindent}{0.5cm}
    \item \textbf{person:} \textit{Pedestrian}

    \item \textbf{car:} \textit{Car}

    \item \textbf{motorcycle:} \textit{Bike}
\end{itemize}


\subsubsection{\texorpdfstring{Drone (VisDrone~\cite{zhu_detection_2022})}{Drone (VisDrone)}}
VisDrone~\cite{zhu_detection_2022} is a large-scale dataset comprising over 10,000 images and 263 video clips captured by drones in 14 cities across China. It includes dense urban and suburban scenes with millions of object annotations, facilitating aerial-based object detection and tracking research.

\begin{figure}[t]
  \centering
  \begin{tabular}{cc}
    \includegraphics[width=0.45\linewidth]{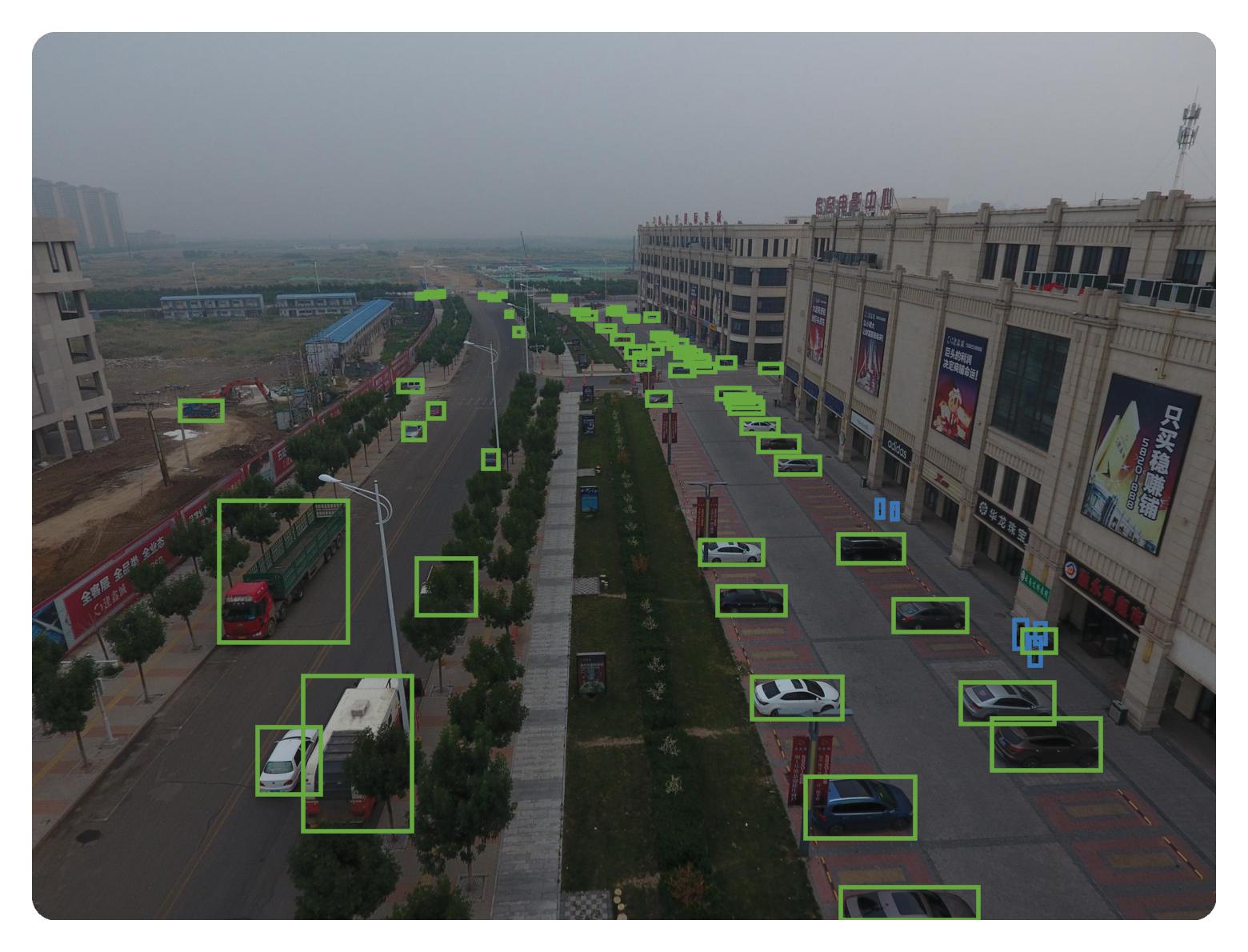} &
    \includegraphics[width=0.45\linewidth]{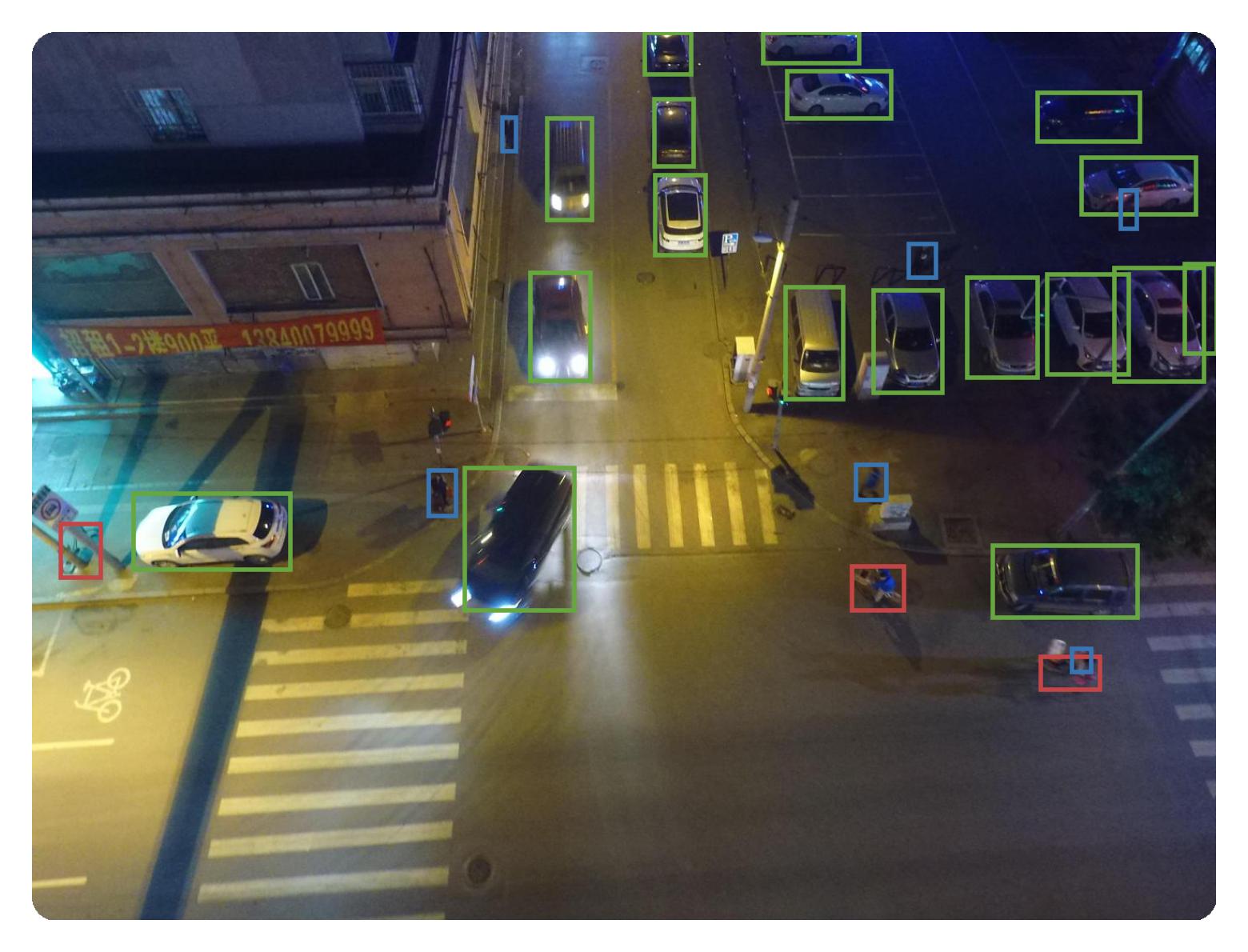} \\
    \includegraphics[width=0.45\linewidth]{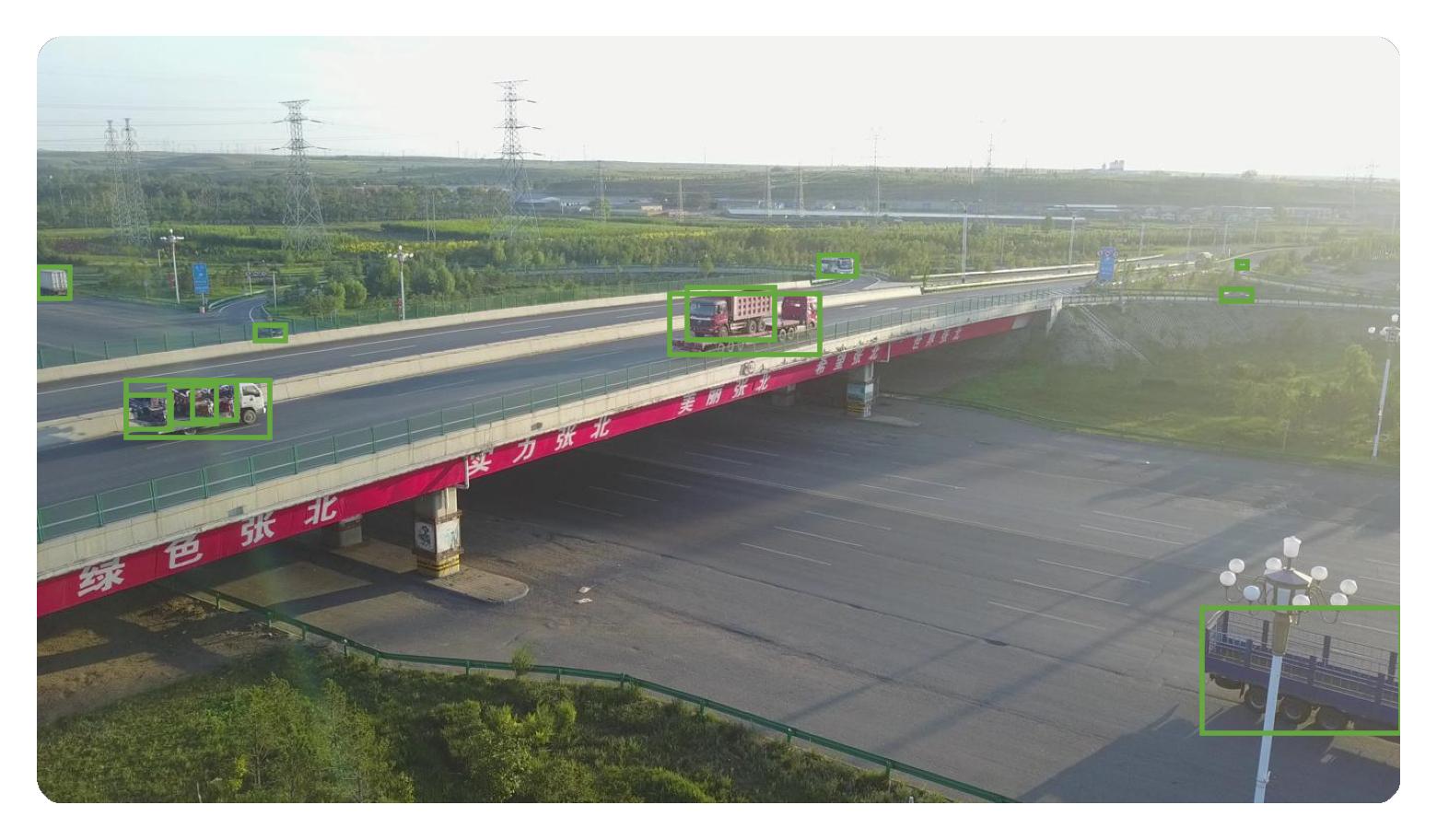} &
    \includegraphics[width=0.45\linewidth]{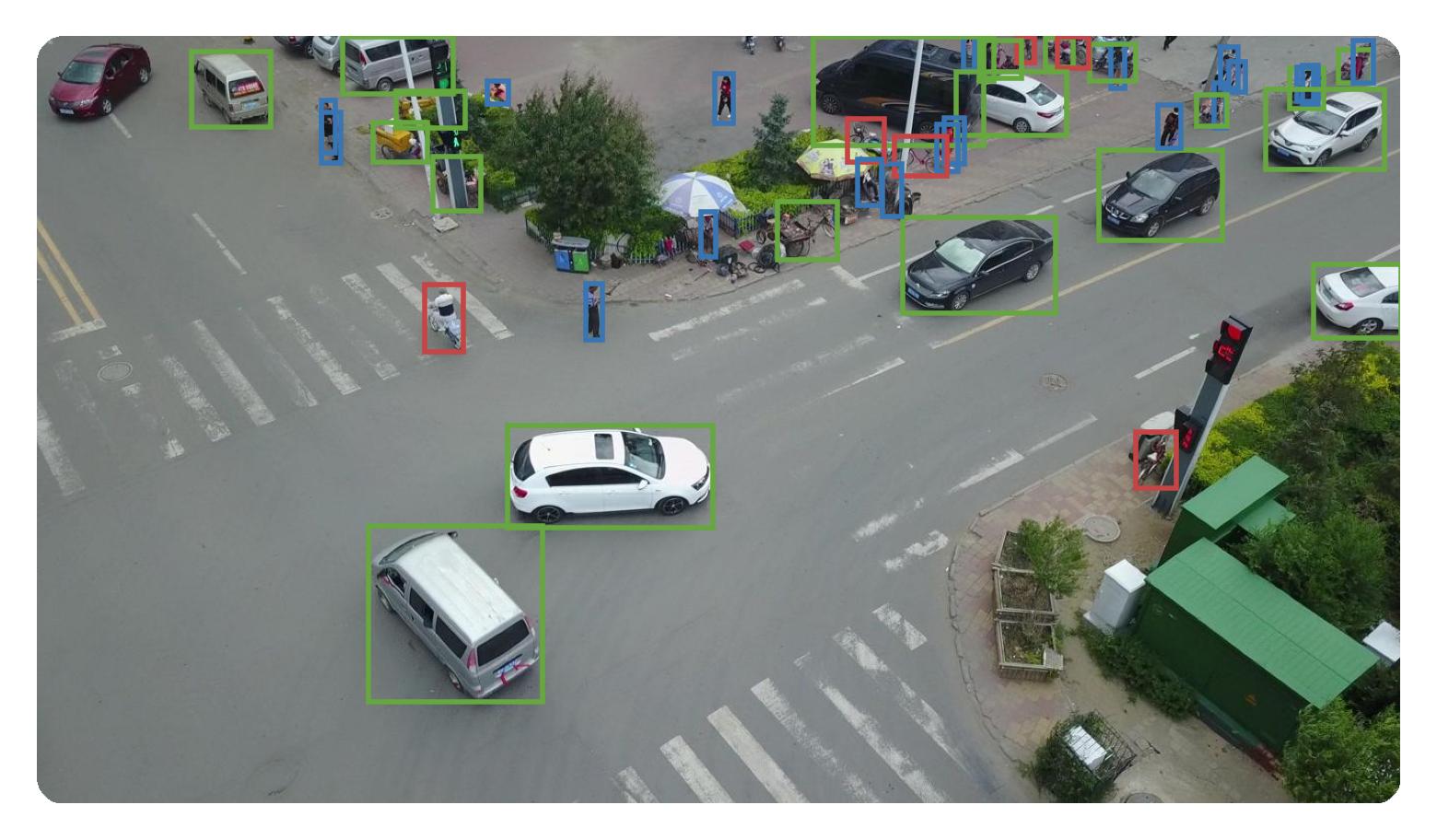} \\
  \end{tabular}
  \caption{Example images of the drone task.}
  \label{fig:drone-task}
\end{figure}

\noindent\textbf{Dataset Processing.}
We remove images with regions labeled as \textit{ignore}.
These regions have many small objects that are not individually labeled.
Bicycle and rider are merged into a combined bicycle class, as described in Section~\ref{sec:supp:thermal-task}.
Figure~\ref{fig:drone-task} depicts example images.

\noindent\textbf{Train, Val, and Test Split.}
The dataset provides a scene ID from which the train, validation, and test split is created like described in Section~\ref{supp:tasks:nuimages}.

\noindent\textbf{Labeling Policy}
Compared to the reference labeling policy (see Section~\ref{supp:tasks:nuimages}), the rider of a motorcycle is labeled as a person and is not part of the motorcycle class.
Some objects are left unlabeled; however, since the total number of labels is large, this is only a small fraction.
The objects have amodal bounding boxes, but due to the bird's-eye view, these are generally not much different from visual bounding boxes.

\noindent\textbf{D-RICO Classes.}

\begin{itemize} \setlength{\itemindent}{0.5cm}
    \item \textbf{person:} \textit{pedestrian}, \textit{people}

    \item \textbf{bicycle:} \textit{bicycle}

    \item \textbf{vehicle:} \textit{car}, \textit{van}, \textit{truck}, \textit{tricycle}, \textit{awning-tricycle}, \textit{bus}, \textit{motor}, \textit{others}
\end{itemize}

\noindent\textbf{EC-RICO Classes.}

\begin{itemize} \setlength{\itemindent}{0.5cm}
    \item \textbf{person:} \textit{Pedestrian}

    \item \textbf{car:} \textit{car}, \textit{van},\textit{tricycle}, \textit{awning-tricycle},

    \item \textbf{bicycle:} \textit{bicycle},
    \item \textbf{motorcycle:} \textit{motor},
    \item \textbf{truck:} \textit{truck}
    \item \textbf{bus:} \textit{bus}
\end{itemize}


\subsubsection{\texorpdfstring{Simulation (SHIFT~\cite{sun_shift_2022})}{Simulation (SHIFT)}}
SHIFT~\cite{sun_shift_2022} is a synthetic dataset created using the CARLA simulator, which contains 2.5 million annotated frames across 4,800 driving sequences. It encompasses a variety of environmental conditions, including weather changes, time of day, and traffic density, to investigate robustness in autonomous driving models.

\begin{figure}[t]
  \centering
  \begin{tabular}{cc}
    \includegraphics[width=0.45\linewidth]{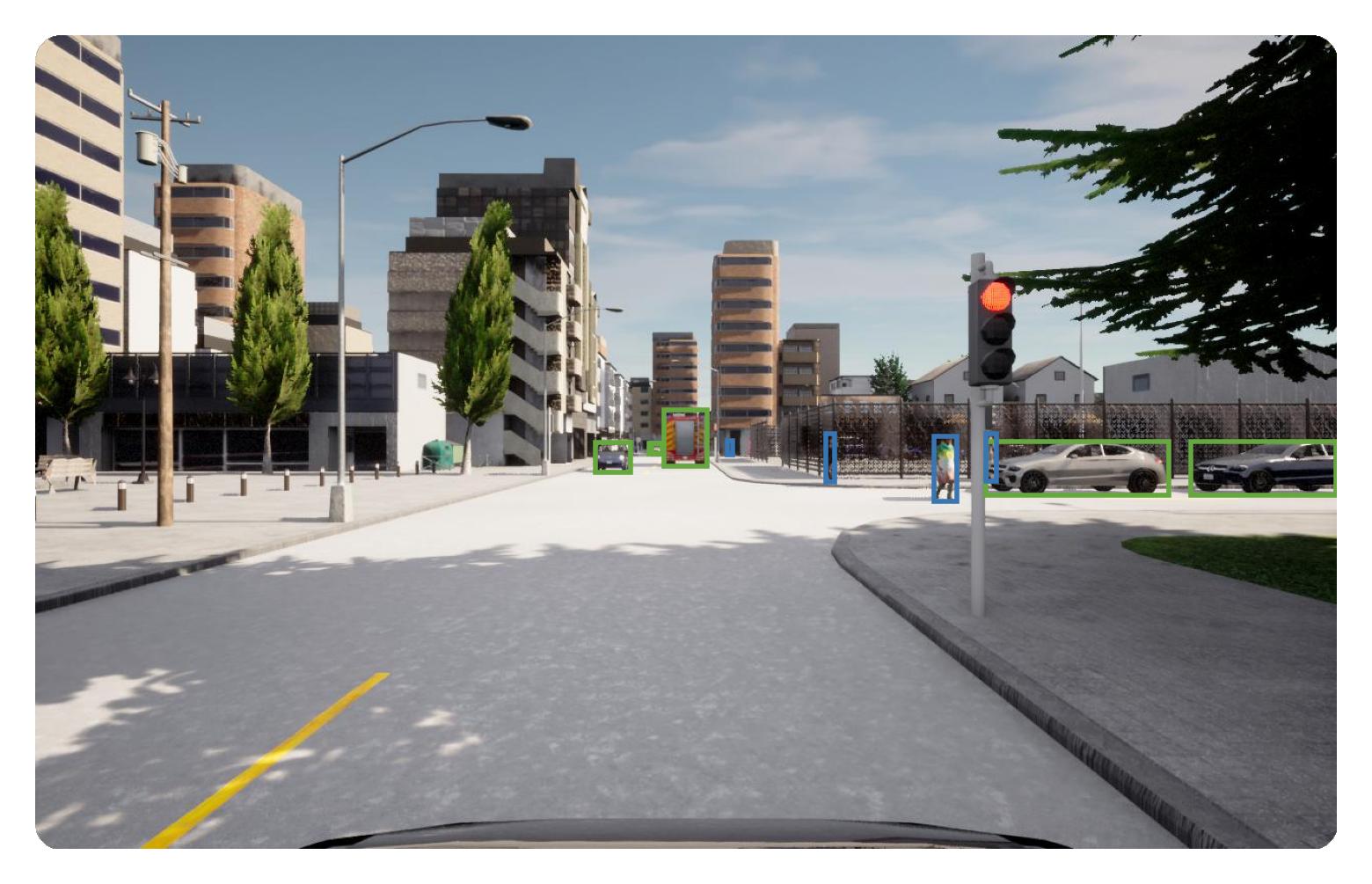} &
    \includegraphics[width=0.45\linewidth]{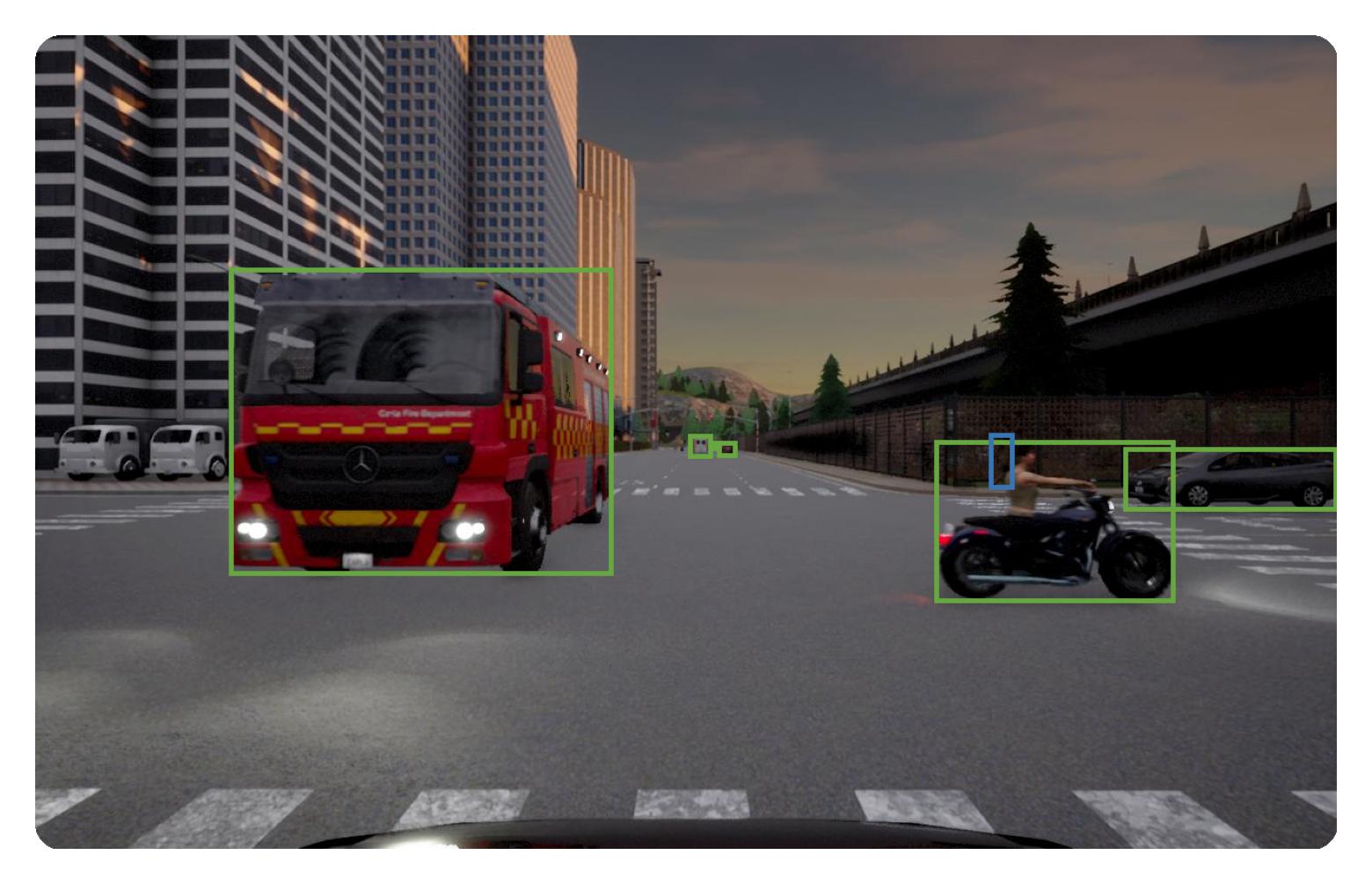} \\
    \includegraphics[width=0.45\linewidth]{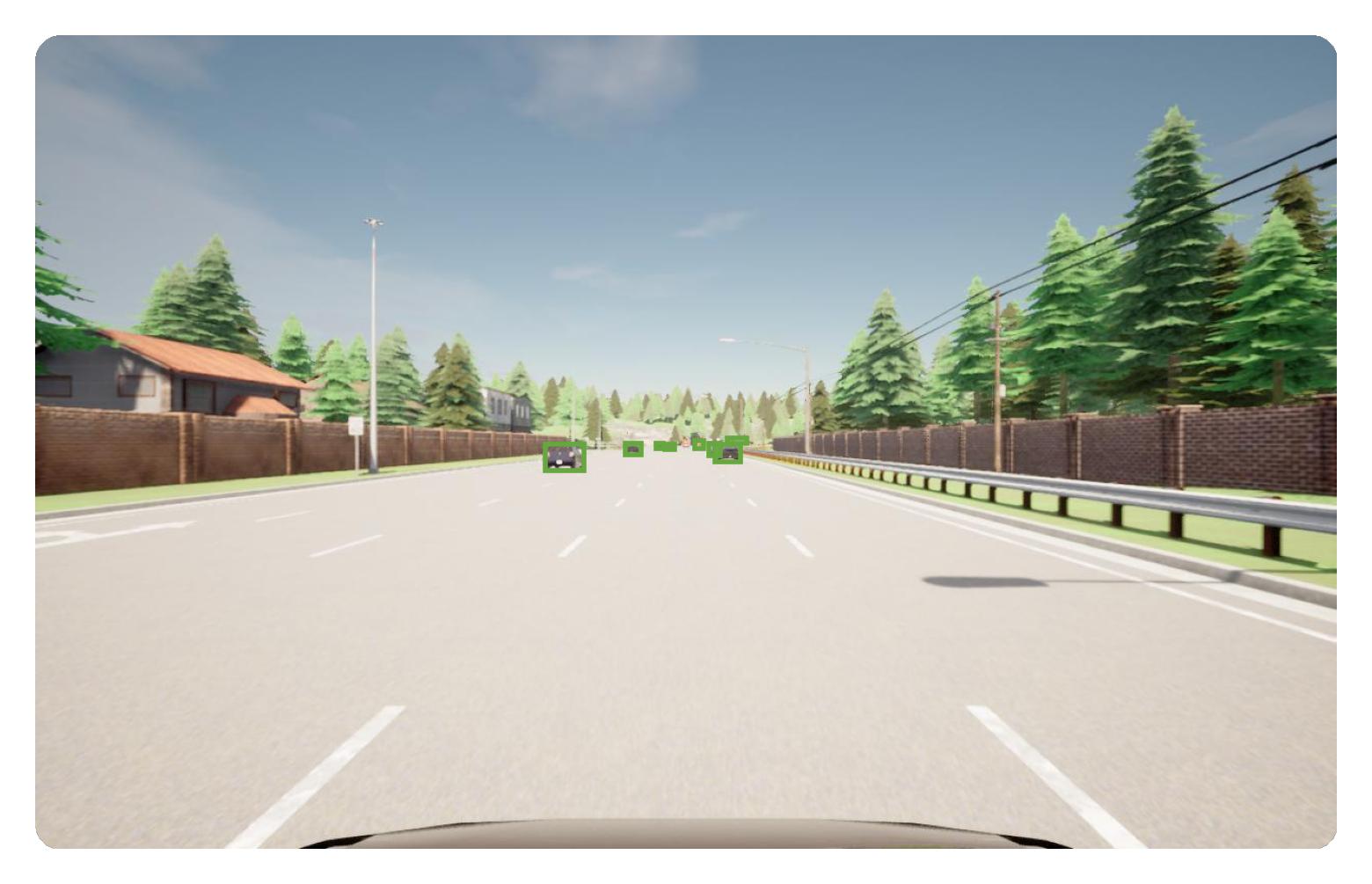} &
    \includegraphics[width=0.45\linewidth]{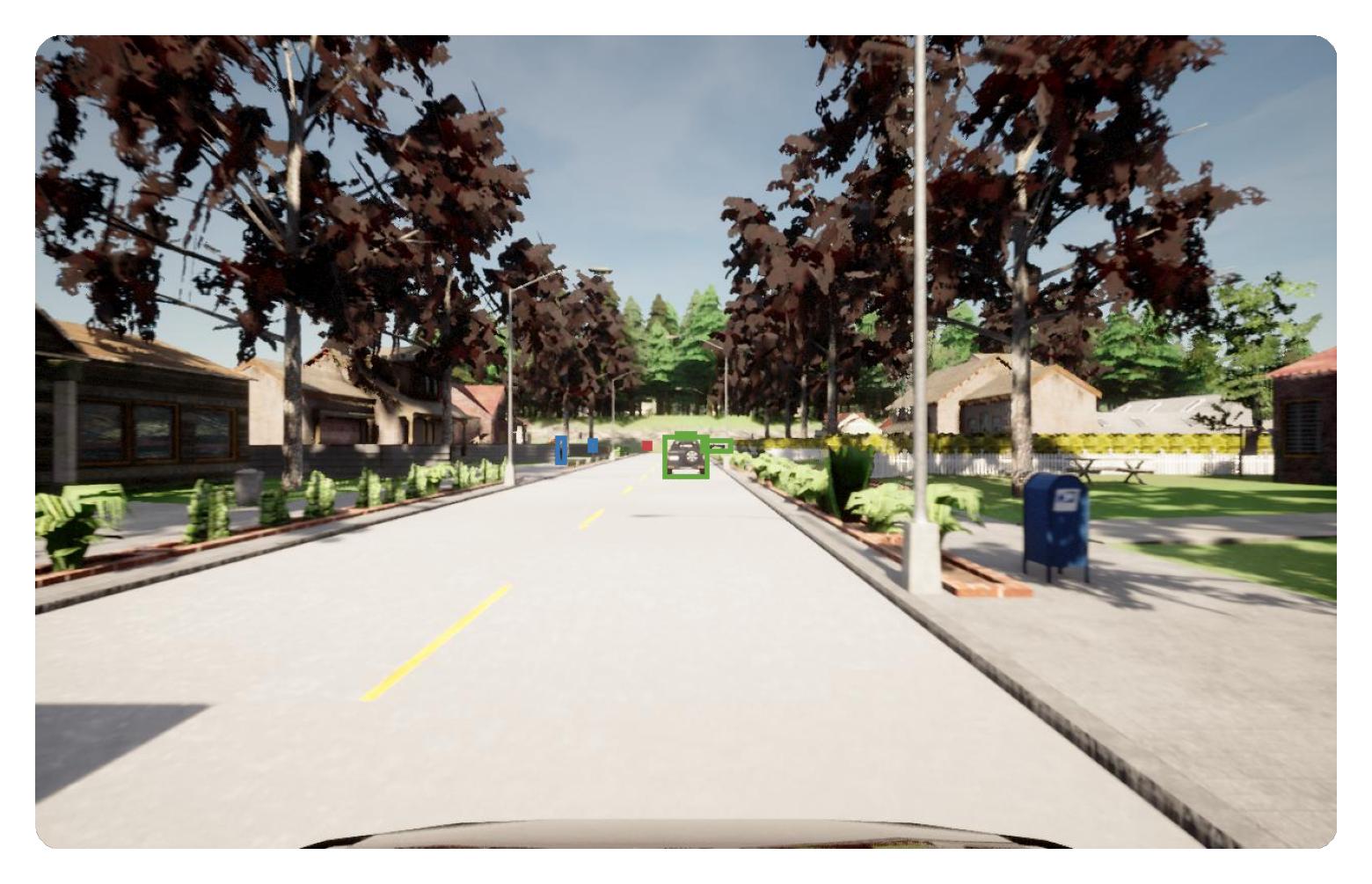} \\
  \end{tabular}
  \caption{Example images of the simulation task.}
  \label{fig:simulation-task}
\end{figure}

\noindent\textbf{Dataset Processing.}
As the frames are samples at a high frequency, we keep only every 50th frame.
Example images are given in Figure~\ref{fig:simulation-task}.

\noindent\textbf{Train, Val, and Test Split.}
The dataset provides a scene ID from which the train, validation, and test split is created like described in Section~\ref{supp:tasks:nuimages}.

\noindent\textbf{Labeling Policy}
The labeling policy matches the reference of Section~\ref{supp:tasks:nuimages}.

\noindent\textbf{D-RICO Classes.}

\begin{itemize} \setlength{\itemindent}{0.5cm}
    \item \textbf{person:} \textit{pedestrian}

    \item \textbf{bicycle:} \textit{bicycle}

    \item \textbf{vehicle:} \textit{car}, \textit{truck}, \textit{bus}, \textit{motorcycle}
\end{itemize}

\noindent\textbf{EC-RICO Classes.}

\begin{itemize} \setlength{\itemindent}{0.5cm}
    \item \textbf{person:} \textit{Pedestrian}

    \item \textbf{car:} \textit{car}

    \item \textbf{bicycle:} \textit{bicycle},
    \item \textbf{motorcycle:} \textit{motorcycle},
    \item \textbf{truck:} \textit{truck}
\end{itemize}

\subsubsection{\texorpdfstring{Fisheye Car (WoodScape~\cite{yogamani_woodscape_2021})}{Fisheye Car (WoodScape)}}\label{sec:supp:fisheye-task}
WoodScape~\cite{yogamani_woodscape_2021} is a fisheye dataset created for autonomous driving, featuring over 100,000 images captured with surround-view cameras. It offers multi-task annotations for object detection, depth estimation, and semantic segmentation, emphasizing addressing fisheye distortion in perception models.

\begin{figure}[t]
  \centering
  \begin{tabular}{cc}
    \includegraphics[width=0.45\linewidth]{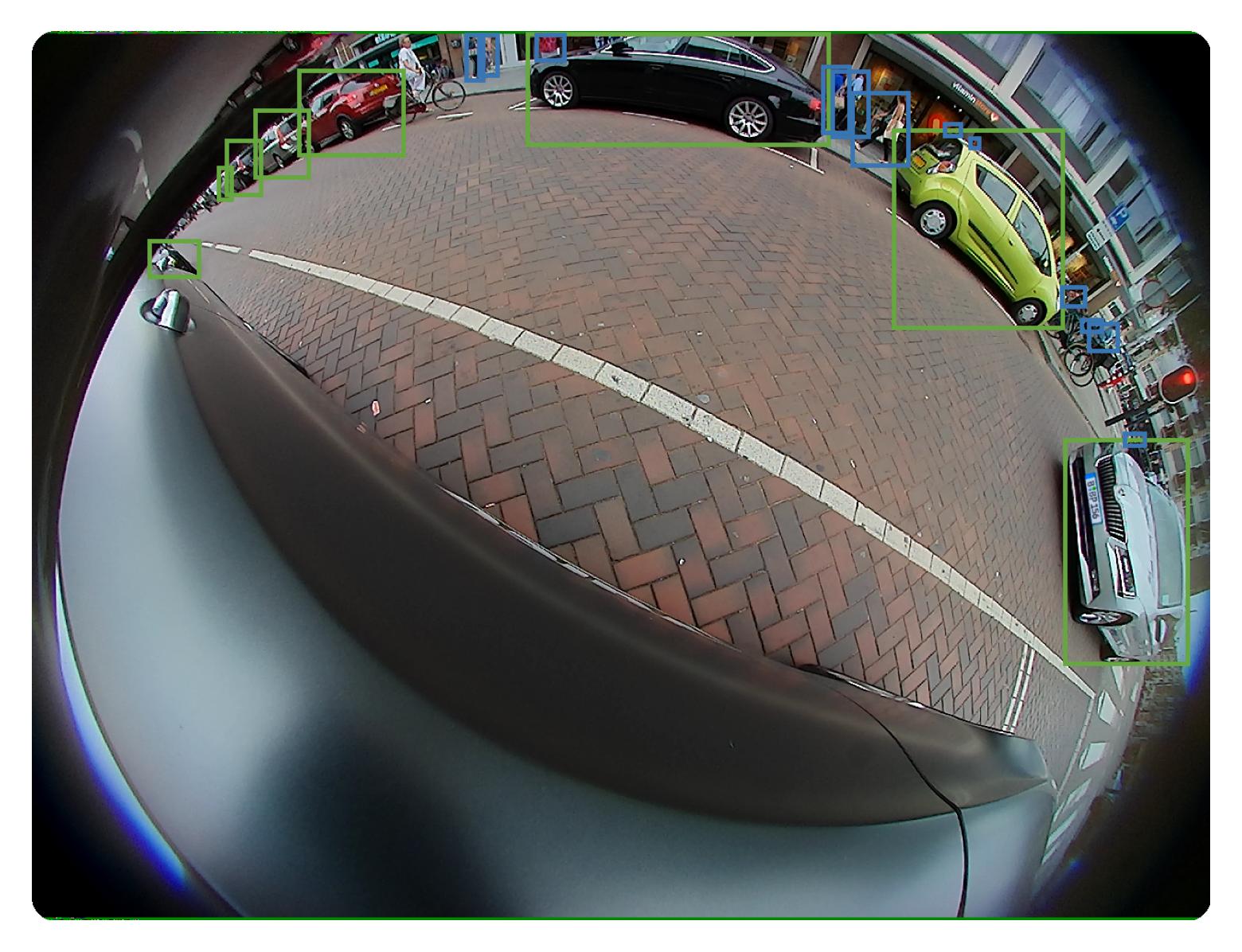} &
    \includegraphics[width=0.45\linewidth]{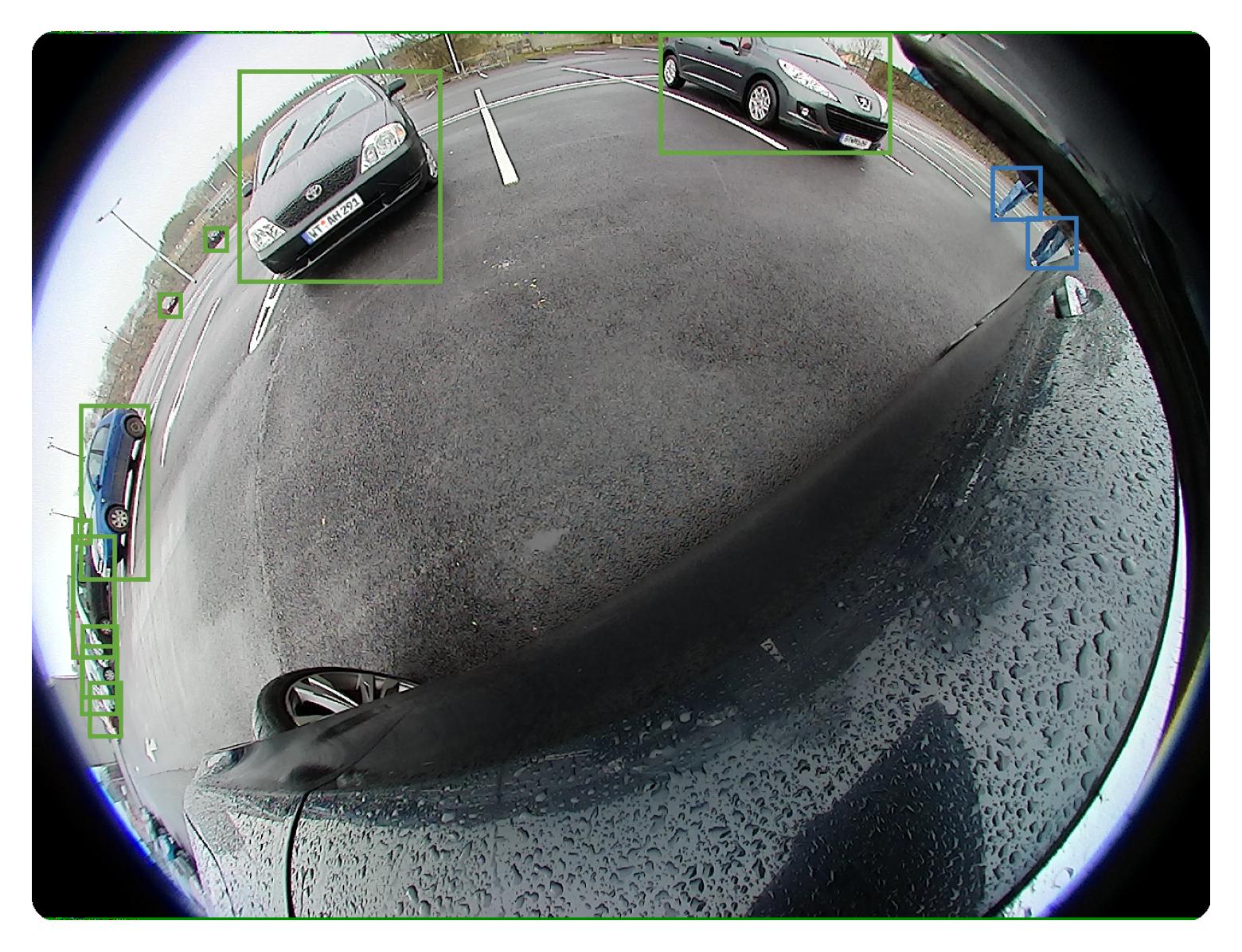} \\
    \includegraphics[width=0.45\linewidth]{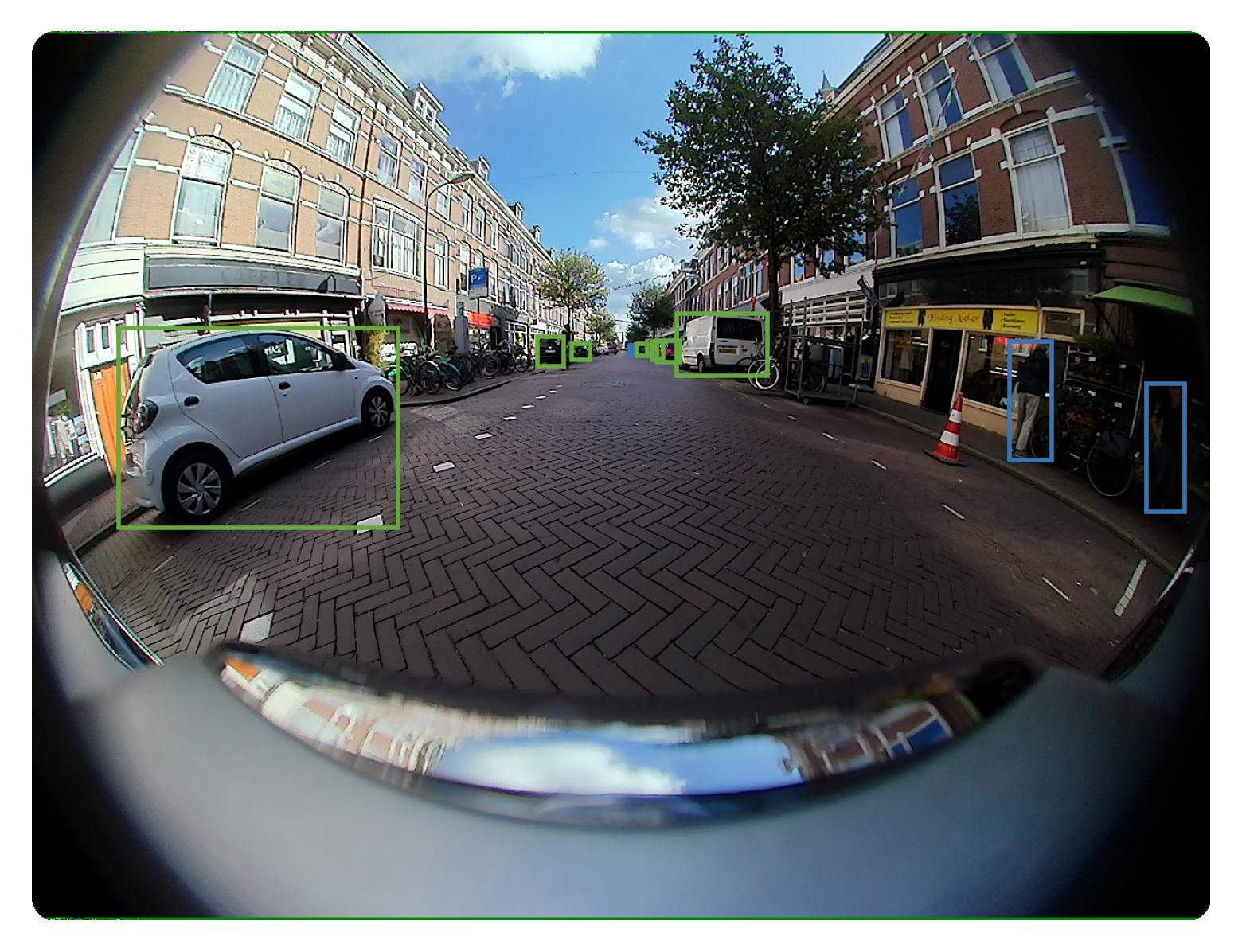} &
    \includegraphics[width=0.45\linewidth]{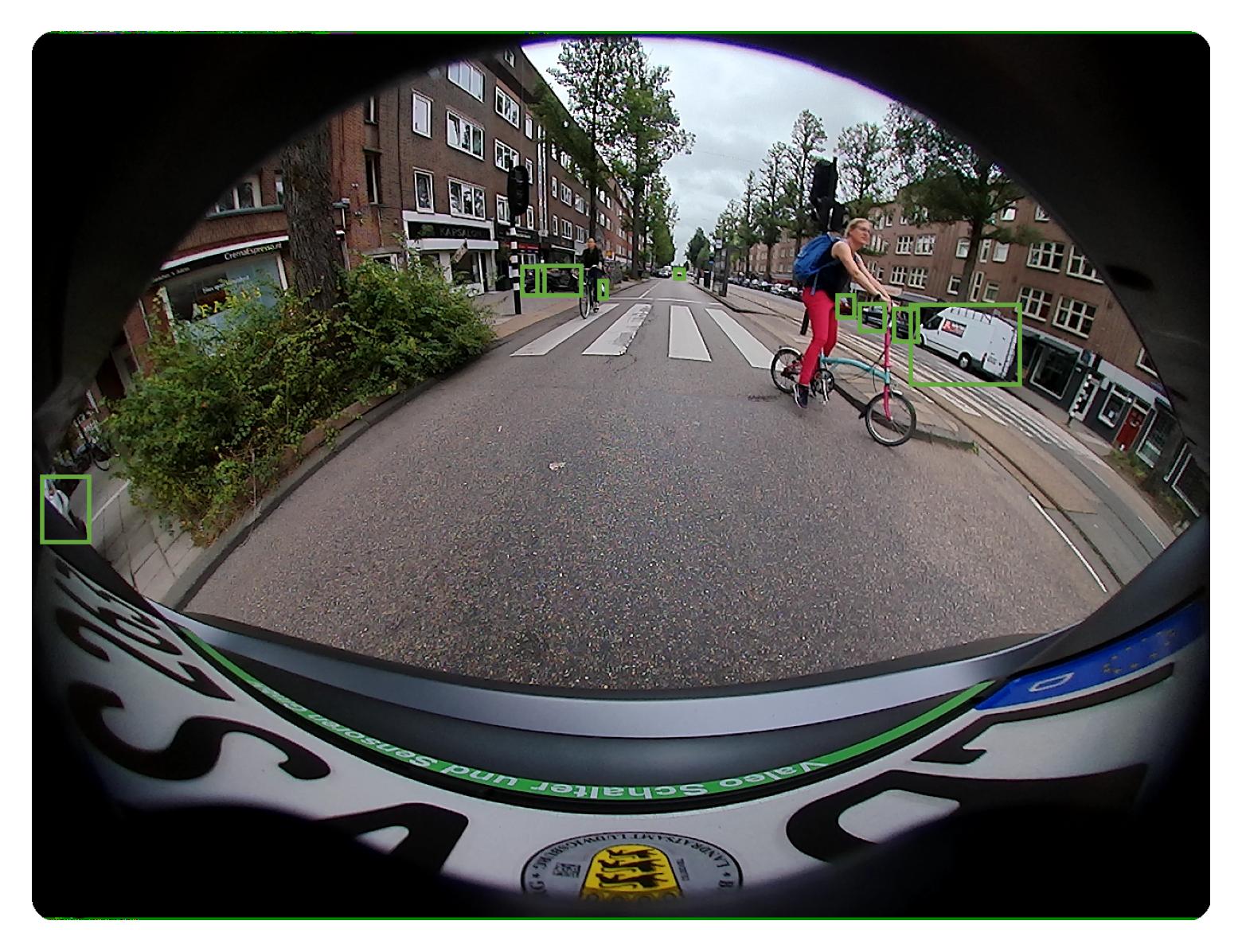} \\
  \end{tabular}
  \caption{Example images of the fisheye car task.}
  \label{fig:fisheye-car-task}
\end{figure}

\noindent\textbf{Dataset Processing.}
We recalculate the bounding boxes from the semantic and instance segmentation masks, as the provided bounding boxes are low-quality.
We discard annotations with bounding boxes smaller than $25\times25$.
To remove nighttime images from the dataset, we calculate each image's mean grayscale value and discard those below 0.3.
This value was found by manual inspection of the images.
See Figure~\ref{fig:fisheye-car-task} for examples.

\noindent\textbf{Train, Val, and Test Split.}
First, we discard two large sections of the image sequence to ensure a proper dataset split without overlap. This is necessary because no scene ID is provided, and we assume the data consists of a single continuous sequence. By removing sufficiently large portions, we break the sequence into three distinct parts while ensuring that the remaining data is split into approximately 60\% for training, 10\% for validation, and 30\% for testing. This approach ensures that each subset represents a separate segment of the sequence, preventing consecutive frames from appearing across different splits and improving the robustness of the evaluation.

\noindent\textbf{Labeling Policy}
Creating bounding boxes from segmentation masks can result in some individual objects being labeled with multiple bounding boxes due to interruptions in the segmentation.
Aside from that, the labeling policy aligns with the reference in Section~\ref{supp:tasks:nuimages}.

\noindent\textbf{D-RICO Classes.}

\begin{itemize} \setlength{\itemindent}{0.5cm}
    \item \textbf{person:} \textit{person}

    \item \textbf{vehicle:} \textit{car}, \textit{train/tram}, \textit{truck}, \textit{trailer}, \textit{van}, \textit{caravan}, \textit{bus}, \textit{motorcycle}
\end{itemize}

\subsubsection{\texorpdfstring{RGB + Thermal Fusion (SMOD~\cite{chen_amfd_2024})}{RGB + Thermal Fusion (SMOD)}}
The SMOD dataset~\cite{chen_amfd_2024} comprises 8,676 well-aligned RGB and thermal image pairs, gathered for pedestrian detection and multispectral object recognition. The dataset features four object categories with comprehensive occlusion annotations, highlighting challenging low-light scenarios.

\begin{figure}[t]
  \centering
  \begin{tabular}{cc}
    \includegraphics[width=0.45\linewidth]{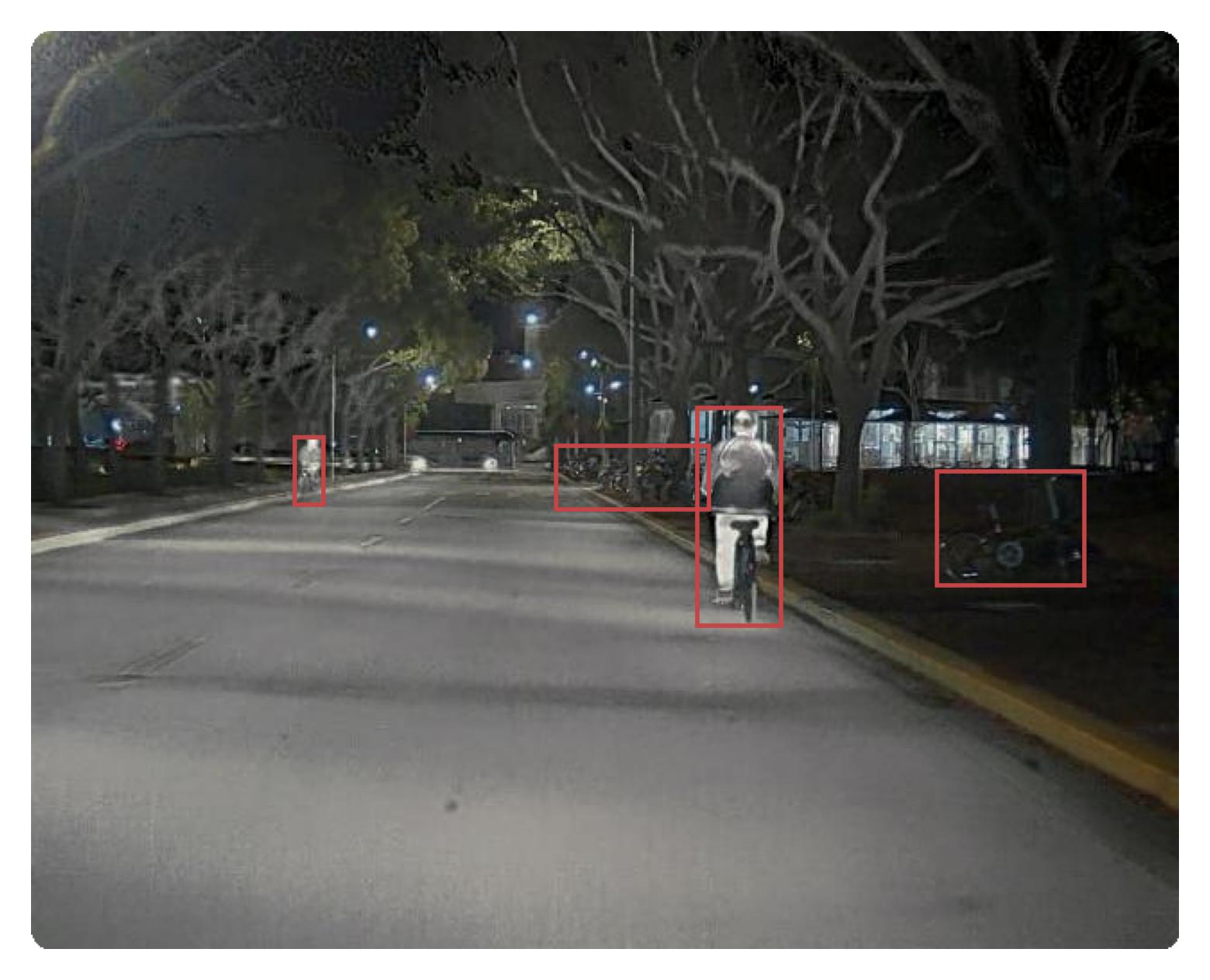} &
    \includegraphics[width=0.45\linewidth]{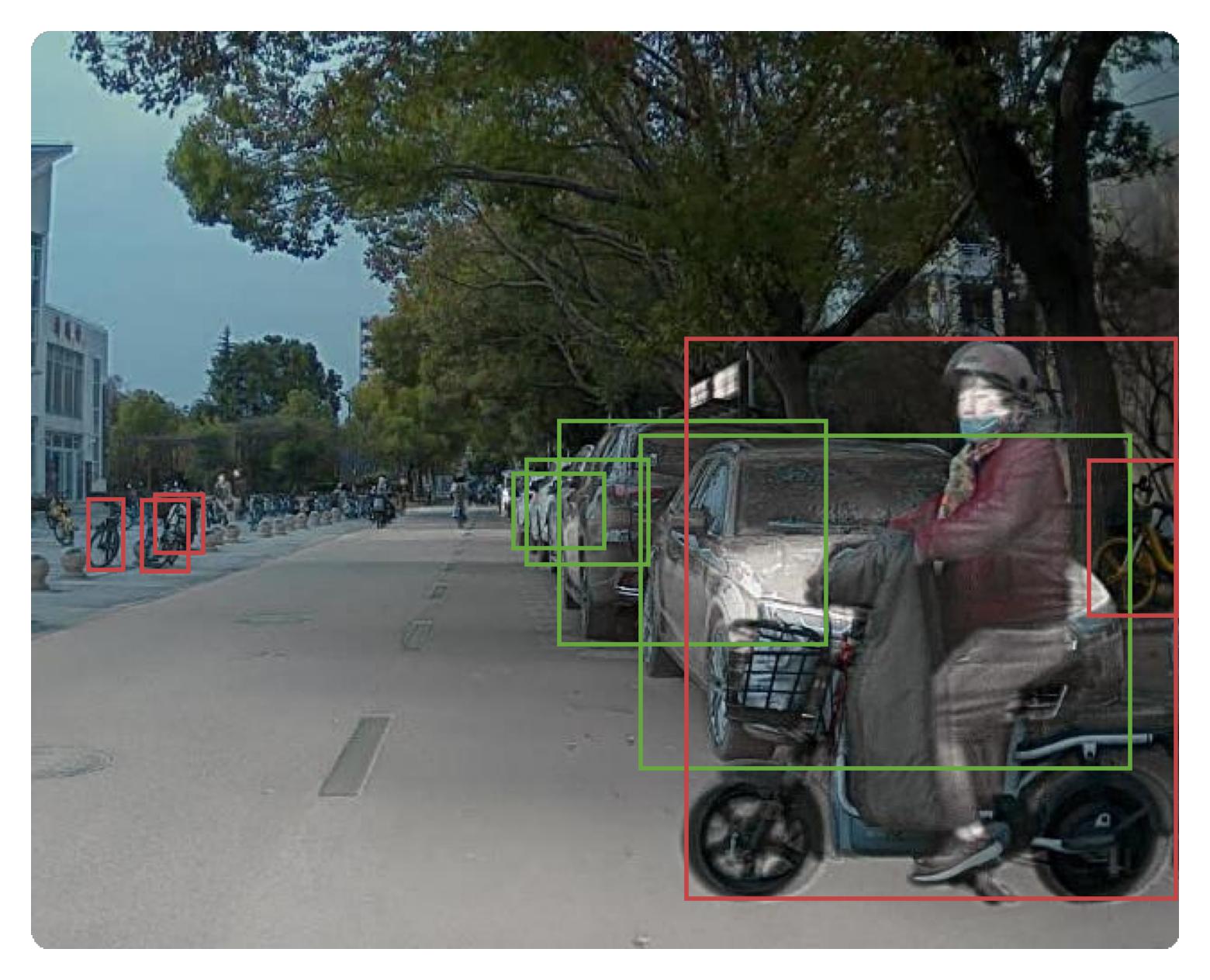} \\
    \includegraphics[width=0.45\linewidth]{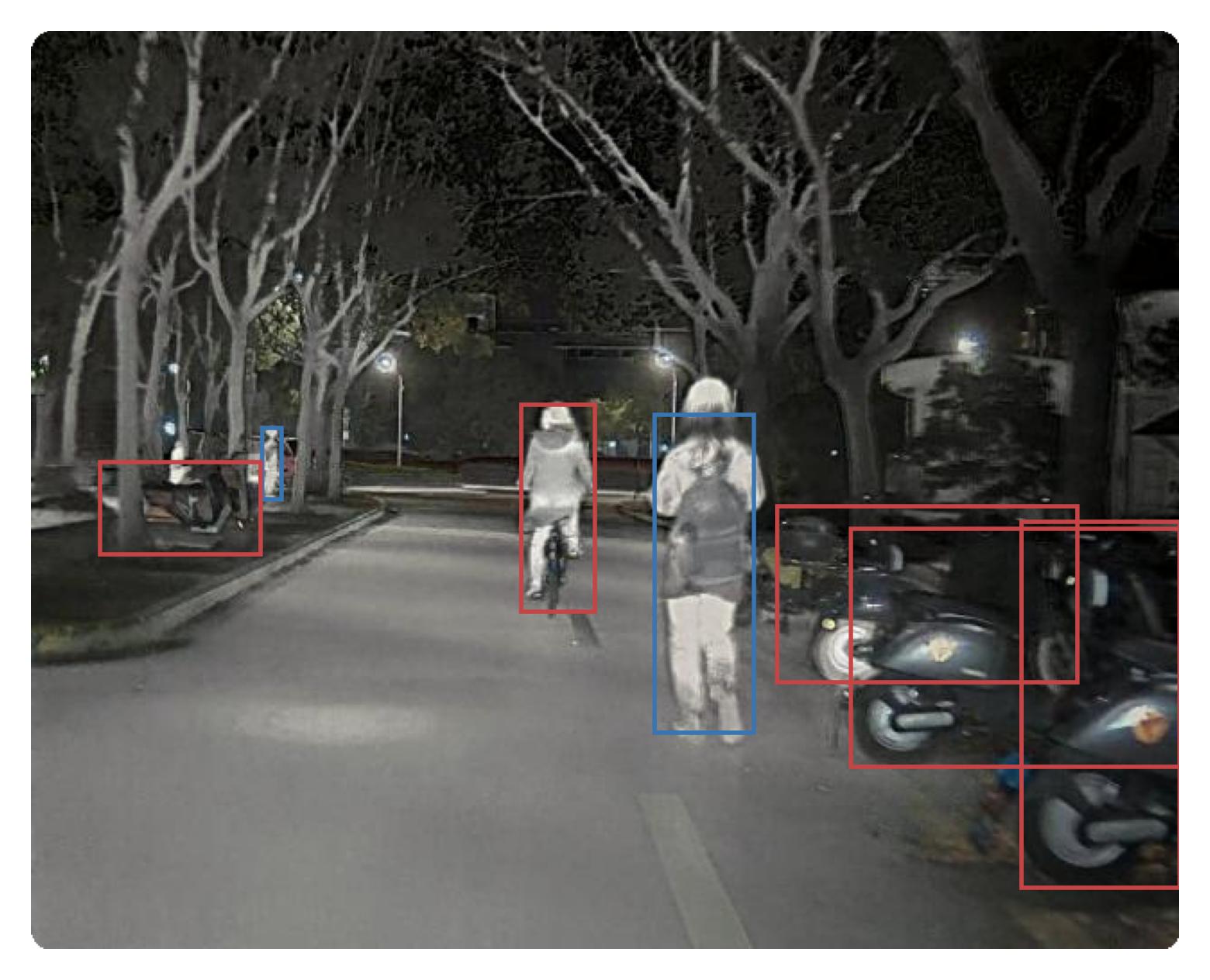} &
    \includegraphics[width=0.45\linewidth]{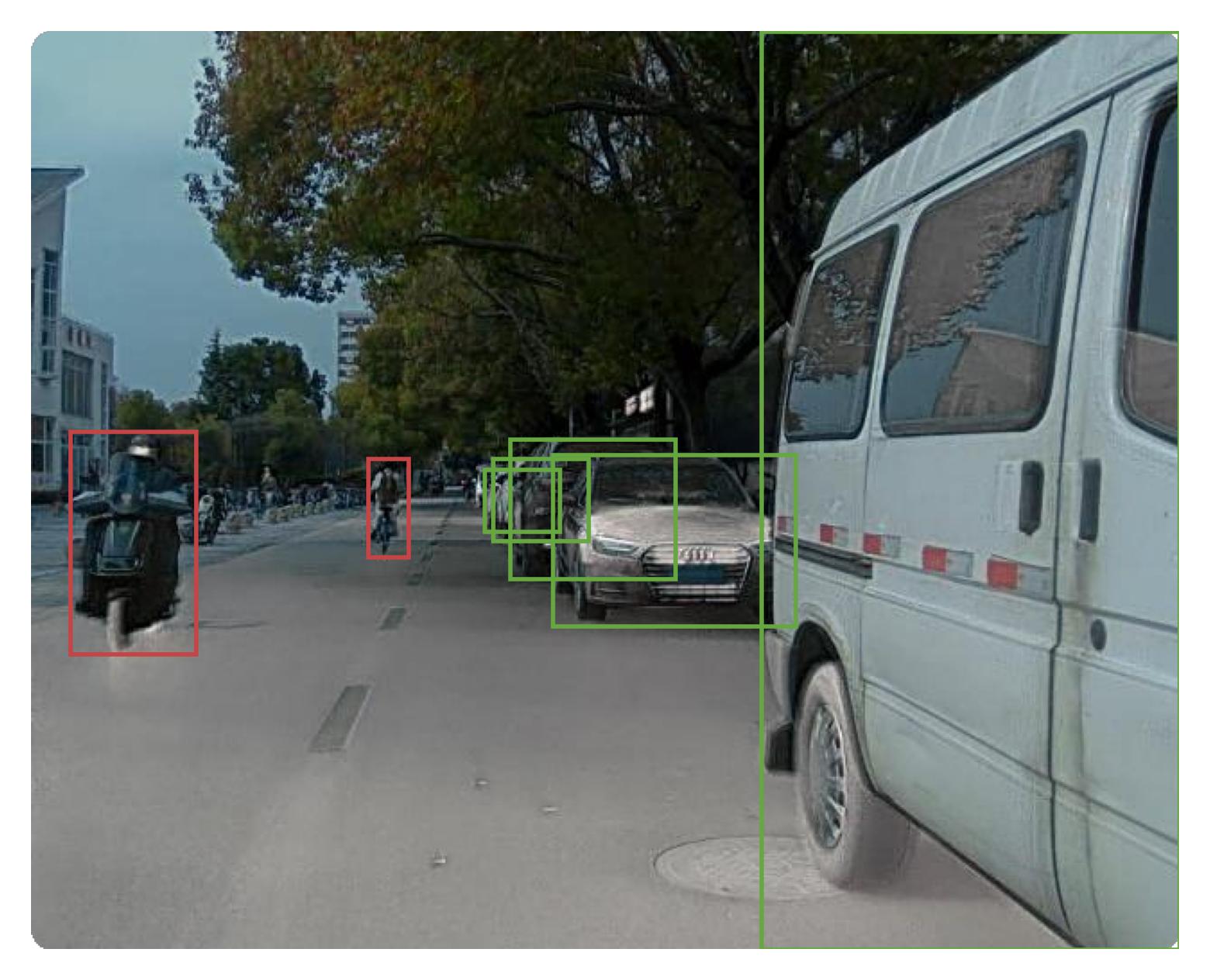} \\
  \end{tabular}
  \caption{Example images of the RGB + thermal fusion task.}
  \label{fig:rgb-thermal-fusion-task}
\end{figure}

\noindent\textbf{Dataset Processing.}
We first select low visibility images by calculating the mean gray value of each image and keep only those with a value below 0.35.
For fusing the two modalities we use the neural network approach by~\cite{zhang_ifcnn_2020}.
It inputs the normalized RGB and thermal image and outputs a fused RGB image.
See Figure~\ref{fig:rgb-thermal-fusion-task} for examples.

\noindent\textbf{Train, Val, and Test Split.}
As no scene IDs are available, the dataset is split into train, val, and test set like described in Section~\ref{sec:supp:fisheye-task}

\noindent\textbf{Labeling Policy}
In this dataset, motorcycles are labeled as bicycles and, hence, are not part of the vehicle class.
There are also numerous objects not labeled.

\noindent\textbf{D-RICO Classes.}

\begin{itemize} \setlength{\itemindent}{0.5cm}
    \item \textbf{person:} \textit{person}
    \item \textbf{bicycle:} \textit{bicycle}, \textit{rider}
    \item \textbf{vehicle:} \textit{car}
\end{itemize}

\subsubsection{\texorpdfstring{Video Game (Sim10k~\cite{johnson-roberson_driving_2017})}{Video Game (Sim10k)}}
SIM10K~\cite{johnson-roberson_driving_2017} is a synthetic dataset created using the \textit{Grand Theft Auto V} game, consisting of 10,000 images annotated with vehicle information. It is used to investigate domain adaptation and transfer learning from synthetic to real-world driving datasets.

\begin{figure}[t]
  \centering
  \begin{tabular}{cc}
    \includegraphics[width=0.45\linewidth]{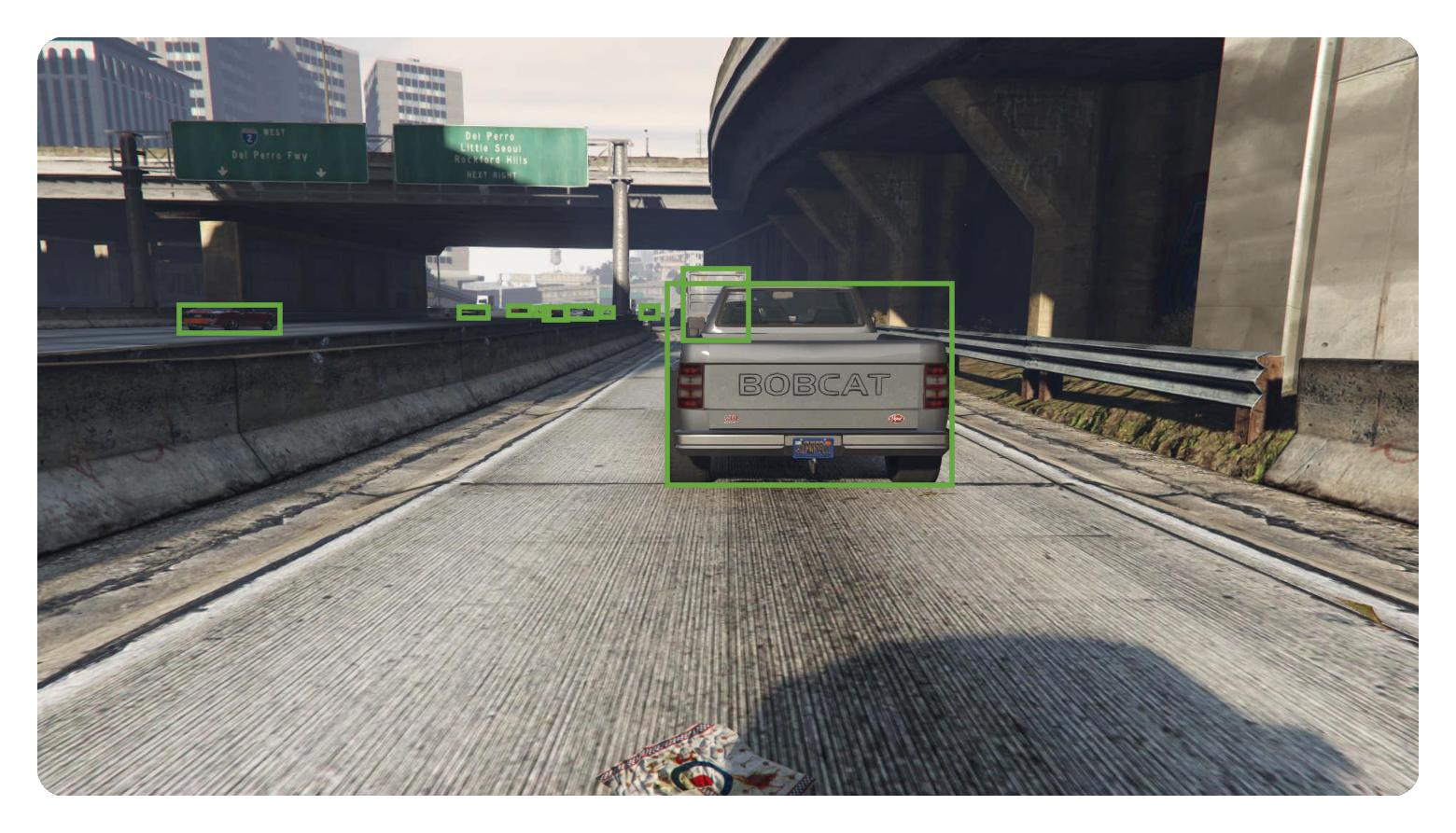} &
    \includegraphics[width=0.45\linewidth]{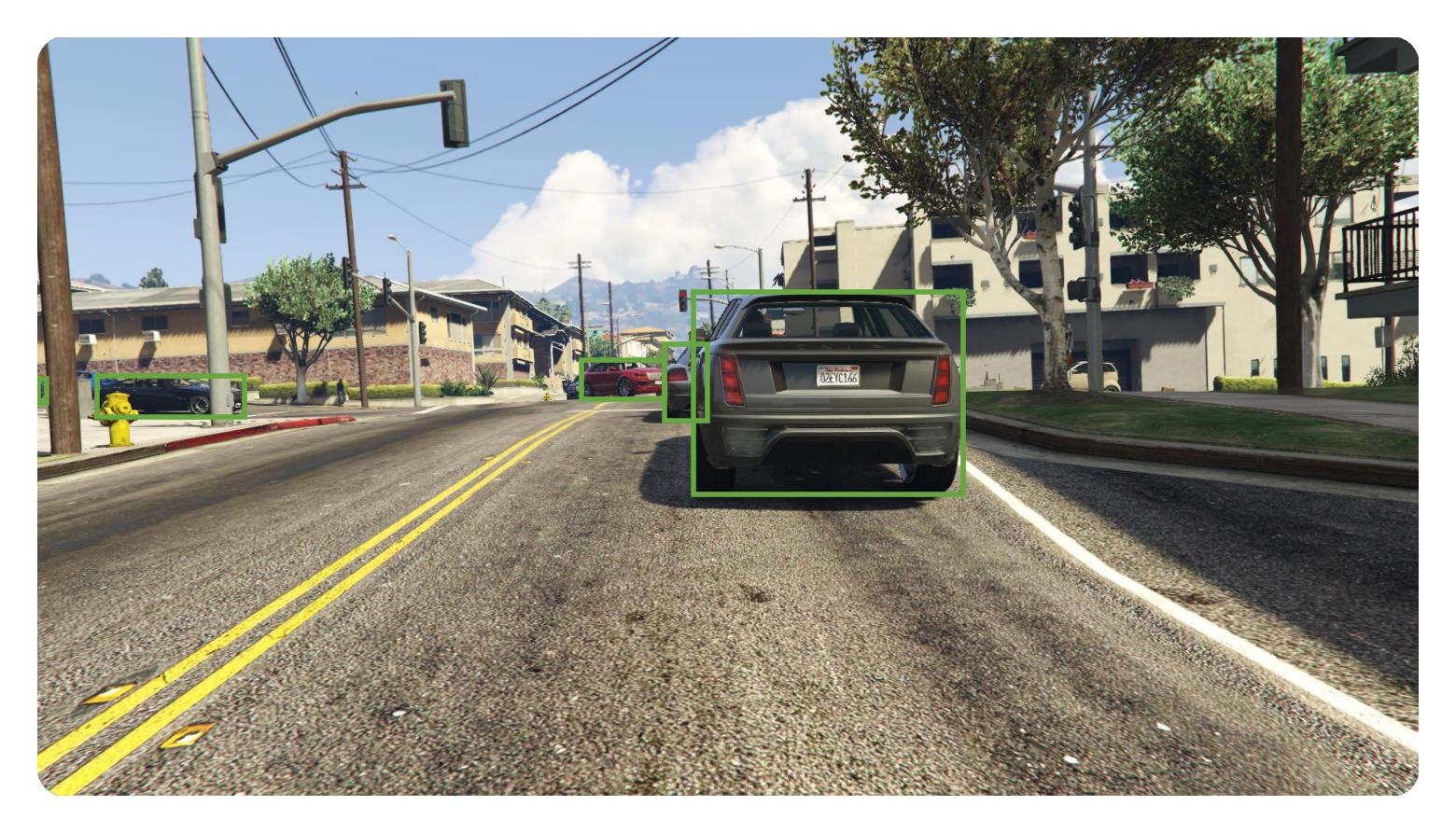} \\
    \includegraphics[width=0.45\linewidth]{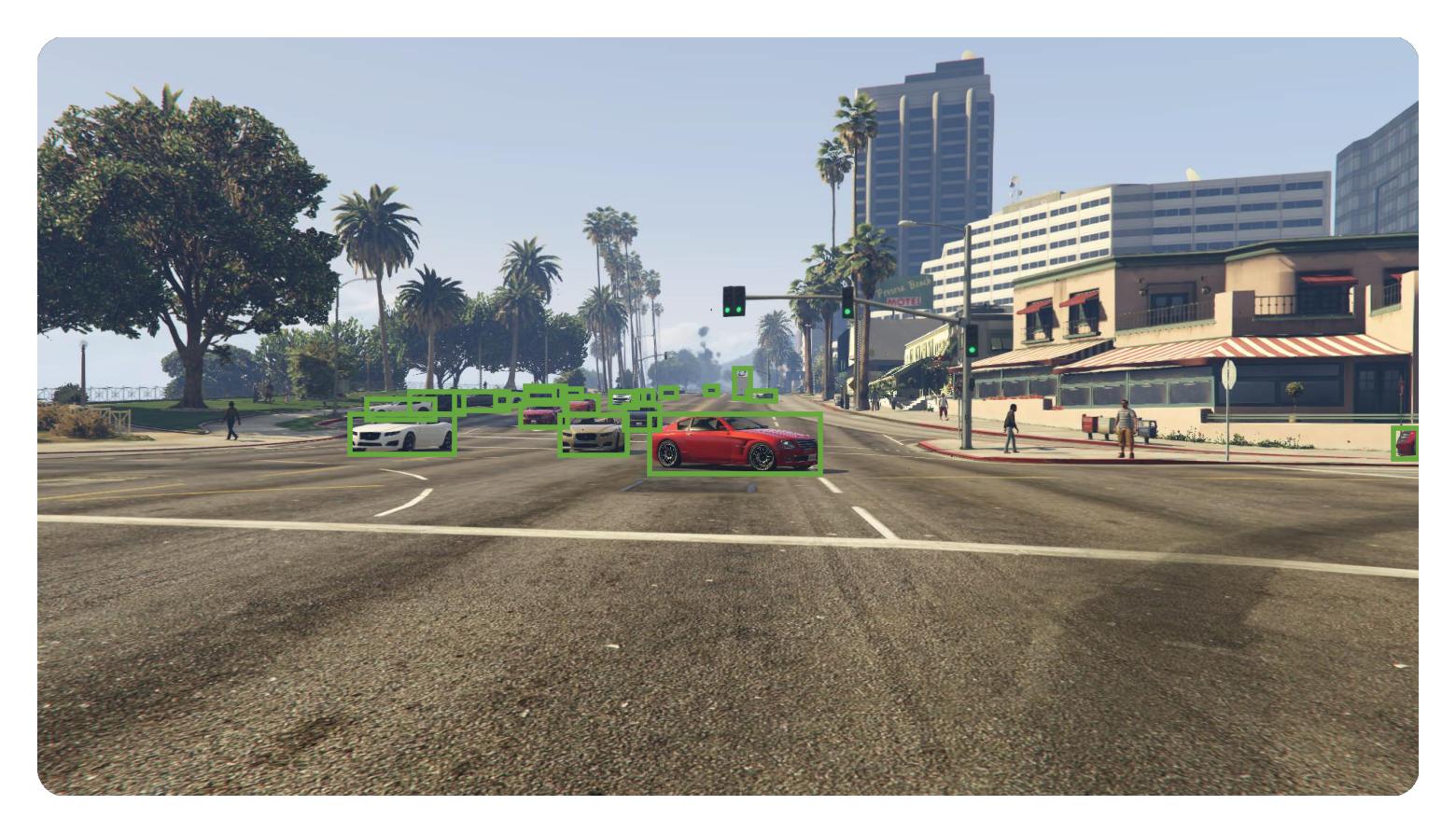} &
    \includegraphics[width=0.45\linewidth]{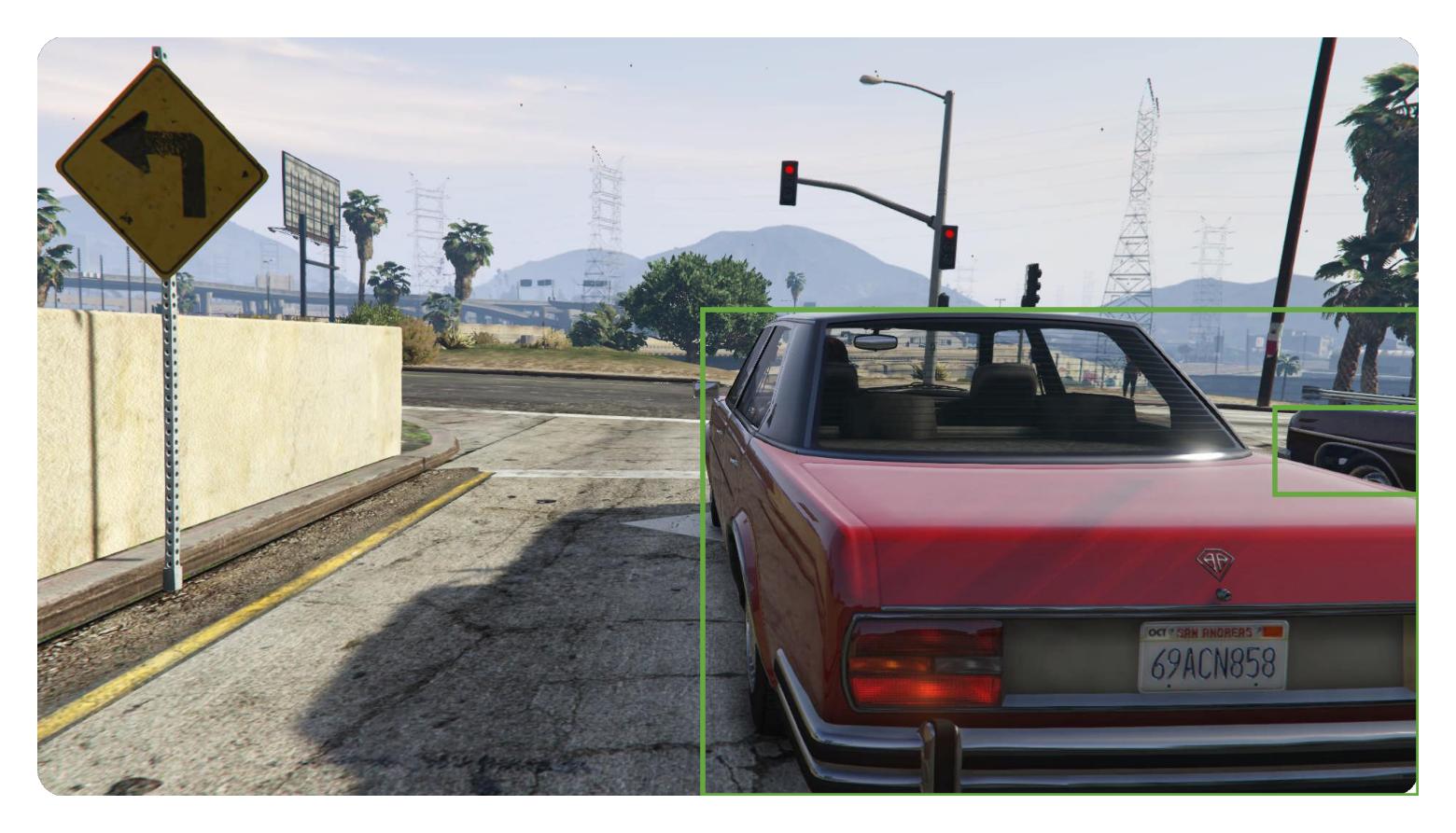} \\
  \end{tabular}
  \caption{Example images of the video game task.}
  \label{fig:video-game-task}
\end{figure}

\noindent\textbf{Dataset Processing.}
We select high visible images like described in Section~\ref{sec:supp:fisheye-task} with a gray value threshold of 0.4.
Examples are presented in Figure~\ref{fig:video-game-task}.

\noindent\textbf{Train, Val, and Test Split.}
As the dataset does not include scene IDs, we split it up like in Section~\ref{sec:supp:fisheye-task} described.

\noindent\textbf{Labeling Policy}
Compared to Section~\ref{supp:tasks:nuimages}, the rider on the motorcycle is not included in the bounding box.
The rider, however, is also not labeled as a person.

\noindent\textbf{D-RICO Classes.}
We only use the vehicle class, constructed from \textit{car} and \textit{motorbike} classes.

\subsubsection{\texorpdfstring{Nighttime (BDD100K~\cite{yu_bdd100k_2020})}{Nighttime (BDD100K)}}
BDD100K~\cite{yu_bdd100k_2020} is a large-scale driving dataset of 100,000 videos captured under various conditions, including nighttime driving. It offers extensive annotations for multiple perception tasks, making it one of the most comprehensive datasets in autonomous driving.

\begin{figure}[t]
  \centering
  \begin{tabular}{cc}
    \includegraphics[width=0.45\linewidth]{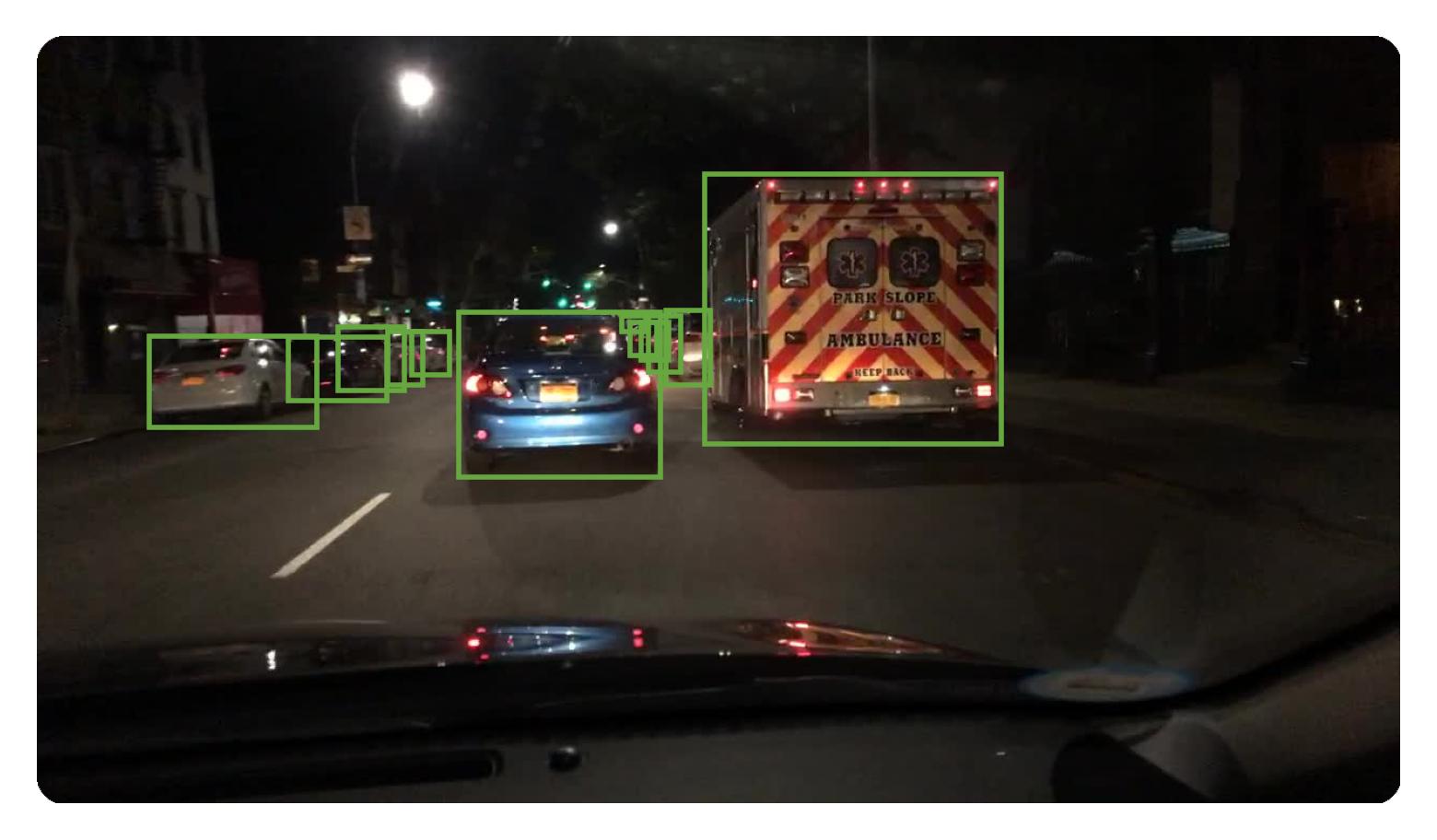} &
    \includegraphics[width=0.45\linewidth]{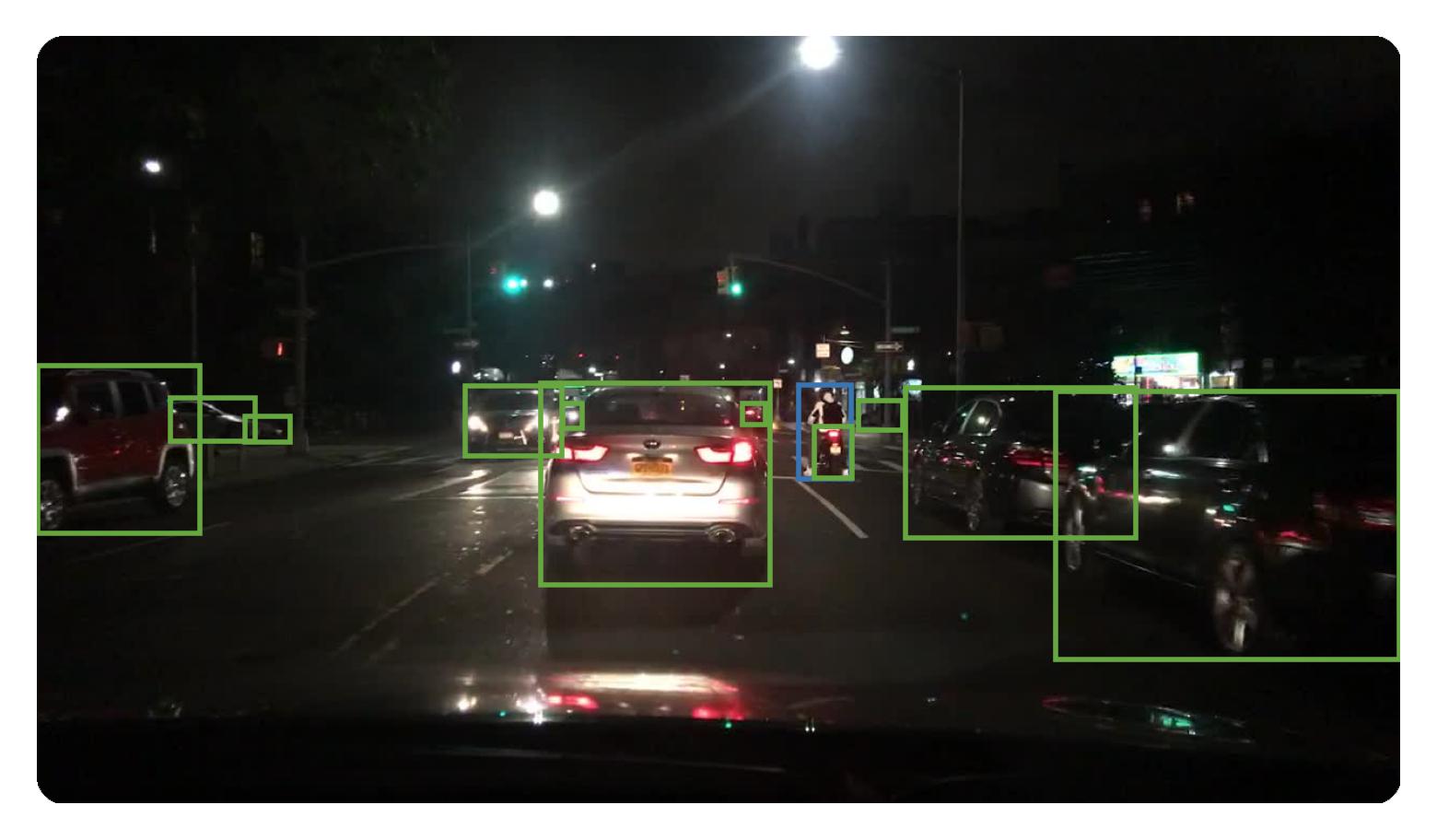} \\
    \includegraphics[width=0.45\linewidth]{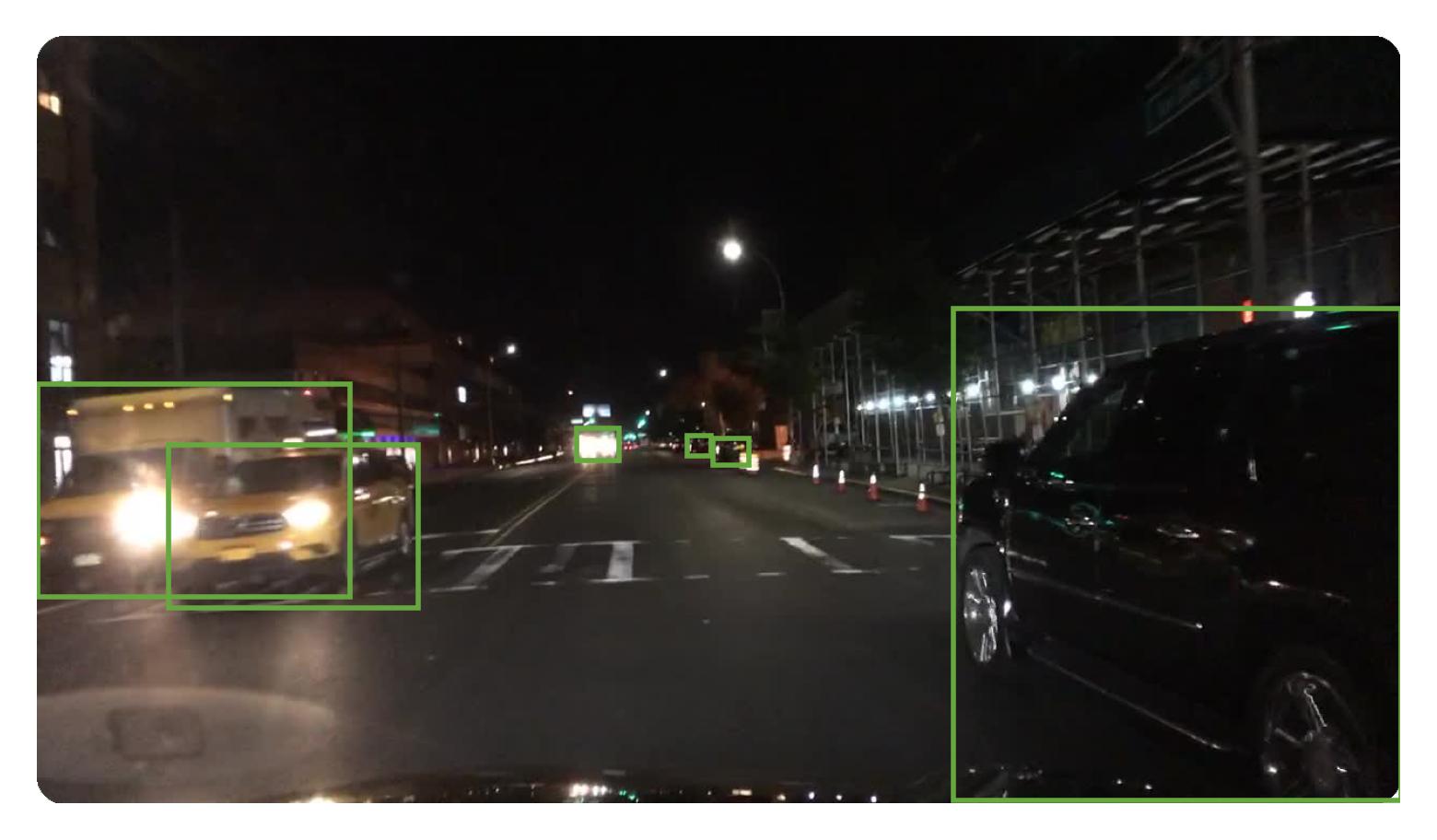} &
    \includegraphics[width=0.45\linewidth]{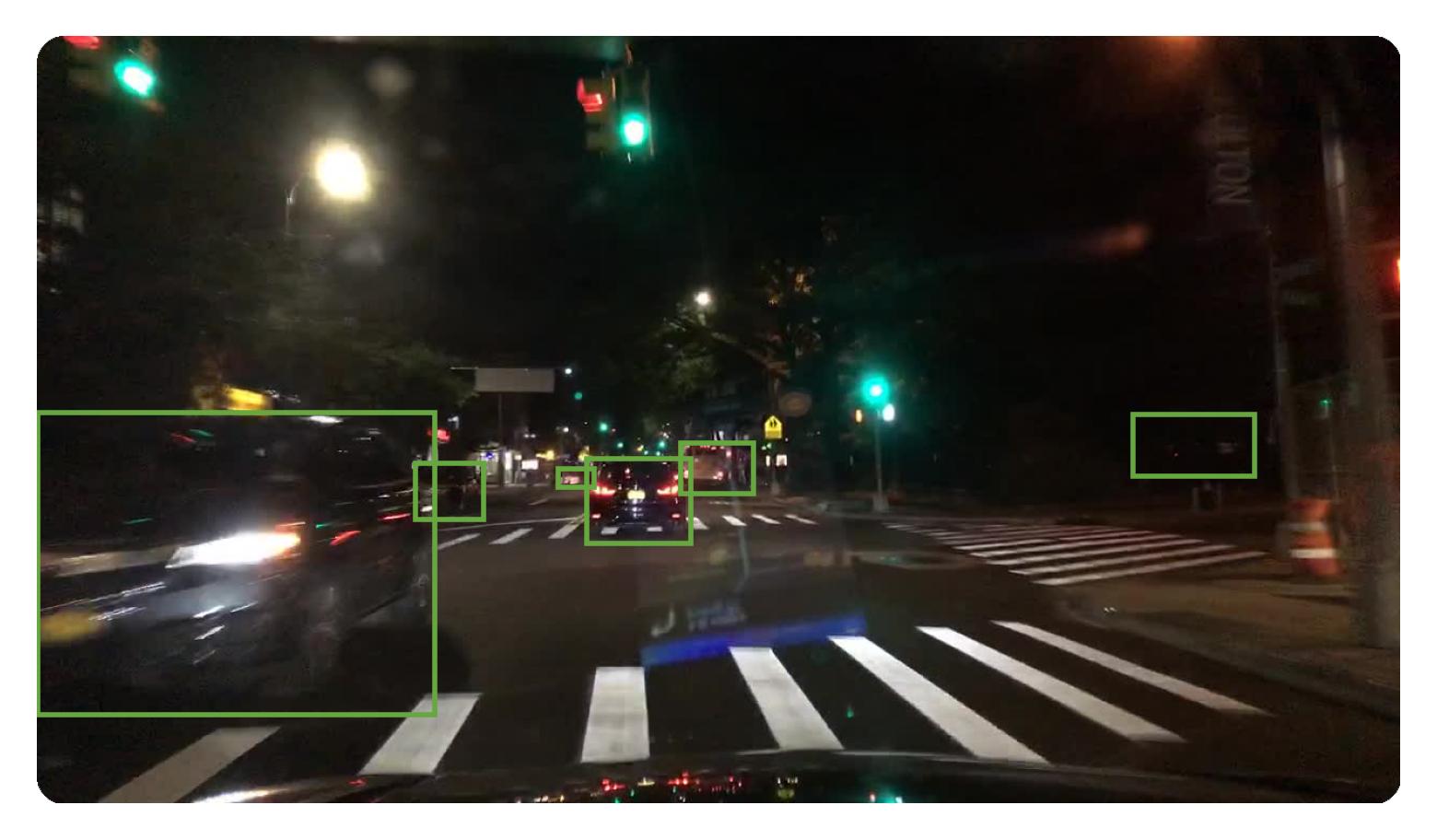} \\
  \end{tabular}
  \caption{Example images of the nighttime task.}
  \label{fig:nighttime-task}
\end{figure}

\noindent\textbf{Dataset Processing.}
Based on the provided meta-data, we select only those images from the dataset that are labeled as night and clear weather.
Rider and bicycle are merged like in Section~\ref{sec:supp:thermal-task} described.
Figure~\ref{fig:nighttime-task} shows example images.

\noindent\textbf{Train, Val, and Test Split.}
Based on the available scene IDs, the data is split into training, validation, and test like in Section~\ref{supp:tasks:nuimages} described.

\noindent\textbf{Labeling Policy}
The labeling policy aligns with the reference in Section~\ref{supp:tasks:nuimages}.

\noindent\textbf{D-RICO Classes.}
\begin{itemize} \setlength{\itemindent}{0.5cm}
    \item \textbf{person:} \textit{person}, \textit{rider}
    \item \textbf{bicycle:} \textit{bike}
    \item \textbf{vehicle:} \textit{car}, \textit{truck}, \textit{motor}, \textit{train}, \textit{bus}
\end{itemize}

\noindent\textbf{EC-RICO Classes.}

\begin{itemize} \setlength{\itemindent}{0.5cm}
    \item \textbf{person:} \textit{person}, \textit{rider}
    \item \textbf{car:} \textit{car}
    \item \textbf{bicycle:} \textit{bike},
    \item \textbf{motorcycle:} \textit{motor},
    \item \textbf{truck:} \textit{truck}
    \item \textbf{bus:} \textit{bus}
    \item \textbf{traffic light:} \textit{traffic light}
    \item \textbf{street sign:} \textit{street sign}
\end{itemize}

\subsubsection{\texorpdfstring{Fisheye Indoor (LOAF~\cite{yang_large-scale_2023})}{Fisheye Indoor (LOAF)}}
LOAF~\cite{yang_large-scale_2023} is a dataset designed for person detection in fisheye images, comprising 43,000 frames extracted from 70 videos filmed in indoor and outdoor surveillance environments. It features radius-aligned bounding boxes specifically adapted for fisheye distortion, emphasizing person localization from overhead camera perspectives.

\begin{figure}[t]
  \centering
  \begin{tabular}{cc}
    \includegraphics[width=0.45\linewidth]{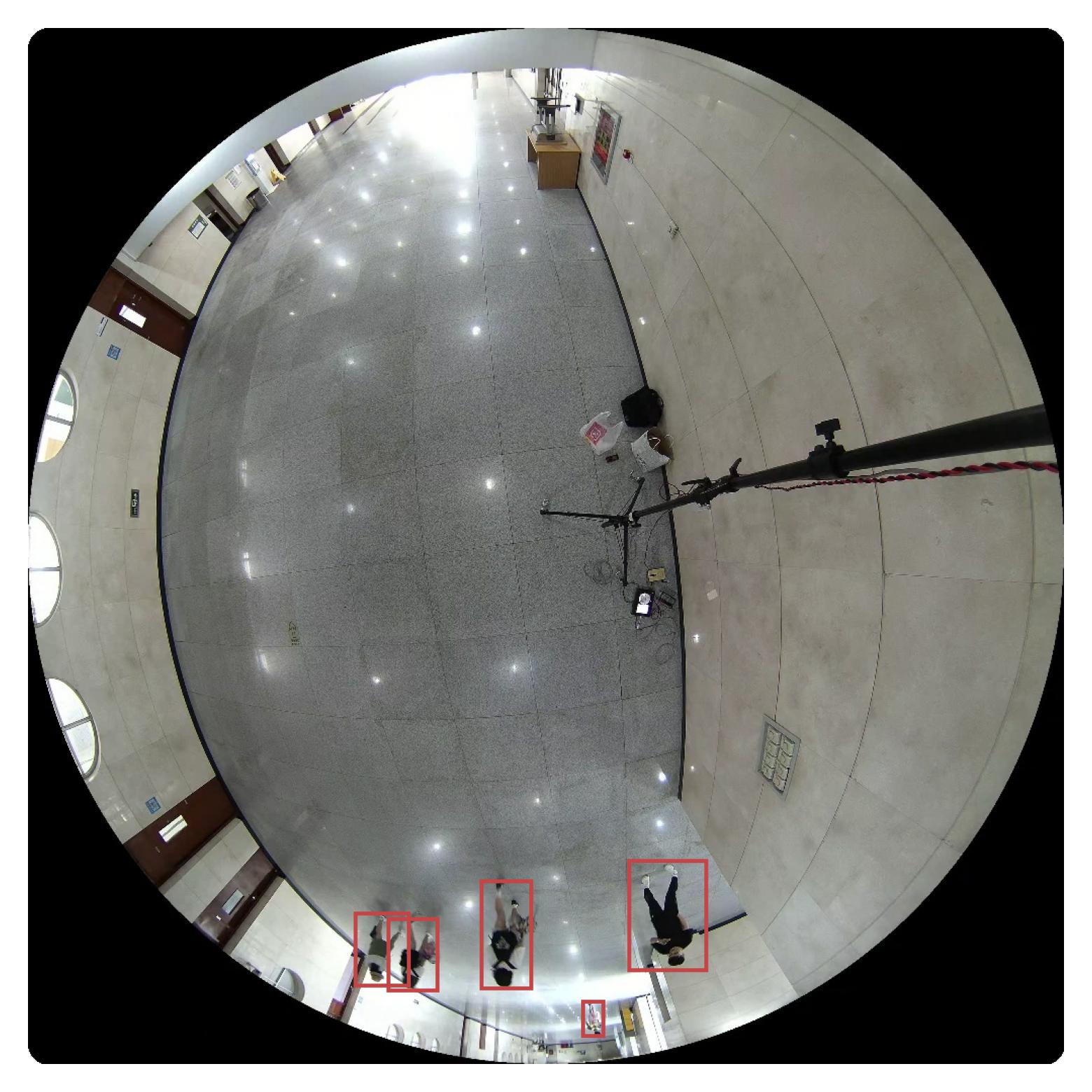} &
    \includegraphics[width=0.45\linewidth]{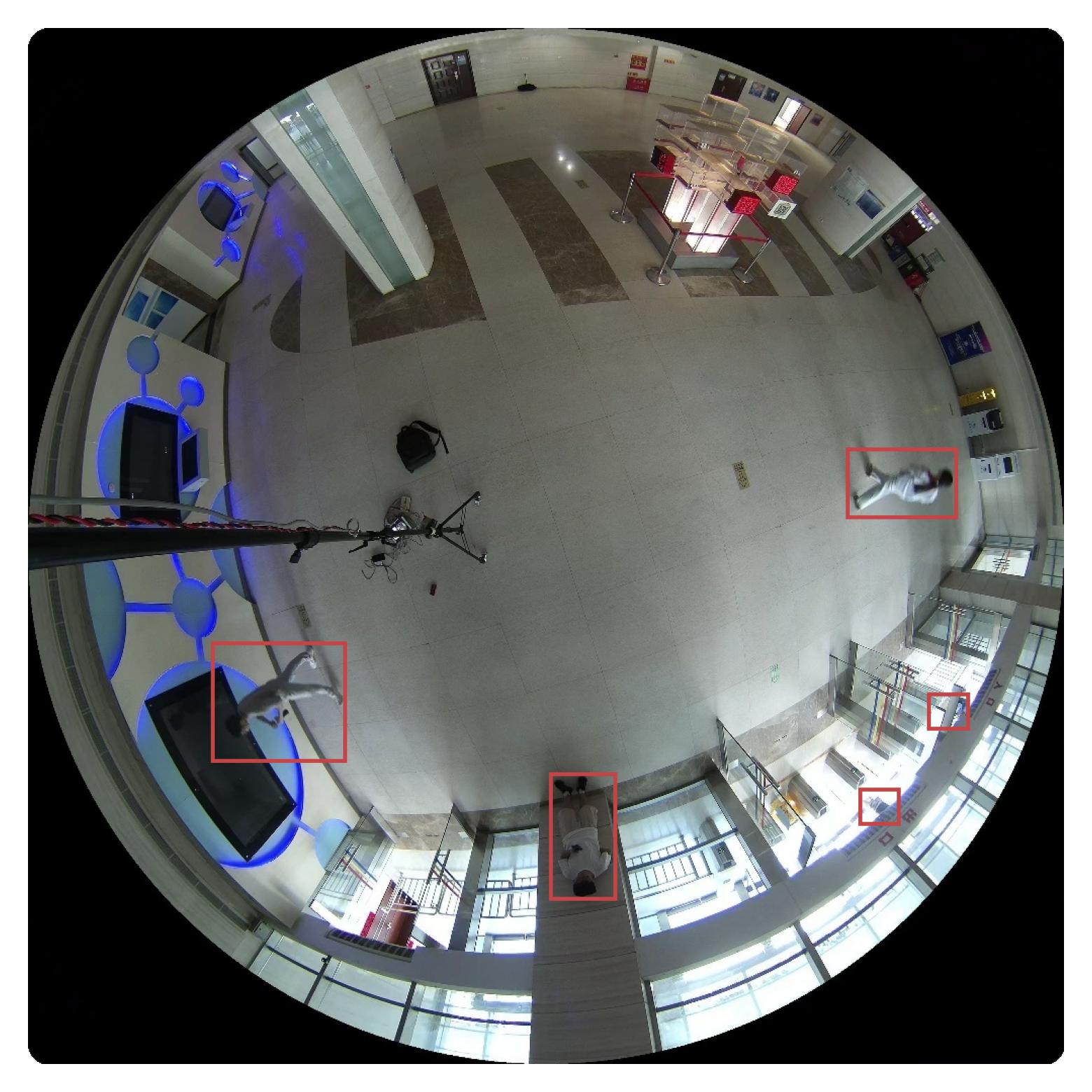} \\
    \includegraphics[width=0.45\linewidth]{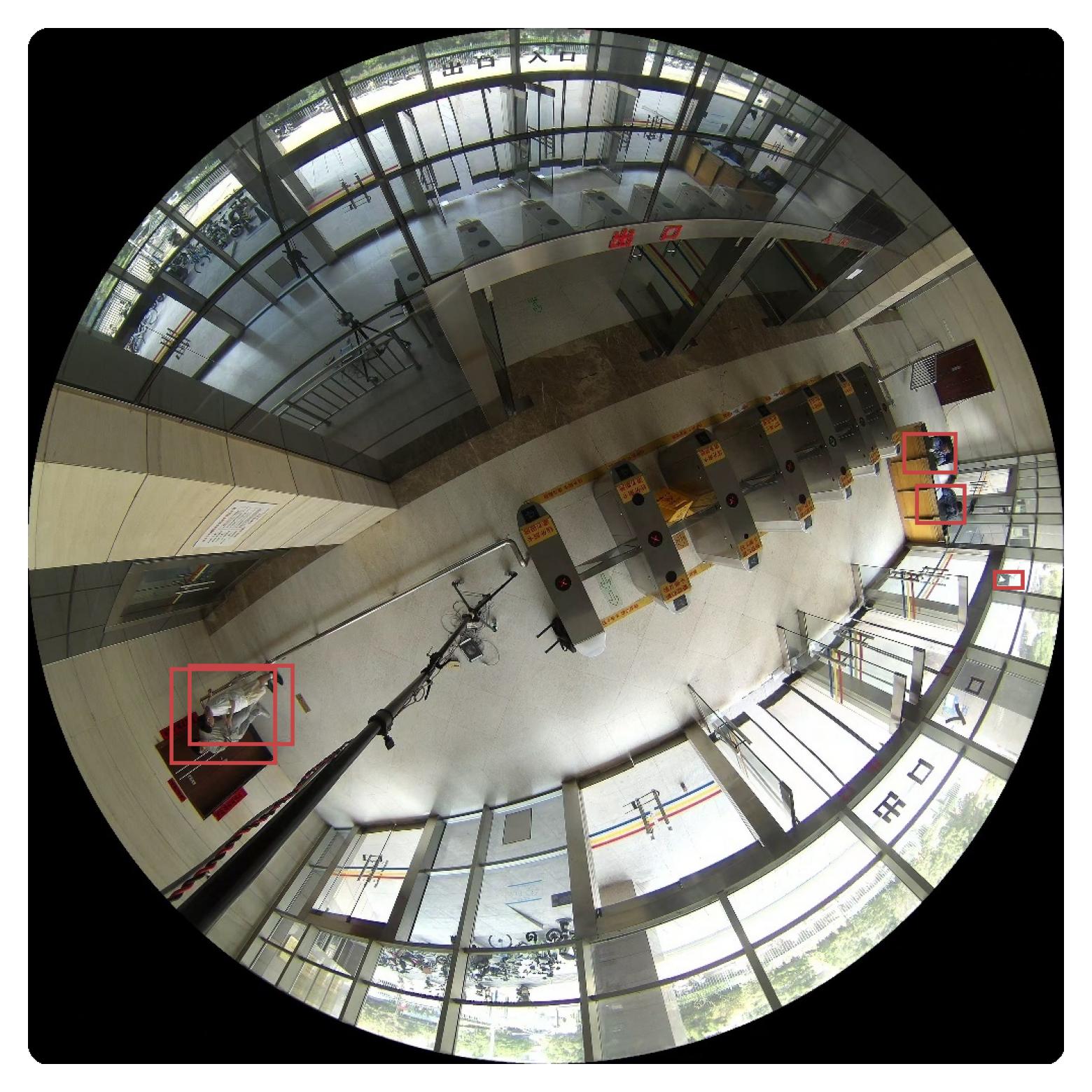} &
    \includegraphics[width=0.45\linewidth]{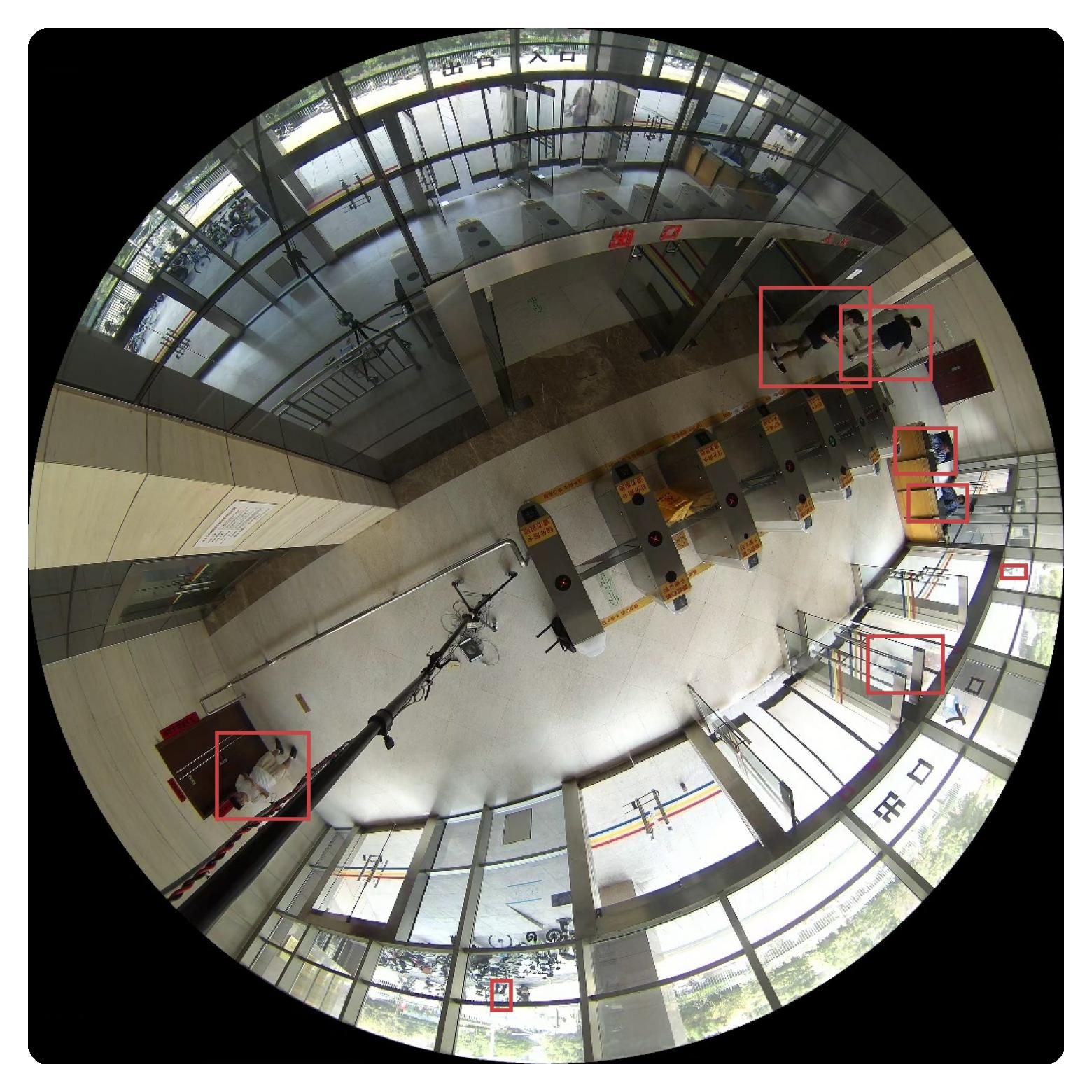} \\
  \end{tabular}
  \caption{Example images of the fisheye indoor task.}
  \label{fig:fisheye-indoor-task}
\end{figure}

\noindent\textbf{Dataset Processing.}
The dataset only provides rotated bounding boxes. We transform them into non-rotated bounding boxes using the given angle. However, this results in boxes that aren't tightly fitted around the object. To narrow the task, we manually choose only indoor scenes.
Example images are illustrated in Figure~\ref{fig:fisheye-indoor-task}.

\noindent\textbf{Train, Val, and Test Split.}
Based on the available scene IDs, the data is split into training, validation, and test like in Section~\ref{supp:tasks:nuimages} described.

\noindent\textbf{Labeling Policy}
The labeling policy corresponds with the reference in Section~\ref{supp:tasks:nuimages}.

\noindent\textbf{D-RICO Classes.}
The dataset only provides labels for the \textit{person} class.

\subsubsection{\texorpdfstring{Gated (DENSE~\cite{bijelic_seeing_2020})}{Gated (DENSE)}}\label{sec:supp:gated-task}
DENSE~\cite{bijelic_seeing_2020} is a multimodal dataset collected over 10,000 kilometers of driving, featuring data from gated cameras, LiDAR, and radar. It focuses on autonomous perception in adverse weather conditions like fog, rain, and snow.

\noindent\textbf{Dataset Processing.}
We select all non-inclement images from the dataset.
We further remove all images that include one of the following classes: \textit{DontCare}, \textit{LargeVehicle\_is\_group}, \textit{Vehicle\_is\_group}, \textit{PassengerCar\_is\_group}, \textit{RidableVehicle\_is\_group}, and \textit{Pedestrian\_is\_group}
This is because the objects of the group could be separated and therefore create contradictions.
We remove all objects smaller than $7\times7$, as such small objects are not labeled in other datasets.

\begin{figure}[t]
  \centering
  \begin{tabular}{cc}
    \includegraphics[width=0.45\linewidth]{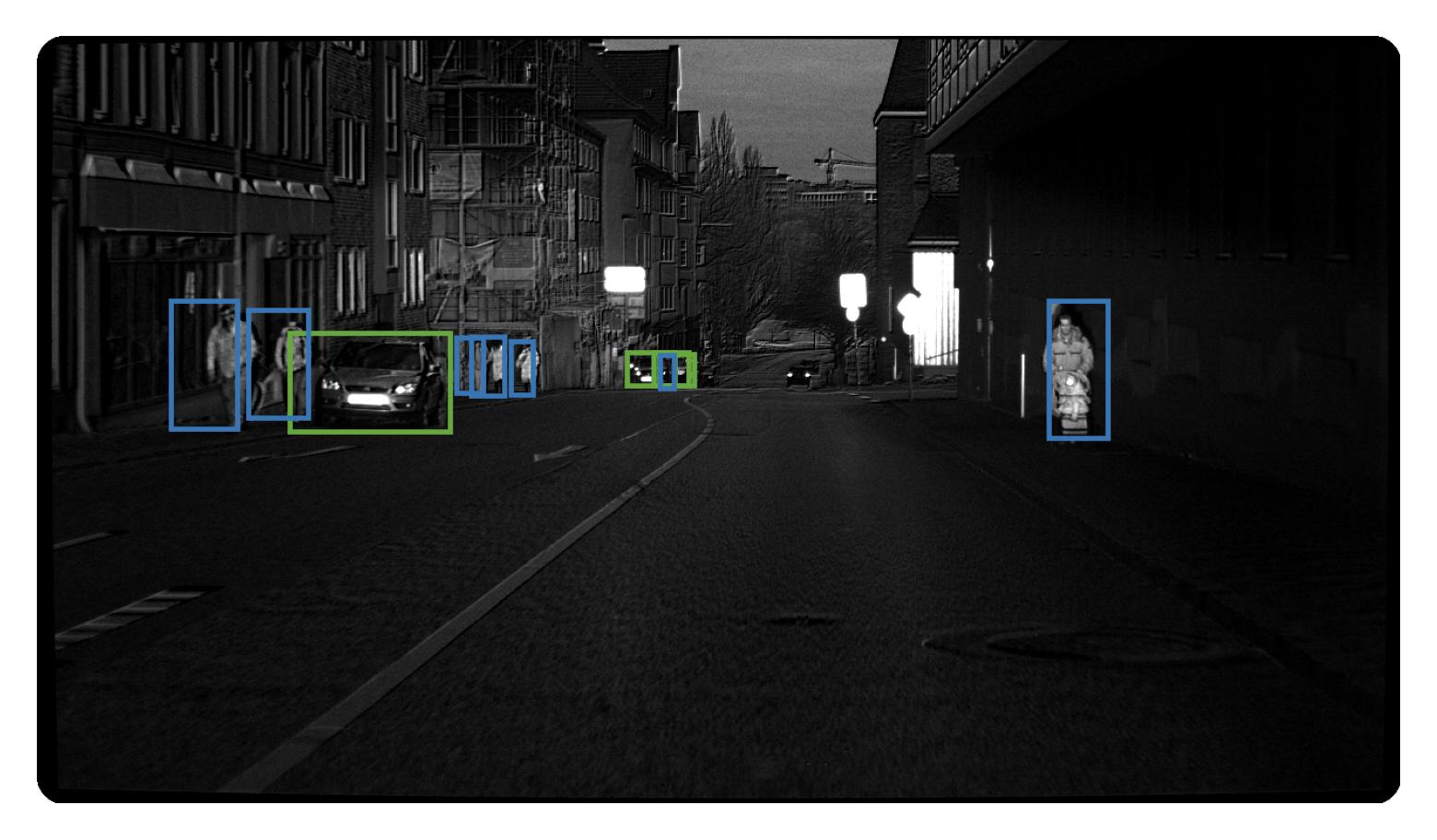} &
    \includegraphics[width=0.45\linewidth]{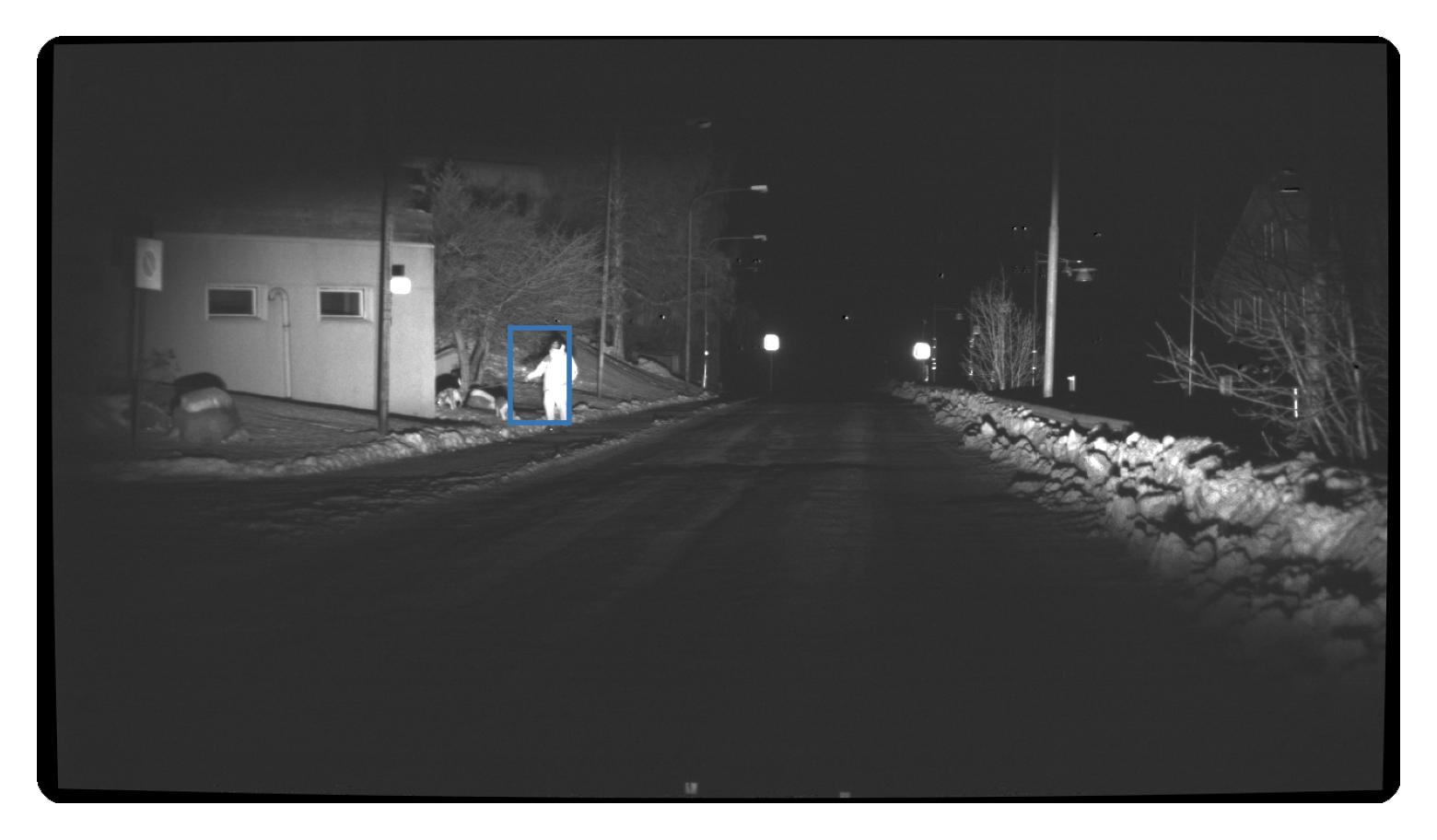} \\
    \includegraphics[width=0.45\linewidth]{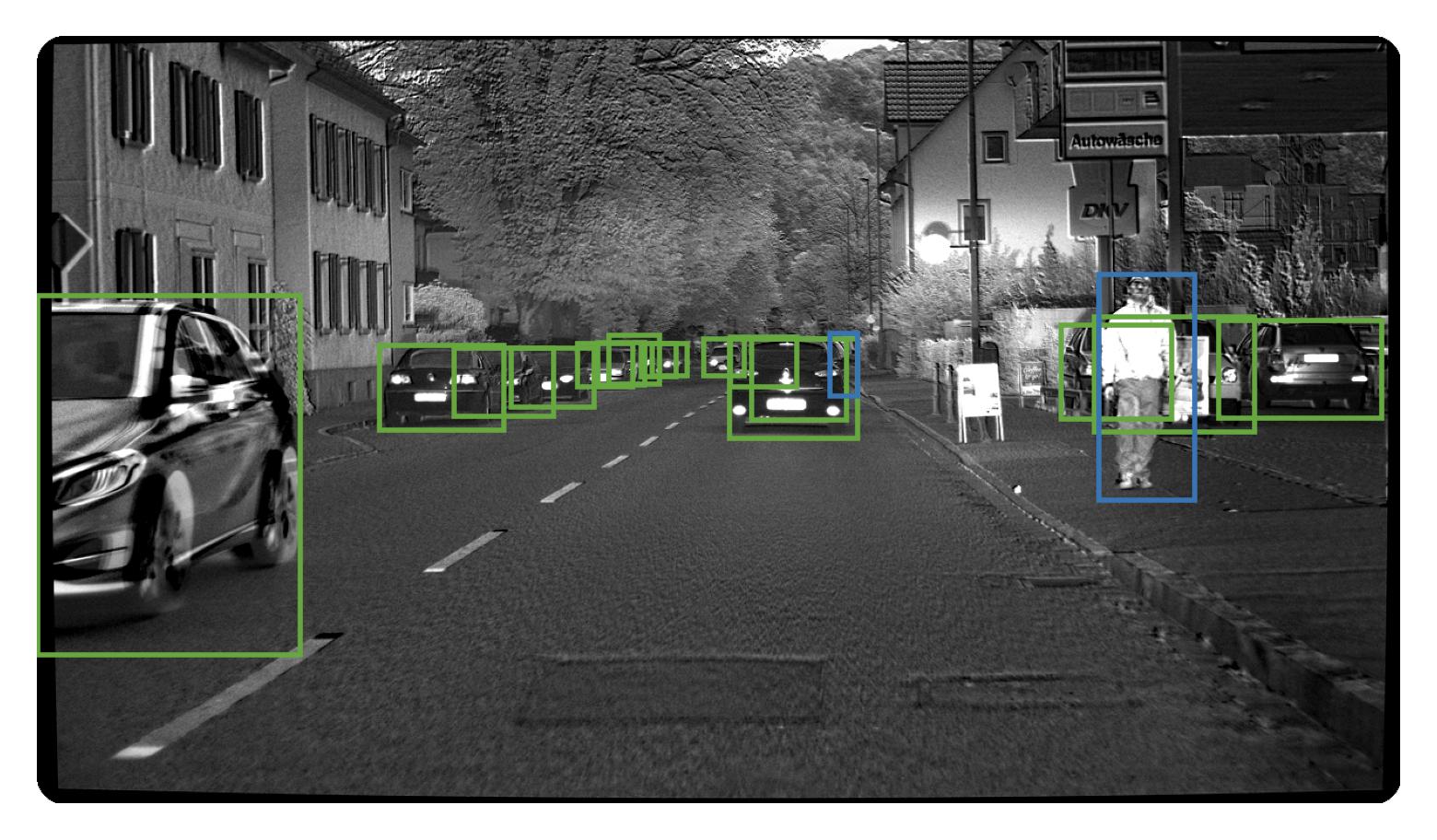} &
    \includegraphics[width=0.45\linewidth]{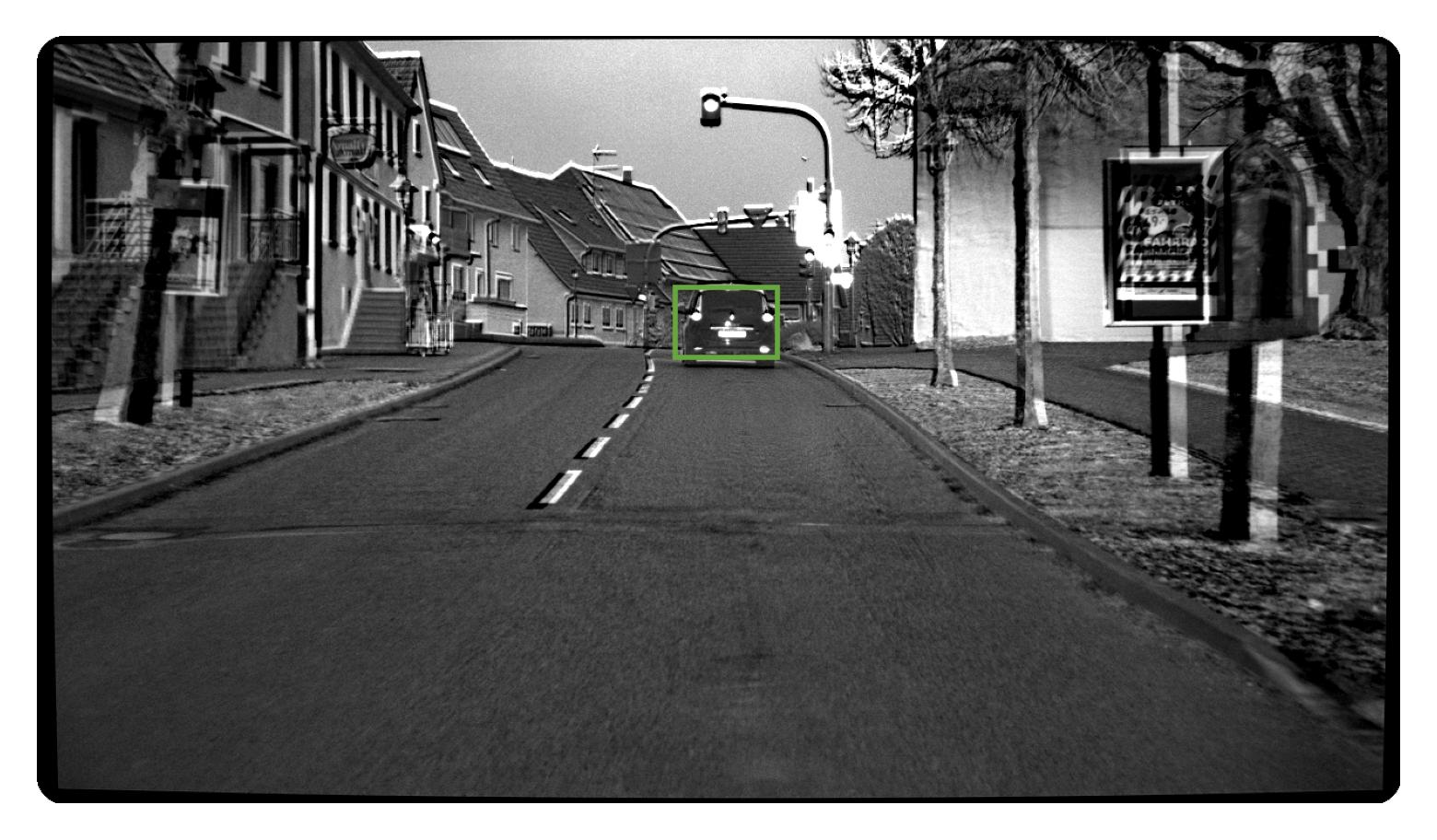} \\
  \end{tabular}
  \caption{Example images of the gated task.}
  \label{fig:gated-task}
\end{figure}

Figure~\ref{fig:gated-task} provides example images.

\noindent\textbf{Train, Val, and Test Split.}
The data is divided into training, validation, and test sets based on the available scene IDs, as outlined in Section~\ref{supp:tasks:nuimages}.

\noindent\textbf{Labeling Policy}
The labeling policy aligns with Section~\ref{supp:tasks:nuimages}.

\noindent\textbf{D-RICO Classes.}
\begin{itemize} \setlength{\itemindent}{0.5cm}
    \item \textbf{person:} \textit{Pedestrian},
    \item \textbf{vehicle:} \textit{PassengerCar}, \textit{Vehicle}, \textit{LargeVehicle}
\end{itemize}

\noindent\textbf{EC-RICO Classes.}

\begin{itemize} \setlength{\itemindent}{0.5cm}
    \item \textbf{person:} \textit{Pedestrian}
    \item \textbf{car:} \textit{PassengerCar}
\end{itemize}

\subsubsection{\texorpdfstring{Realistic Simulation (Synscapes~\cite{wrenninge_synscapes_2018})}{Realistic Simulation (Synscapes)}}
Synscapes~\cite{wrenninge_synscapes_2018} is a synthetic dataset of 25,000 photorealistic urban driving images created using physically based rendering techniques. It provides per-pixel semantic segmentation and instance annotations, which are designed for training perception models with realistic lighting and material properties.

\begin{figure}[t]
  \centering
  \begin{tabular}{cc}
    \includegraphics[width=0.45\linewidth]{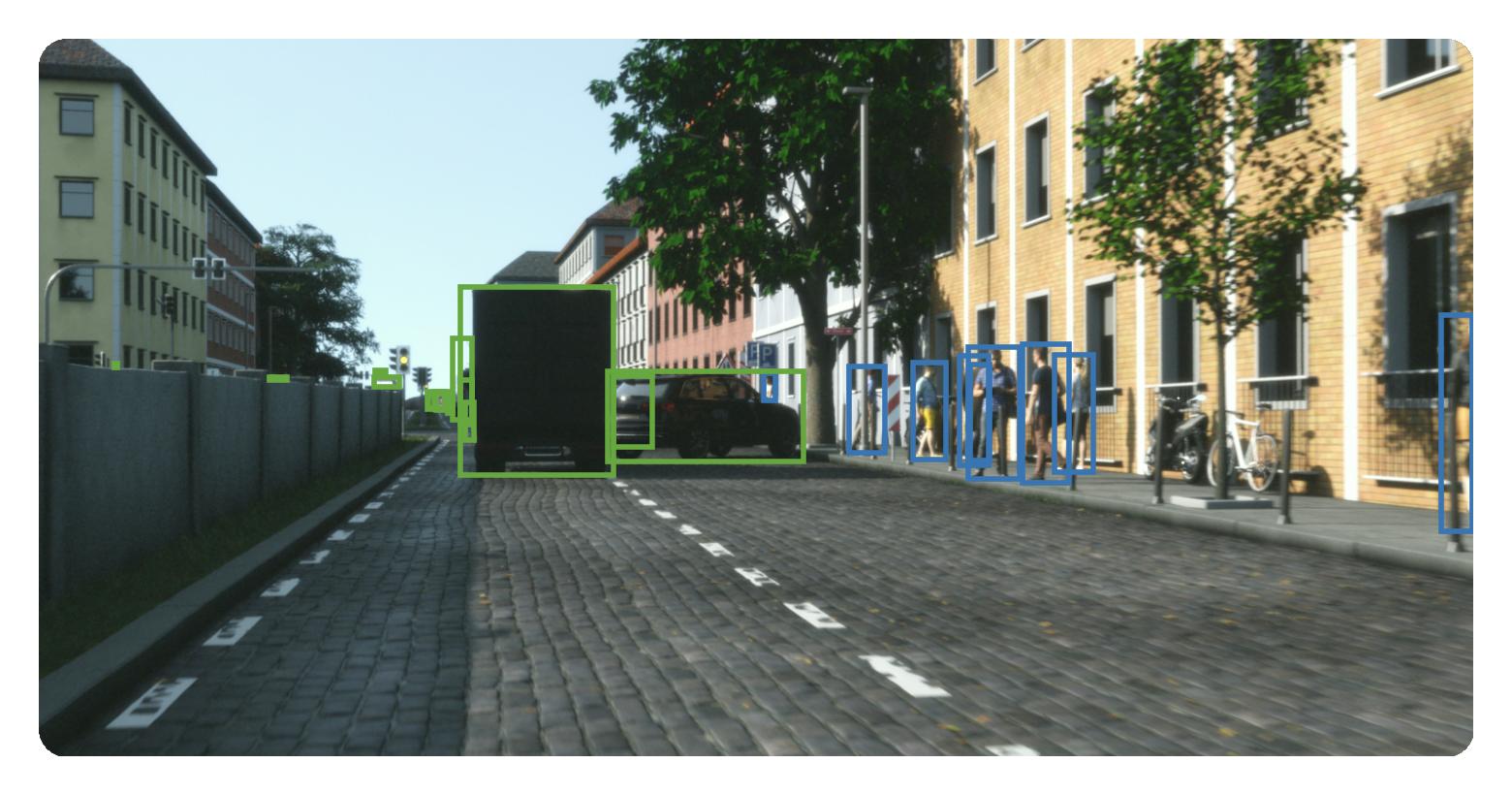} &
    \includegraphics[width=0.45\linewidth]{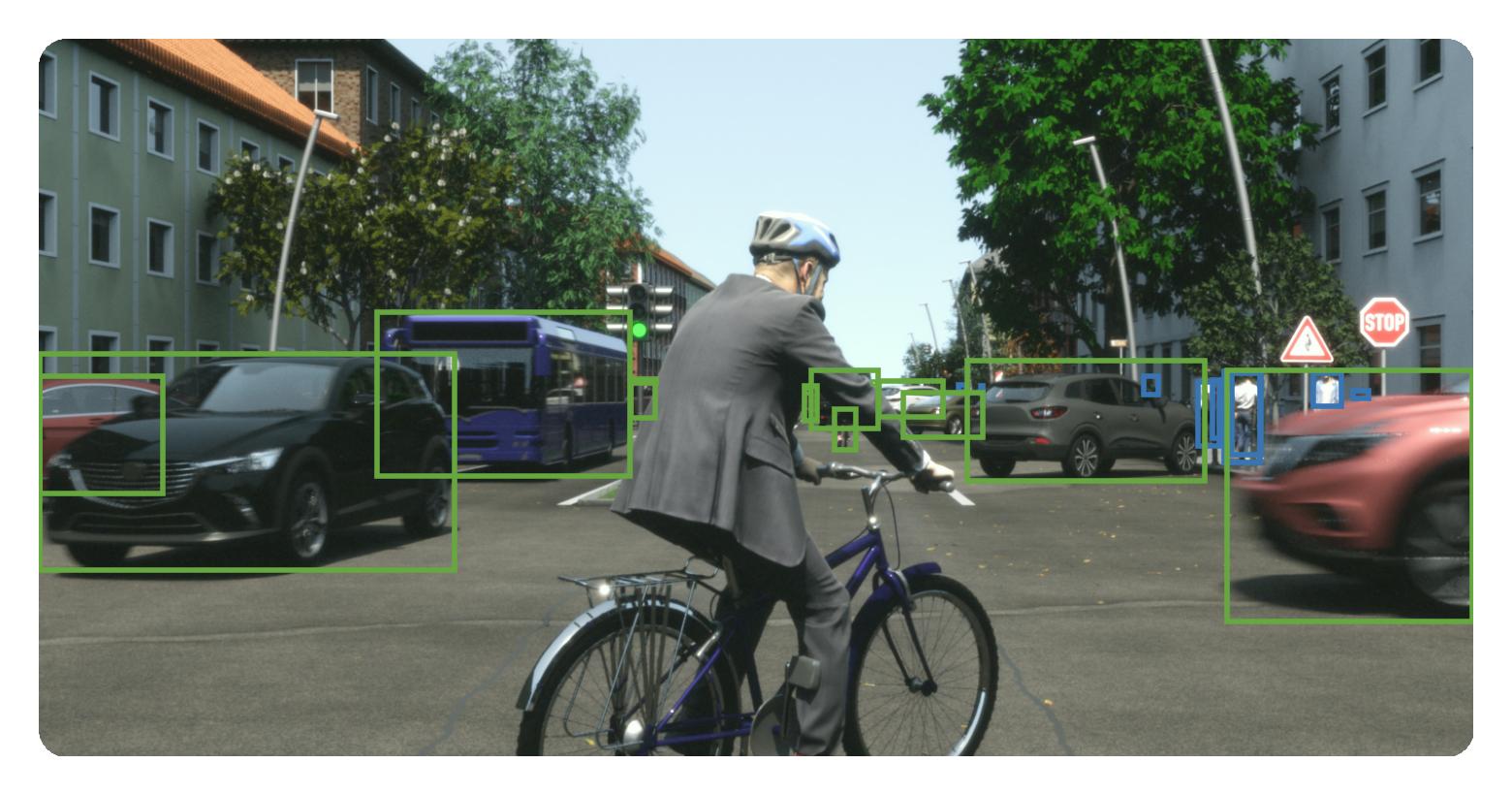} \\
    \includegraphics[width=0.45\linewidth]{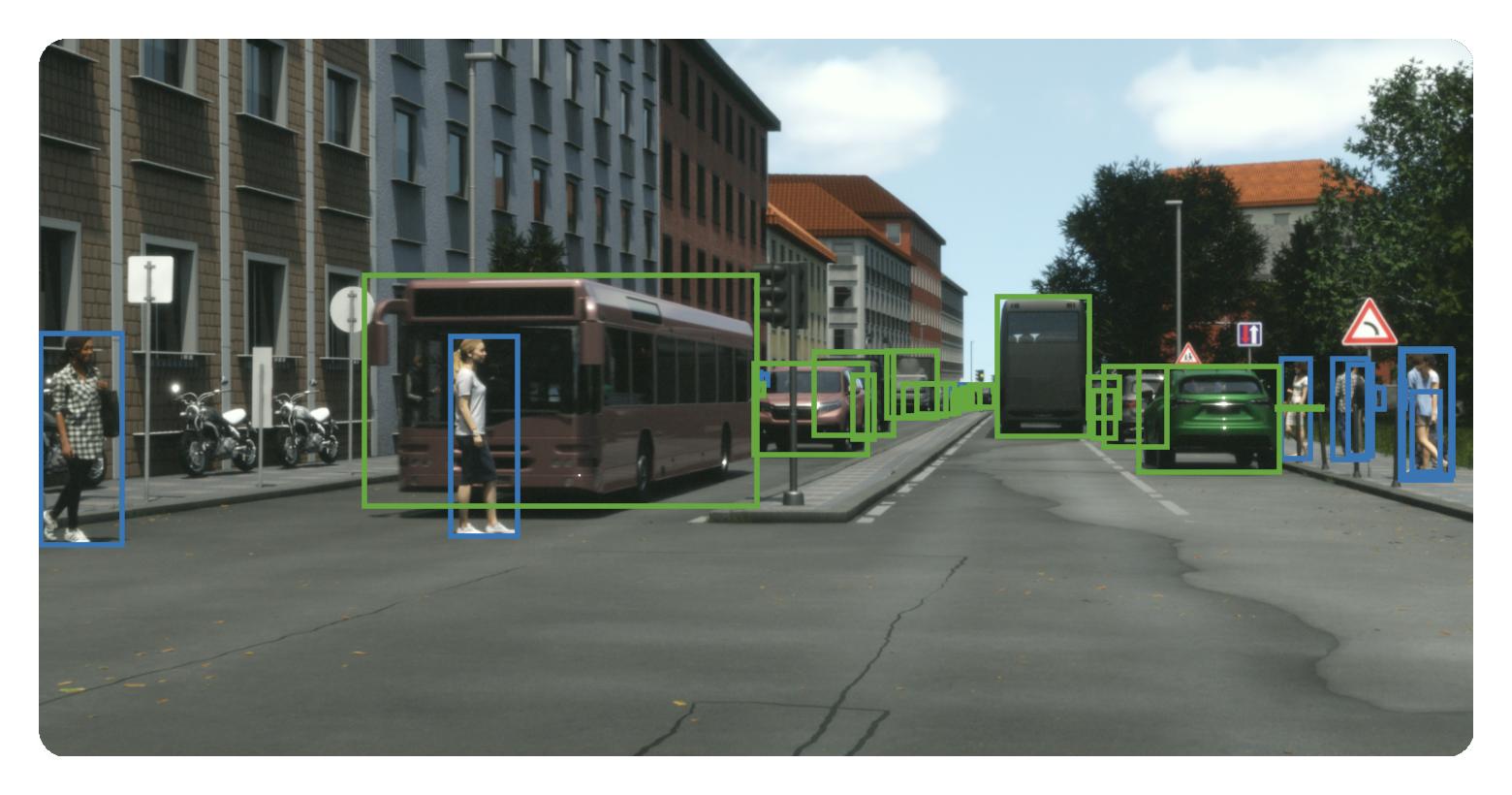} &
    \includegraphics[width=0.45\linewidth]{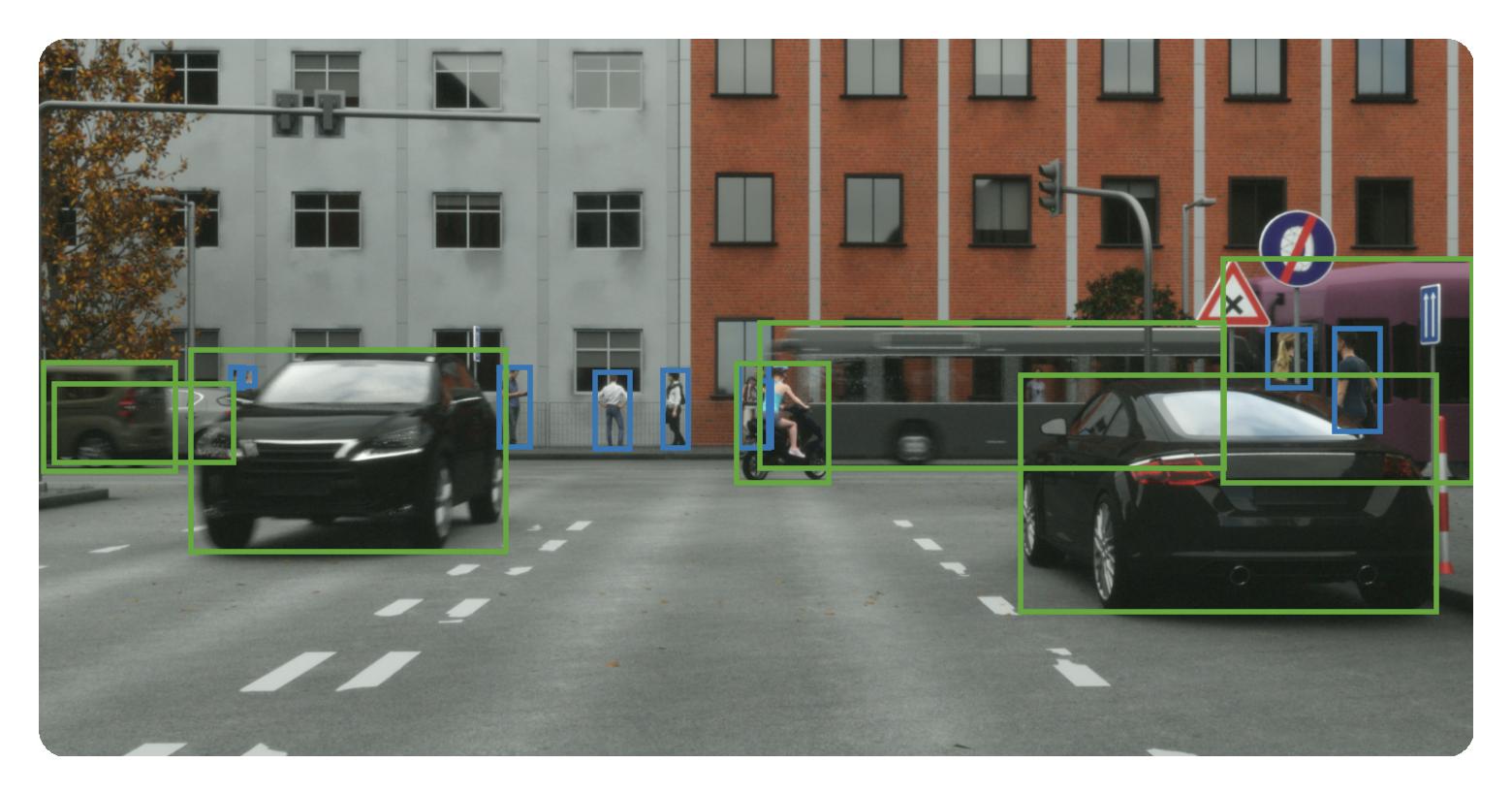} \\
  \end{tabular}
  \caption{Example images of the photorealistic simulation task.}
  \label{fig:photorealistic-simulation-task}
\end{figure}

\noindent\textbf{Dataset Processing.}
The dataset provides bounding boxes; however, these also include labels for objects that are completely hidden. We calculate the bounding boxes for each visible object based on the semantic and instance segmentation masks. We only select images with mean gray values above 0.38 to exclude low visibility and night images (see Section~\ref{sec:supp:fisheye-task}). Refer to Figure~\ref{fig:photorealistic-simulation-task} for examples.

\noindent\textbf{Train, Val, and Test Split.}
Each image is rendered from a unique scene, and the images can, therefore, be randomly assigned into training, validation, and test splits.

\noindent\textbf{Labeling Policy}
Multiple objects are not labeled, and some bicycles are labeled as vehicles. 

\noindent\textbf{D-RICO Classes.}
\begin{itemize} \setlength{\itemindent}{0.5cm}
    \item \textbf{person:} \textit{person},
    \item \textbf{vehicle:} \textit{car}, \textit{truck}, \textit{bus}, \textit{train}, \textit{motorcycle}, \textit{rider}
\end{itemize}

\subsubsection{\texorpdfstring{Thermal Fisheye Indoor (TIMo~\cite{schneider_timodataset_2022})}{Thermal Fisheye Indoor (TIMo)}}
TIMo~\cite{schneider_timodataset_2022} is a thermal imaging dataset designed for indoor person detection, comprising over 612,000 frames captured with infrared cameras. It features detailed annotations for tracking individuals and detecting anomalies in low-visibility environments.

\begin{figure}[t]
  \centering
  \begin{tabular}{cc}
    \includegraphics[width=0.45\linewidth]{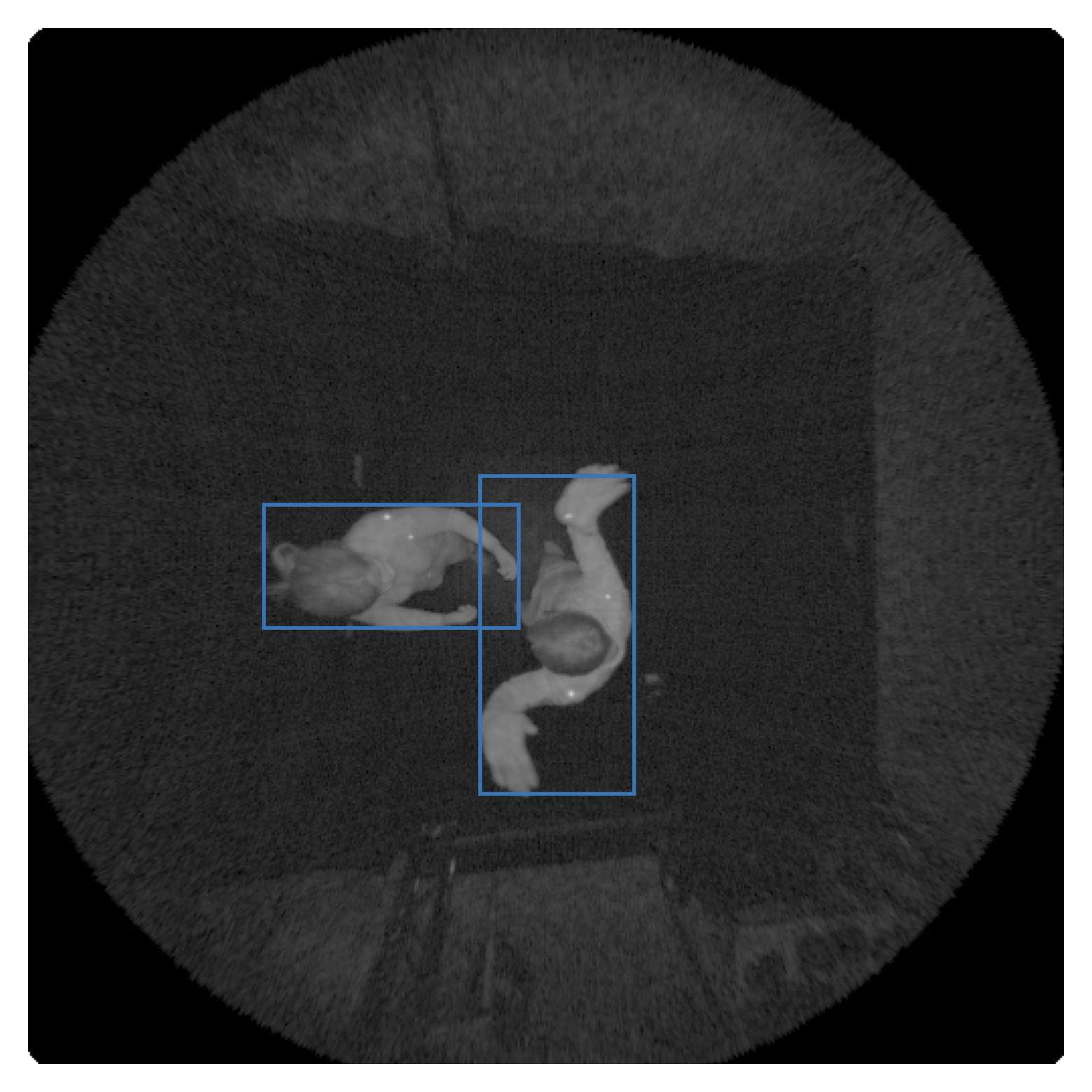} &
    \includegraphics[width=0.45\linewidth]{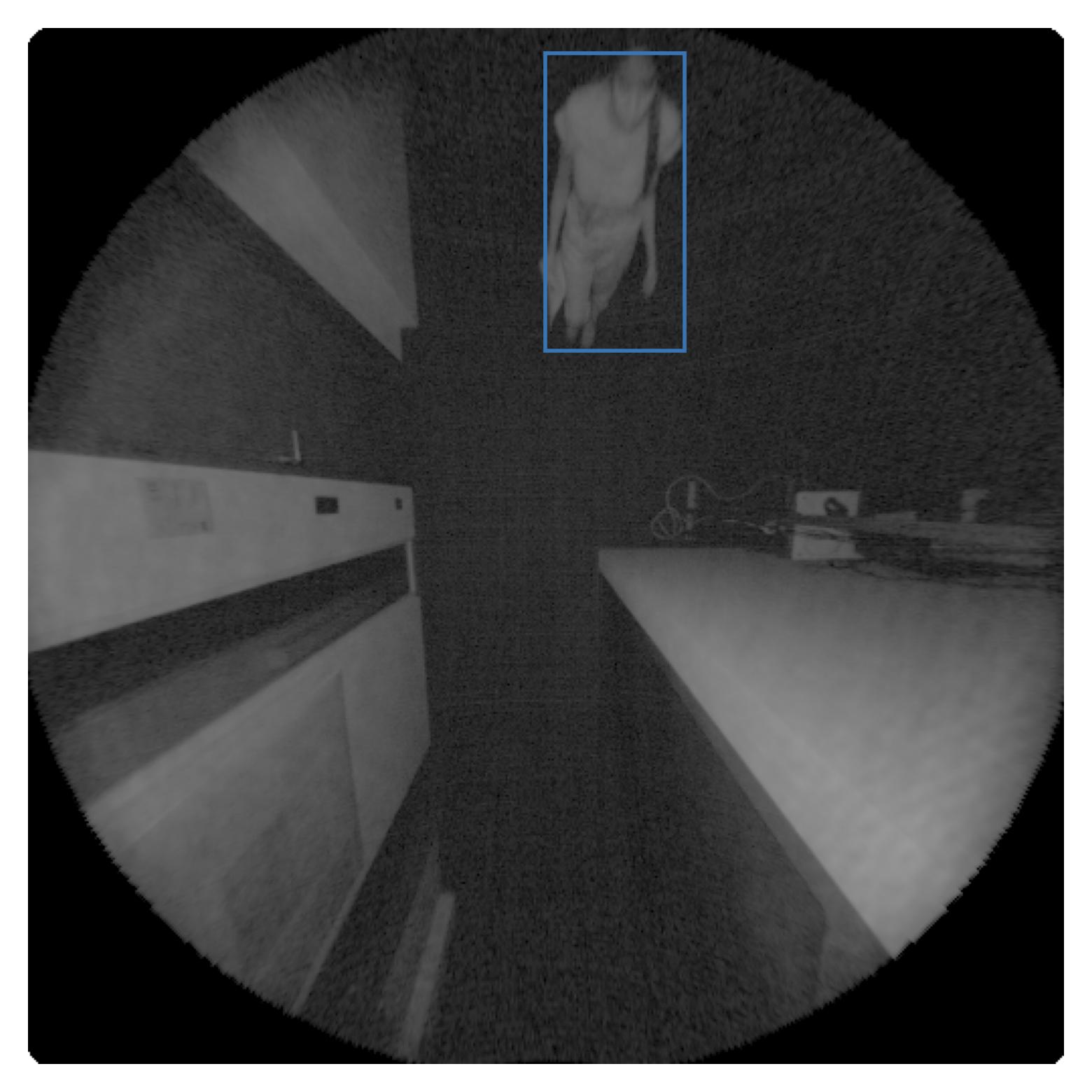} \\
    \includegraphics[width=0.45\linewidth]{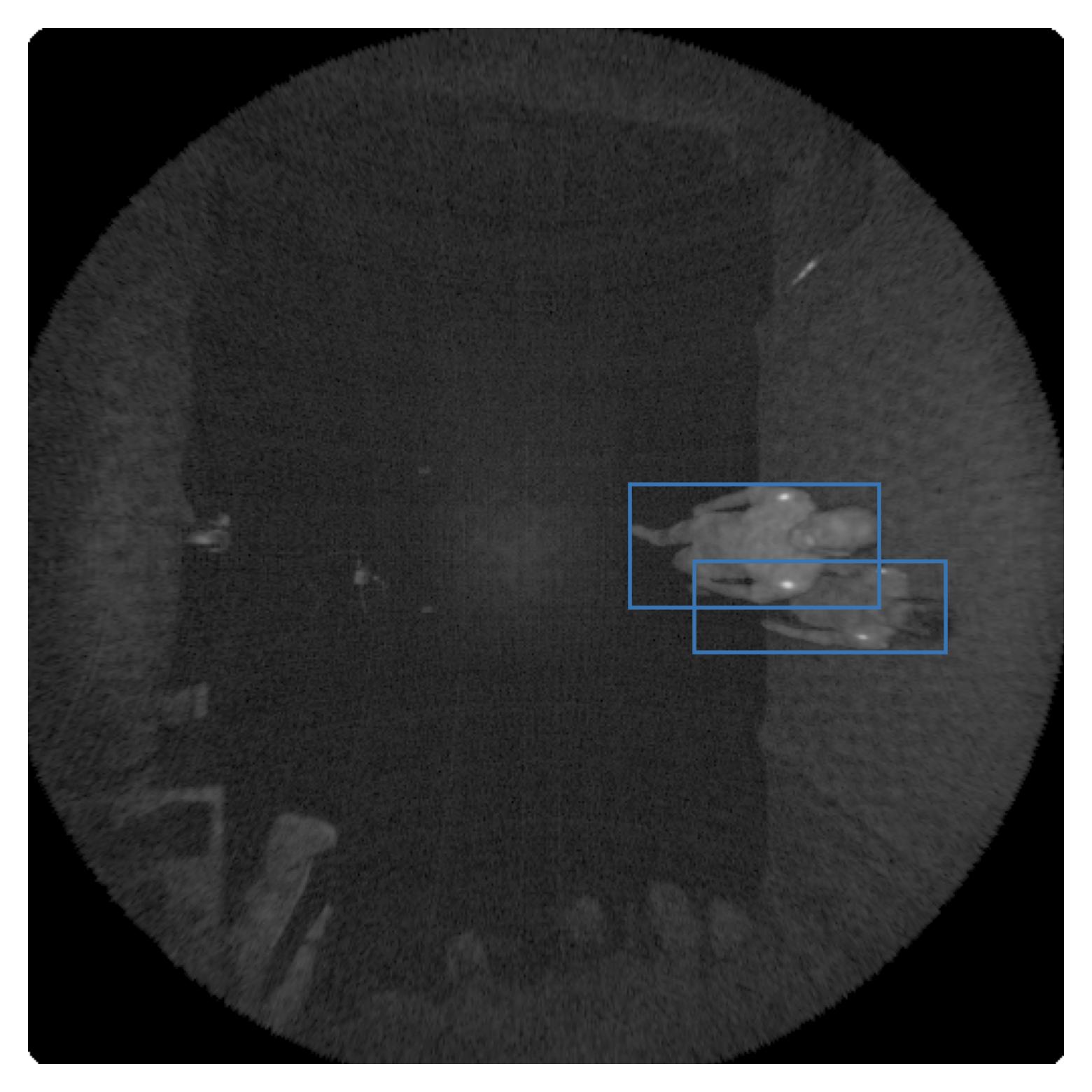} &
    \includegraphics[width=0.45\linewidth]{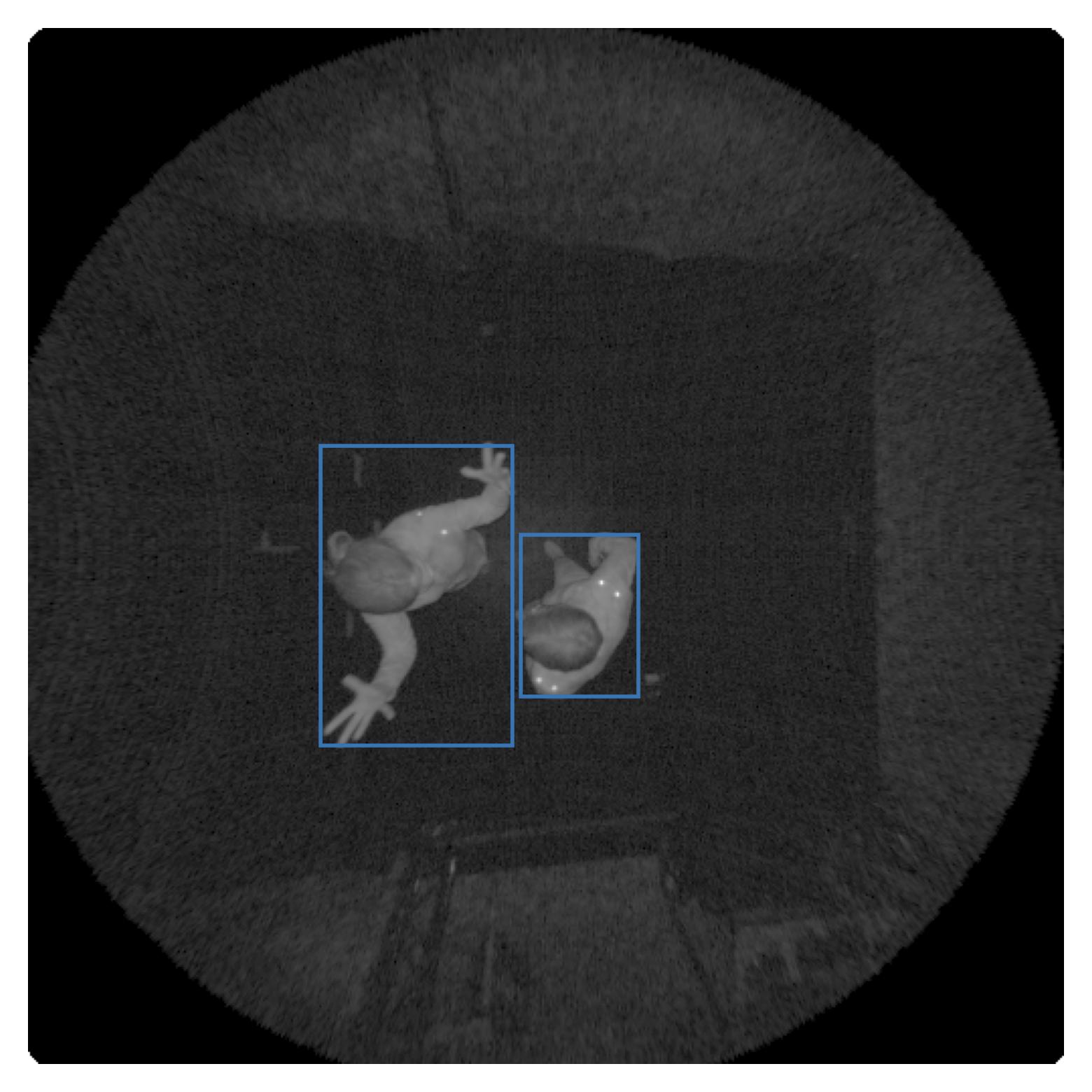} \\
  \end{tabular}
  \caption{Example images of the thermal fisheye indoor task.}
  \label{fig:thermal-fisheye-indoor-task}
\end{figure}

\noindent\textbf{Dataset Processing.}
We use only the infrared images from the dataset and keep every sixth frame. The images provided in the dataset are in signal strength. We calculate the logarithm of the images and normalize them.
Examples are displayed in Figure~\ref{fig:thermal-fisheye-indoor-task}.

\noindent\textbf{Train, Val, and Test Split.}
The dataset is separated into training, validation, and test sets according to the scene IDs as detailed in Section~\ref{supp:tasks:nuimages}.

\noindent\textbf{Labeling Policy}
The labeling policy is consistent with Section~\ref{supp:tasks:nuimages}.

\noindent\textbf{D-RICO Classes.}
The dataset exclusively offers labels for the \textit{person} class.

\subsubsection{\texorpdfstring{Inclement (DENSE~\cite{bijelic_seeing_2020})}{Inclement (DENSE)}}
This subset of DENSE~\cite{bijelic_seeing_2020} includes images taken in non-clear weather conditions like heavy fog, snow, and rain. It offers a challenging testbed for assessing robustness in perception systems under limited visibility.

\begin{figure}[t]
  \centering
  \begin{tabular}{cc}
    \includegraphics[width=0.45\linewidth]{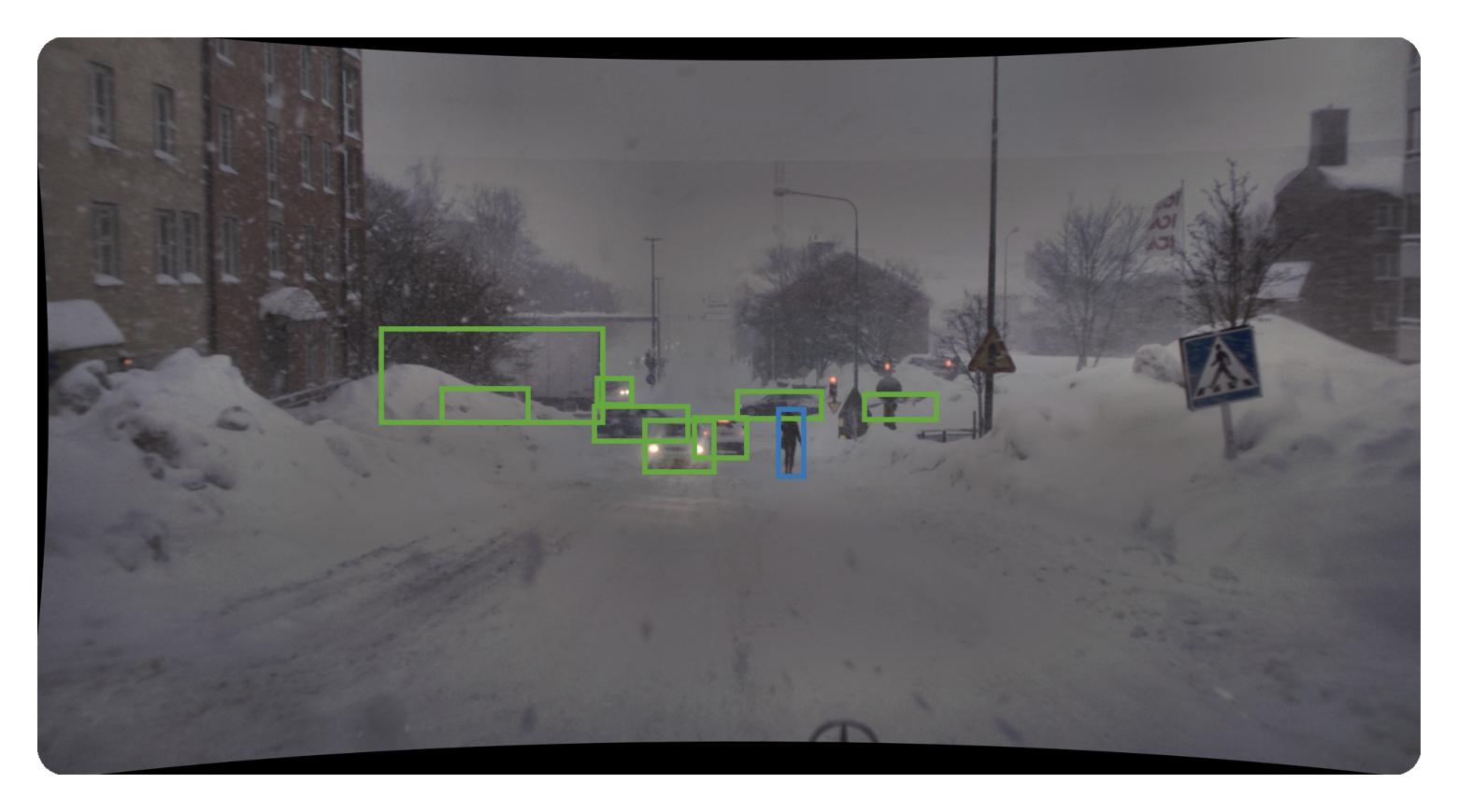} &
    \includegraphics[width=0.45\linewidth]{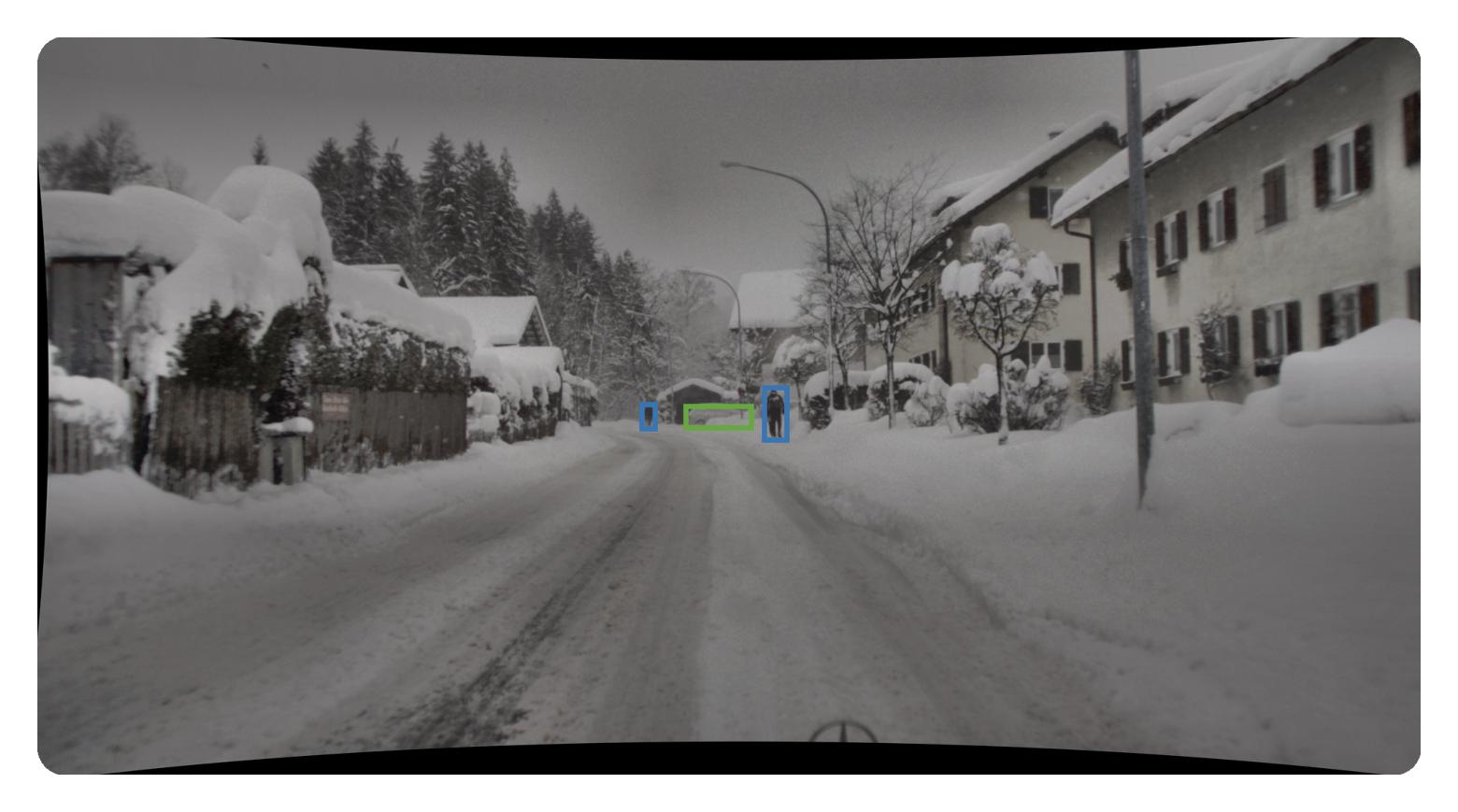} \\
    \includegraphics[width=0.45\linewidth]{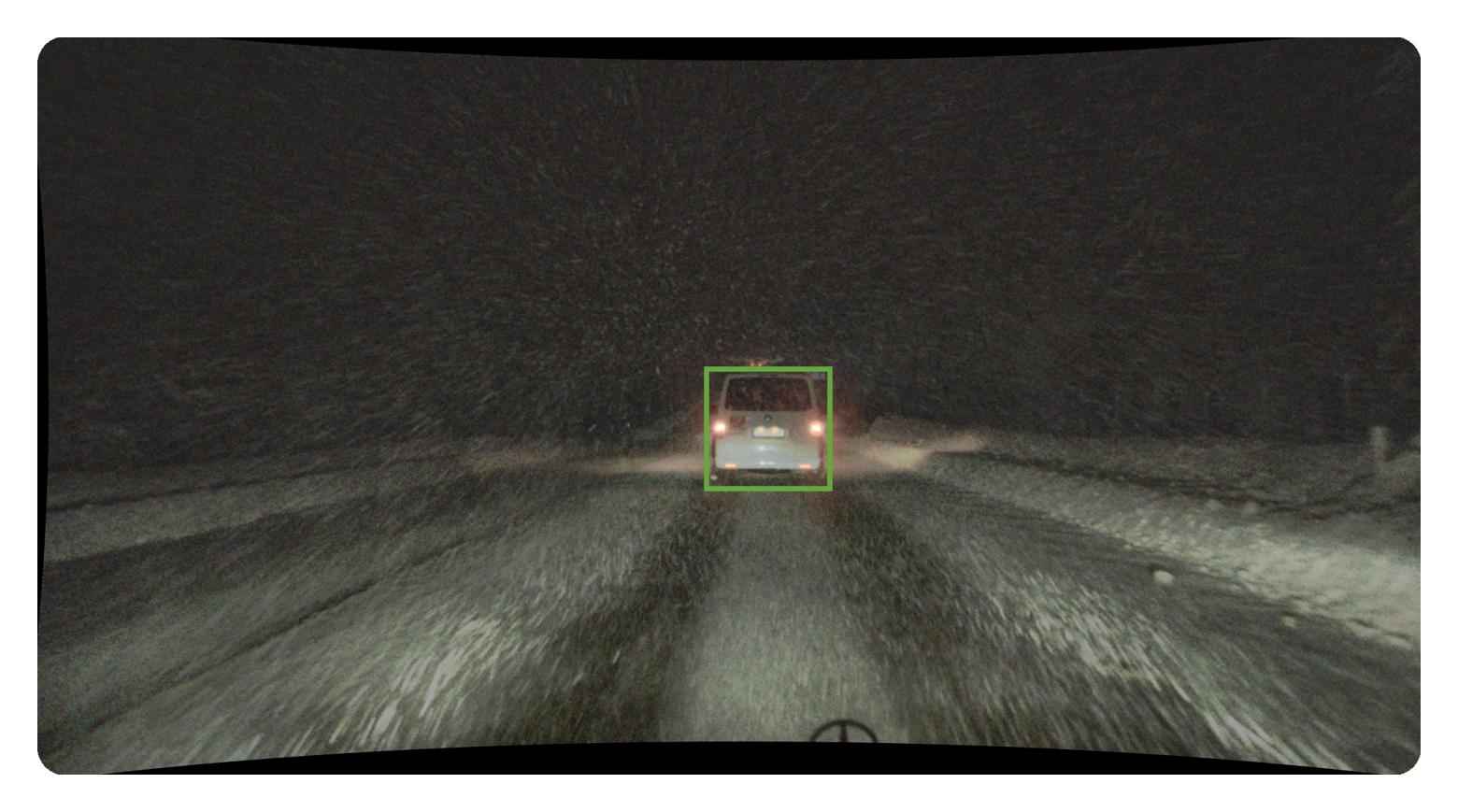} &
    \includegraphics[width=0.45\linewidth]{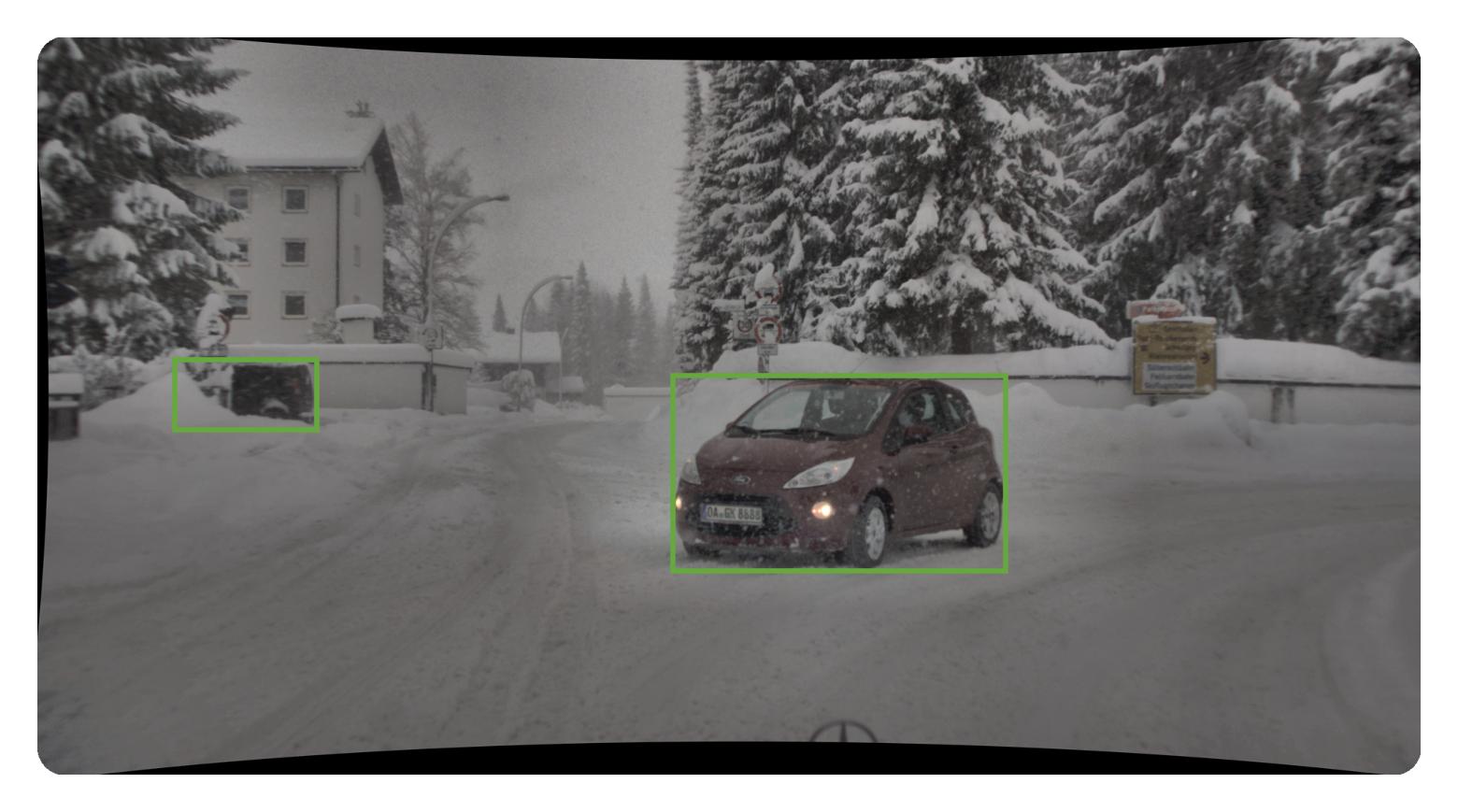} \\
  \end{tabular}
  \caption{Example images of the inclement task.}
  \label{fig:inclement-task}
\end{figure}

\noindent\textbf{Dataset Processing.}
We manually sort each image according to the weather conditions. We select those that depict intense fog, snow, and heavy rain. Additionally, we eliminate small objects and images based on the group classes, as described in Section~\ref{sec:supp:gated-task}.
Example images are shown in Figure~\ref{fig:inclement-task}.

\noindent\textbf{Train, Val, and Test Split.}
The dataset is divided into training, validation, and test sets based on the scene IDs outlined in Section~\ref{supp:tasks:nuimages}.

\noindent\textbf{Labeling Policy}
The labeling policy is consistent with Section~\ref{supp:tasks:nuimages}.

\noindent\textbf{D-RICO Classes.}
\begin{itemize} \setlength{\itemindent}{0.5cm}
    \item \textbf{person:} \textit{Pedestrian},
    \item \textbf{vehicle:} \textit{PassengerCar}, \textit{Vehicle}, \textit{LargeVehicle}
\end{itemize}

\subsubsection{\texorpdfstring{Event Camera (DSEC~\cite{gehrig_low-latency_2024, gehrig_dsec_2021})}{Event Camera (DSEC)}}
DSEC~\cite{gehrig_low-latency_2024, gehrig_dsec_2021} is a multimodal dataset that features high-resolution stereo event cameras, as well as RGB and LiDAR data. It captures 53 driving sequences in urban and rural settings, facilitating research on event-based perception in dynamic environments.

\begin{figure}[t]
  \centering
  \begin{tabular}{cc}
    \includegraphics[width=0.45\linewidth]{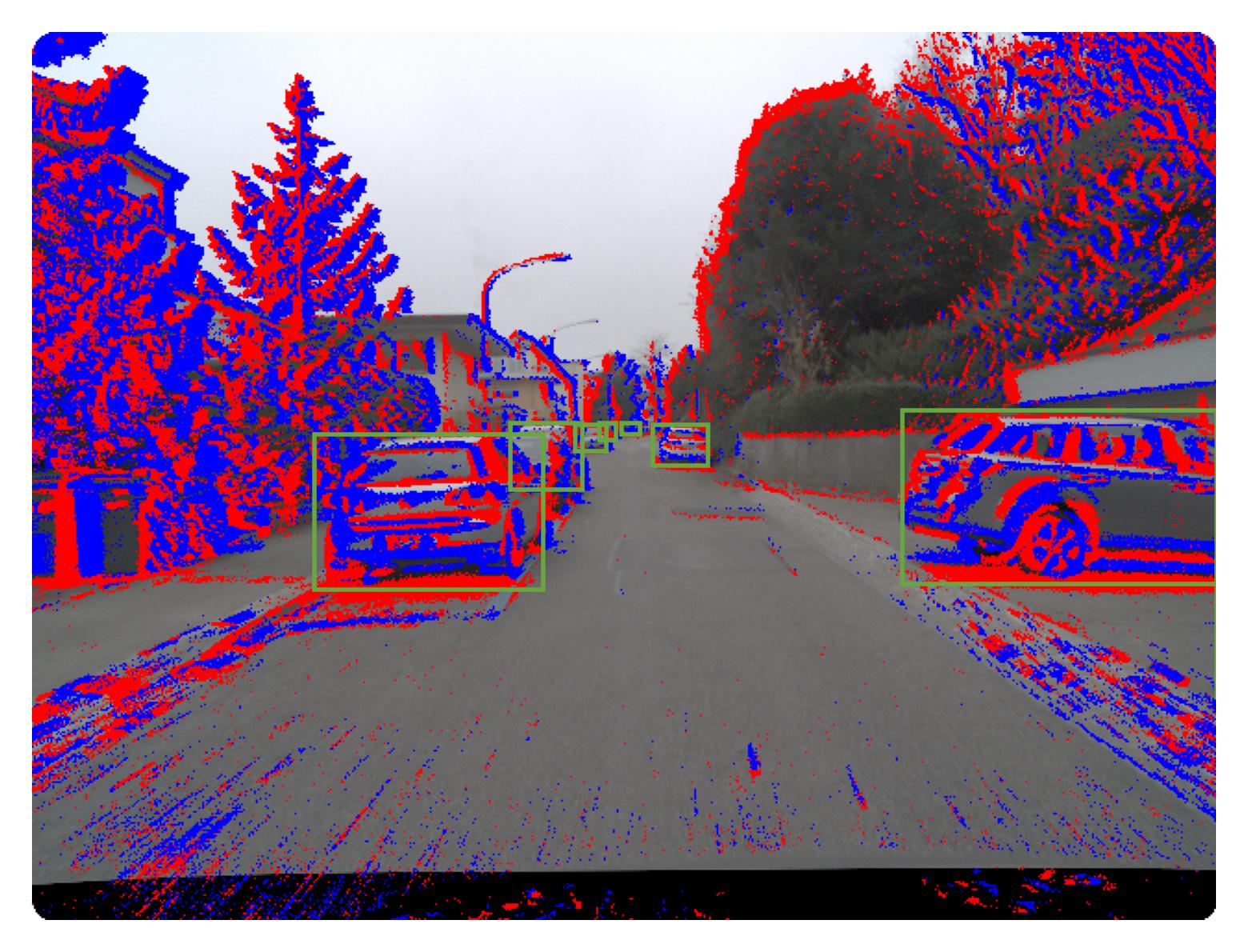} &
    \includegraphics[width=0.45\linewidth]{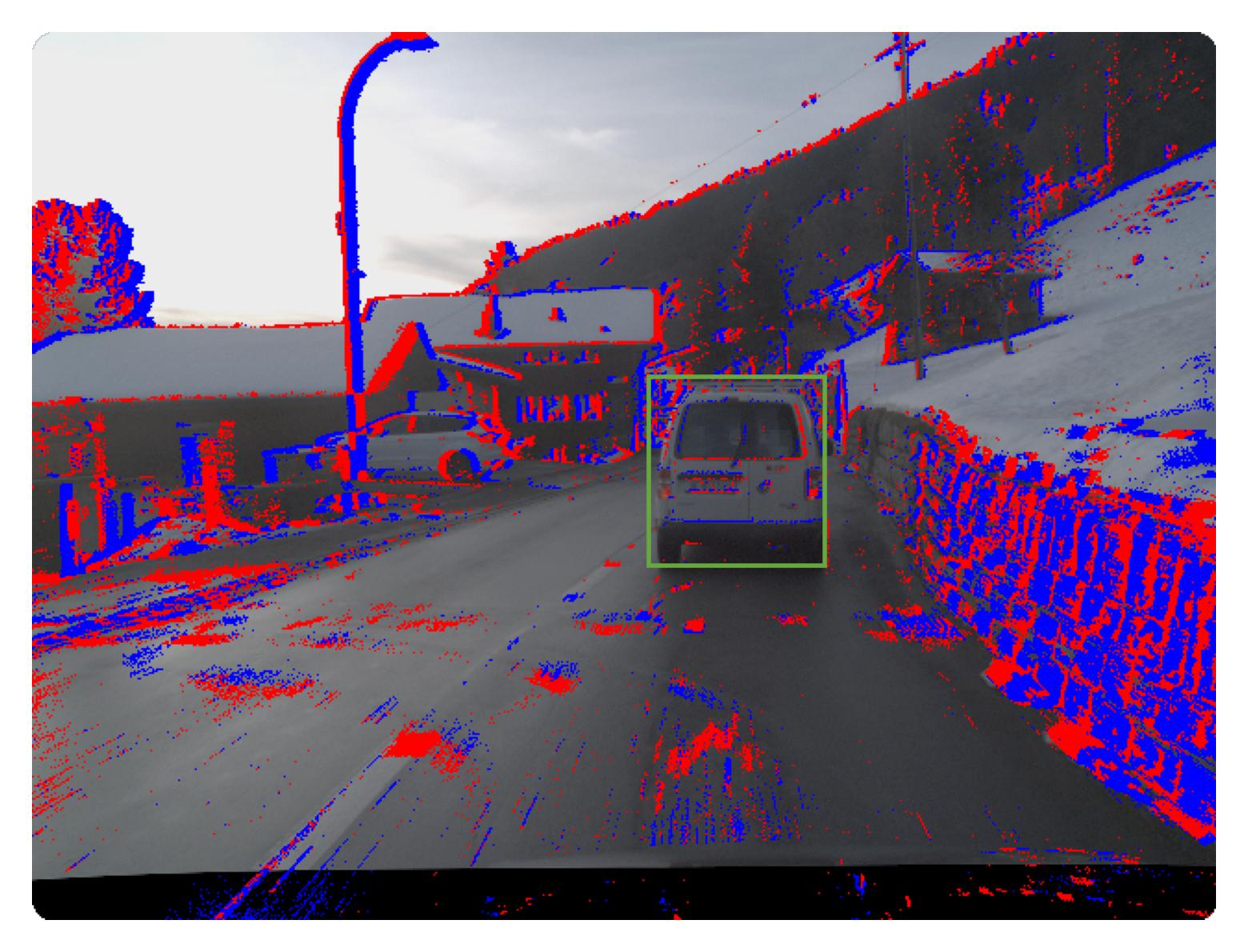} \\
    \includegraphics[width=0.45\linewidth]{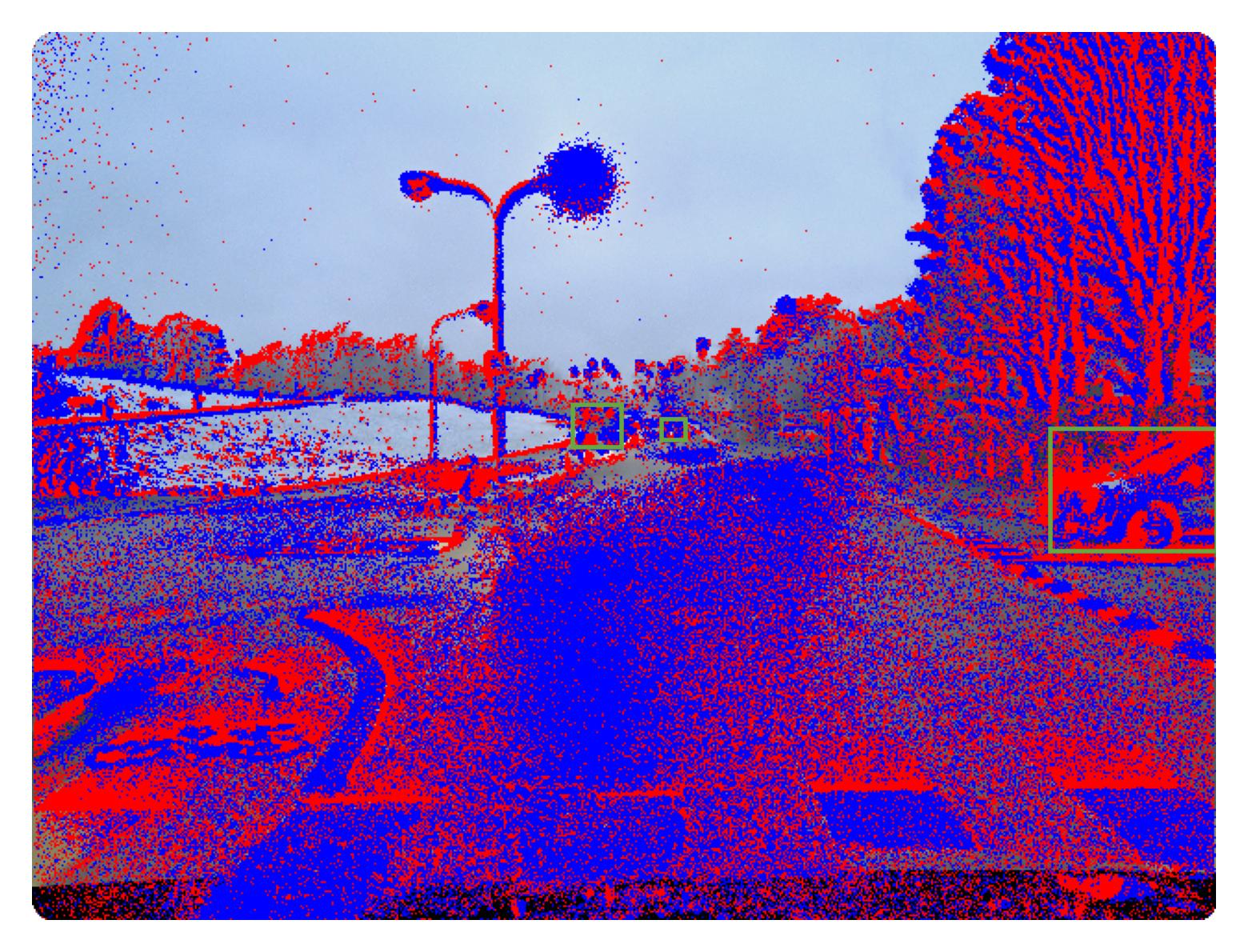} &
    \includegraphics[width=0.45\linewidth]{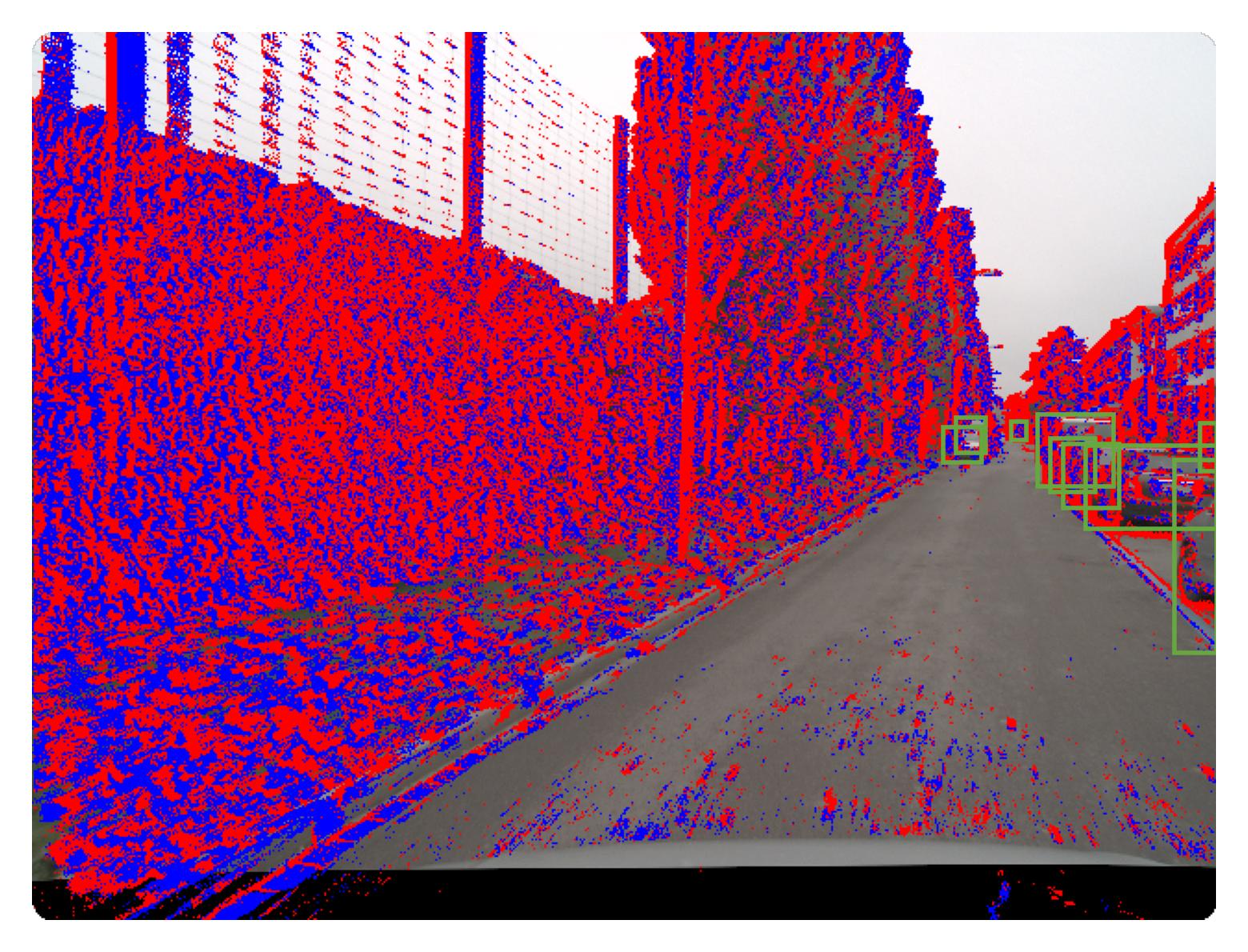} \\
  \end{tabular}
  \caption{Example images of the event camera task.}
  \label{fig:event-camera-task}
\end{figure}

\noindent\textbf{Dataset Processing.}
The gated and RGB images are overlaid according to the provided algorithm~\cite{gehrig_low-latency_2024, gehrig_dsec_2021}. Since this creates bounding boxes that extend beyond the images, we adjust them to fit the image size. As detailed in Section~\ref{sec:supp:thermal-task}, the rider and bicycle are combined. Figure~\ref{fig:event-camera-task} showcases example images.

\noindent\textbf{Train, Val, and Test Split.}
Since the dataset lacks scene IDs, we divided it as outlined in Section~\ref{sec:supp:fisheye-task}.

\noindent\textbf{Labeling Policy}
The labeling policy matches with Section~\ref{supp:tasks:nuimages}, except for the missing labels for trams.

\noindent\textbf{D-RICO Classes.}
\begin{itemize} \setlength{\itemindent}{0.5cm}
    \item \textbf{person:} \textit{pedestrian}, \textit{rider}
    \item \textbf{bicycle} \textit{bicycle}
    \item \textbf{vehicle:} \textit{car}, \textit{bus}, \textit{truck}, \textit{motorcycle}, \textit{train}
\end{itemize}

\noindent\textbf{EC-RICO Classes.}

\begin{itemize} \setlength{\itemindent}{0.5cm}
    \item \textbf{person:} \textit{Pedestrian}
    \item \textbf{car:} \textit{PassengerCar}
    \item \textbf{bicycle:} \textit{bicycle}
    \item \textbf{motorcycle:} \textit{motorcycle}
\end{itemize}

\section{Additional Details on the Experimental Setup}
In this section, we provide additional details on the experimental setup, including the hyperparameters and the adaptations made during the implementation of the IL methods.

\subsection{Hyperparameters}
We mostly used the hyperparameters as specified in~\cite{fang_eva-02_2023}.
Table~\ref{tab:supp:hyper-general} lists the general hyperparameters we used for all experiments that differ from~\cite{fang_eva-02_2023}.
In Table~\ref{tab:supp:hyper-joint-D} to~\ref{tab:supp:hyper-ldb-EC}, the experiment-specific hyperparameters are listed.
For all trainings, we conducted a small random hyperparameters search on the learning rate and the method-specific hyperparameters.

\begin{table}
    \centering
    \caption{General hyperparameters used in all experiments if not specified differently.}
    \label{tab:supp:hyper-general}
    \begin{tabular}{ll}
    \toprule
        \textbf{Hyperparameter} & \textbf{Value} \\
    \midrule
       batch size  & $20$ \\
       input image size  & $1536 \times1536\times 3$\\
       training iterations  & $696$\\
       evaluation period  & $500$\\
       warm-up length  &  $0.1$\\
       warm-up factor & $0.001$ \\
       learning rate & $0.001$ \\
       learning rate scheduler & cosine\\
       learning rate scheduler end value & $0$\\
       Aug.: rnd. flip & True\\
       Aug.: resize scale & True\\
       Aug.: fixed size crop & True\\
       Aug.: rnd. brightness & True\\
       Aug.: rnd. contrast & True\\
       Aug.: rnd. saturation & True\\
       Aug.: rnd. lightning & True\\
       Aug.: resize scale, min scale & $0.1$ \\
       Aug.: resize scale, max scale & $2$ \\
       Aug.: rnd. brightness, min. intensity & $0.6$ \\
       Aug.: rnd. brightness, max. intensity & $1.4$ \\
       Aug.: rnd. contrast, min. intensity & $0.6$ \\
       Aug.: rnd. contrast, max. intensity & $1.4$ \\
       Aug.: rnd. saturation, min. intensity & $0.6$ \\
       Aug.: rnd. saturation, max. intensity & $1.4$ \\
       Aug.: rnd. lightning, scale & $0.1$ \\
       Aug. for evaluation and testing & False \\
       
     \bottomrule
    \end{tabular}
\vspace{1mm} 
\caption*{\scriptsize Abbreviations: Aug. = augmentation, min. = minimum, max. maximum, rnd. = random}
\end{table}

\begin{table}
    \centering
    \caption{Hyperparameters for joint training on D-RICO.}
    \label{tab:supp:hyper-joint-D}
    \begin{tabular}{ll}
    \toprule
        \textbf{Hyperparameter} & \textbf{Value} \\
    \midrule
       training iterations & $10,440$\\
       evaluation period & $696$\\
     \bottomrule
    \end{tabular}
\end{table}

\begin{table}
    \centering
    \caption{Hyperparameters for joint training on EC-RICO.}
    \label{tab:supp:hyper-joint-EC}
    \begin{tabular}{ll}
    \toprule
        \textbf{Hyperparameter} & \textbf{Value} \\
    \midrule
       training iterations & $5,568$\\
       evaluation period & $696$\\
     \bottomrule
    \end{tabular}
\end{table}

\begin{table}
    \centering
    \caption{Hyperparameters for ABR~\cite{liu_augmented_2023} training on D-RICO.}
    \label{tab:supp:hyper-abr-D}
    \begin{tabular}{ll}
    \toprule
        \textbf{Hyperparameter} & \textbf{Value} \\
    \midrule
       learning rate & $0.001$\\
       $\alpha$ for class distillation & $1$\\
       $\alpha$ for box distillation & $1$\\
       $\beta$ for box distillation & $0$\\
       $\gamma$ for box distillation & $0$\\
       memory buffer size & $2,000$\\
       training iterations & $870$ \\
     \bottomrule
    \end{tabular}
\end{table}

\begin{table}
    \centering
    \caption{Hyperparameters for ABR~\cite{liu_augmented_2023} training on EC-RICO.}
    \label{tab:supp:hyper-abr-EC}
    \begin{tabular}{ll}
    \toprule
        \textbf{Hyperparameter} & \textbf{Value} \\
    \midrule
       learning rate & $0.001$\\
       $\alpha$ for class distillation & $1$\\
       $\alpha$ for box distillation & $1$\\
       $\beta$ for box distillation & $0$\\
       $\gamma$ for box distillation & $0$\\
       memory buffer size & $2,000$\\
       training iterations & $870$ \\
     \bottomrule
    \end{tabular}
\end{table}

\begin{table}
    \centering
    \caption{Hyperparameters for Meta-IOD~\cite{joseph_incremental_2022} training on D-RICO.}
    \label{tab:supp:hyper-iod-D}
    \begin{tabular}{ll}
    \toprule
        \textbf{Hyperparameter} & \textbf{Value} \\
    \midrule
       learning rate & $0.0001$\\
       Number of features per class & $5000$\\
       Number of images per class & $5000$\\
       Distillation weight & $1$ \\
       Warp Iteration & $20$\\
       Warp Layer & box\_head.2.conv4\\
     \bottomrule
    \end{tabular}
\end{table}

\begin{table}
    \centering
    \caption{Hyperparameters for Meta-IOD~\cite{joseph_incremental_2022} training on EC-RICO.}
    \label{tab:supp:hyper-iod-EC}
    \begin{tabular}{ll}
    \toprule
        \textbf{Hyperparameter} & \textbf{Value} \\
    \midrule
       learning rate  & $0.0001$\\
       Number of features per class & $5000$\\
       Number of images per class & $5000$\\
       Distillation weight & $1$ \\
       Warp Iteration & $20$\\
       Warp Layer & box\_head.2.conv4\\
     \bottomrule
    \end{tabular}
\end{table}

\begin{table}
    \centering
    \caption{Hyperparameters for BPF~\cite{leonardis_bridge_2025} training on D-RICO.}
    \label{tab:supp:hyper-bpf-D}
    \begin{tabular}{ll}
    \toprule
        \textbf{Hyperparameter} & \textbf{Value} \\
    \midrule
       learning rate  & $0.001$\\
       $\alpha$ for class distillation & $0.1$\\
       $\alpha$ for box distillation & $0.15$\\
       memory buffer size & $2000$\\
     \bottomrule
    \end{tabular}
\end{table}

\begin{table}
    \centering
    \caption{Hyperparameters for BPF~\cite{leonardis_bridge_2025} training on EC-RICO.}
    \label{tab:supp:hyper-bpf-EC}
    \begin{tabular}{ll}
    \toprule
        \textbf{Hyperparameter} & \textbf{Value} \\
    \midrule
       learning rate  & $0.001$\\
       $\alpha$ for class distillation & $0.15$\\
       $\alpha$ for box distillation & $0.15$\\
       memory buffer size & $2000$\\
     \bottomrule
    \end{tabular}
\end{table}

\begin{table}
    \centering
    \caption{Hyperparameters for LDB~\cite{song_non-exemplar_2024} training on D-RICO.}
    \label{tab:supp:hyper-ldb-D}
    \begin{tabular}{ll}
    \toprule
        \textbf{Hyperparameter} & \textbf{Value} \\
    \midrule
       learning rate  & $0.001$\\
       batch size & 5\\
       training iterations & 2784\\
     \bottomrule
    \end{tabular}
\end{table}

\begin{table}
    \centering
    \caption{Hyperparameters for LDB~\cite{song_non-exemplar_2024} training on EC-RICO.}
    \label{tab:supp:hyper-ldb-EC}
    \begin{tabular}{ll}
    \toprule
        \textbf{Hyperparameter} & \textbf{Value} \\
    \midrule
       learning rate  & $0.001$\\
       batch size & 5\\
       training iterations & 2784\\
     \bottomrule
    \end{tabular}
\end{table}

\subsection{Implementation Details of IL Methods}

In this section we elaborate on the implementation details of the Replay approach and ABR\cite{liu_augmented_2023}, Meta-ILOD~\cite{joseph_incremental_2022}, BPF~\cite{leonardis_bridge_2025} and LDB~\cite{song_non-exemplar_2024}.

\subsubsection{Replay}
We employ a growing memory buffer that is initialized empty. For each new task, a subset of training data is randomly selected and added to the buffer. During training, the remaining task data is merged with the memory buffer.

A growing buffer is preferable to a static one, especially for long task sequences. A fixed-size buffer either over-replays early task examples, wasting compute, or retains too few examples as tasks accumulate. If memory constraints are not strict, an expanding buffer ensures stability by maintaining consistent replay examples per task.

To keep the number of training iterations on new task data constant, the total iterations must increase over time. Consequently, both memory and compute requirements grow with the number of tasks. This approach is feasible only if resource constraints are not too restrictive.

\subsubsection{ABR}

In the original ABR implementation, the memory buffer accumulates boxes from previous tasks but is only divided by class. We introduce an additional split by task, motivated by the feature dissimilarity across objects, which prevents defining a uniform proximity metric to the class mean.

Our approach retains a fixed-size memory buffer, as in the original method, but after each task, the half containing the least representative boxes for each class and task is discarded. We compute each object's feature vector without augmentations to ensure representative features. Stability is improved by selecting only non-overlapping boxes and excluding those smaller than $20 \times 20$ pixels~\cite{wang_wanderlust_2021}.

For mixup and mosaic, we closely follow the ABR implementation. Since we apply resize augmentation, we also scale replayed boxes in the mixup to match the resized image.

Using the Cascade Faster-RCNN~\cite{cai_cascade_2017} detection head prevents direct application of the same distillation approach. Instead, we generate soft labels from the last (\ie third) cascade stage without filtering empty boxes. Given our large input image size of $1536 \times 1536$, we stabilize the bounding box distillation loss by computing it on the logarithm of the box coordinates.

Since not all objects are labeled for every class in each task, we apply a mask during distillation to ensure it only affects classes seen by the teacher during training.

\subsubsection{Meta-ILOD}

We follow the original implementation for image and feature storage. The \textit{warping} operation is applied to the last convolution layer of the final (\ie third) cascade stage.

Since the backbone is fixed and not trained, we omit the backbone feature distillation loss. The \textit{warp} loss is computed using the last cascade stage.

In all other aspects, we adhere closely to the original implementation.

\subsubsection{BPF}
BPF involves generating pseudo labels using the previous model. However, since we do not operate in a CIL setting, we disable this feature.

For distillation, we obtain soft predictions from the final cascade stage, similar to ABR. Additionally, we apply masking in the distillation loss to ensure learning only from predictions the teacher model was trained on. As the \textit{finetuning} teacher, we use the model trained for the individual training baseline.

\subsubsection{LDB}
Since we apply augmentations during training but not during inference, we cannot directly collect image features for the nearest mean classifier during training. Instead, we compute the image features and class means separately, storing the means for use during IL.

Our IL framework enables task iteration without requiring separate training for each task. To improve efficiency, we store domain bias terms and output layers as matrices, selecting the corresponding row based on the determined task ID. This setup allows a single model to be used for inference, automatically selecting the appropriate model components based on the task ID.

\subsection{Mean Average Precision}
We report the COCO‑style mean Average Precision (mAP), computed by averaging the AP at ten intersection over union (IoU) thresholds from 0.50 to 0.95 in increments of 0.05, denoted as $\text{mAP}@[0.50:0.05:0.95]$.

\subsection{Task Affinity}
For the task affinity metric, only the output layers of the backbone are adapted, meaning the last layer in both the classification and bounding box regression parts.

\section{Additional Results}
The main paper holds the primary results of this work. 
We included here some additional results. 

\subsection{Quantitative Comparison to Existing Benchmarks}~\label{subsec:suppl:quanti-comparison-becnhmarks}
In Table~\ref{tab:comp-different-benchmarks}, we qualitatively compare the proposed benchmarks to existing ones. To highlight these differences quantitatively, we conduct a comparative analysis with the CLAD-D~\cite{verwimp_clad_2023} and VOC Series~\cite{song_non-exemplar_2024} benchmarks (see Section~\ref{sec:related-work} for brief descriptions). Specifically, we evaluate joint training, individual training, naive FT, and replay with 10\% of the training data. The results for CLAD-D and VOC Series are presented in Table~\ref{tab:clad-d-voc-series}, while the results for D-RICO and EC-RICO are shown in Table~\ref{tabel:main-results}.

We observe the most significant difference in naive FT forgetting rates: 19\% for D-RICO compared to just 4.33\% and 4.21\% for CLAD-D and VOC Series, respectively. A similar pattern emerges in the 1\% replay scenario, where forgetting is minimal for CLAD-D and VOC Series but remains relatively high (8.91\%) for D-RICO. Additionally, the $\overline{\bm{\mAP}}$ is lower for D-RICO, indicating that a larger, more comprehensive model is necessary to achieve strong overall performance.

Beyond incremental learning experiments, we also evaluate task affinity (see Section~\ref{subsec:bench:task-affinity}) across these benchmarks. We report mean TA scores of 94\% for CLAD-D, 98\% for VOC Series, and notably lower at 76\% for D-RICO. This indicates that the output layer alone is insufficient for adapting to new tasks in D-RICO, underscoring greater task diversity and dissimilarity.

Overall, these experiments clearly demonstrate that the diversity present in D-RICO and consequently EC-RICO significantly exceeds that of the existing benchmarks, CLAD-D and VOC Series.

\begin{table}[htbp]
\centering
\setlength{\tabcolsep}{3pt}
\scriptsize
\caption{Results on the CLAD-D and VOC Series benchmarks.}
\label{tab:clad-d-voc-series}
\begin{tabular}{lccccc}
\toprule
                \textbf{Benchmark}          & \textbf{Method}            & \textbf{$\overline{\bm{\mAP}}$ \textuparrow} & \textbf{$\mathbf{FM}$  \textdownarrow} & \textbf{$\mathbf{FWT}$ \textuparrow} & \textbf{$\mathbf{IM}$ \textuparrow} \\ \midrule

            & Joint Training         & $59.96$ &  &  &   \\
CLAD-D   & Individual Training    & $55.24$ &  &  &  \\
   & Naive FT               & $49.88$ & $4.33$ & $-1.05$ & $-2.24$  \\
  & Replay 1\%            & $50.36$ & $3.79$ & $-0.36$ & $-2.17$  \\
           
            \midrule
            
            & Joint Training         & $62.54$ &  &  &   \\
VOC Series   & Individual Training    & $64.49$ &  &  &  \\
   & Naive FT               & $59.86$ & $4.21$ & $-1.06$ & $0.46$  \\
  & Replay 1\%            & $61.45$ & $2.24$ & $-0.49$ & $0.59$  \\
            \bottomrule
\end{tabular}%
\end{table}

\subsection{Unfrozen Backbone}

In addition to the experiments with a frozen backbone, we analyze the impact of making the backbone trainable. For simplicity, we use a subset of D-RICO (tasks: $1\veryshortarrow 2 \veryshortarrow 3 \veryshortarrow 11 \veryshortarrow 15$) and compare joint and individual training, naive fine-tuning (FT), and 1\% replay in this reduced setting, both with and without a frozen backbone.

Table~\ref{tab:frozen-backbone} presents the results, showing that unfreezing the backbone improves model plasticity and overall performance. However, in the naive FT scenario, this comes at the cost of increased forgetting. Since 1\% replay is generally effective at mitigating forgetting, it also helps reduce forgetting when the backbone is unfrozen. These findings suggest that while an unfrozen backbone can offer performance benefits, it also amplifies forgetting, making the choice between frozen and unfrozen non-trivial. In such cases, more robust IL methods become necessary to counteract the negative effects of increased plasticity.

\begin{table}[htbp]
\centering
\setlength{\tabcolsep}{3pt}
\scriptsize
\caption{Un/frozen backbones (tasks: $1\veryshortarrow  2\veryshortarrow  3\veryshortarrow  11\veryshortarrow  15$).}
\label{tab:frozen-backbone}
\begin{tabular}{lccccc}
\toprule
                \textbf{Benchmark}          & \textbf{Method}            & \textbf{$\overline{\bm{\mAP}}$ \textuparrow} & \textbf{$\mathbf{FM}$  \textdownarrow} & \textbf{$\mathbf{FWT}$ \textuparrow} & \textbf{$\mathbf{IM}$ \textuparrow} \\ \midrule
        & Joint Training    & $37.30$ &  &  &  \\
frozen   & Individual Training    & $42.15$ &  &  &  \\
backbone   & Naive FT               & $26.77$ & $18.99$ & $-2.76$ & $6.94$   \\
  & Replay 1\%            & $38.53$ & $3.33$ & $-0.16$ & $6.18$   \\
           
            \midrule
            & Joint Training    & $38.33$ &  &  &  \\
unfrozen   & Individual Training    & $44.80$ &  &  &  \\
backbone   & Naive FT               & $29.03$ & $21.57$ & $2.94$ & $9.38$   \\
  & Replay 1\%            & $42.03$ & $2.58$ & $1.21$ & $6.57$   \\
            \bottomrule
\end{tabular}%
\end{table}

\vspace{0.3mm}

\subsection{Detailed Run Results}
\begin{figure}[!htb]
    \centering
    \includegraphics[width=0.92\linewidth]{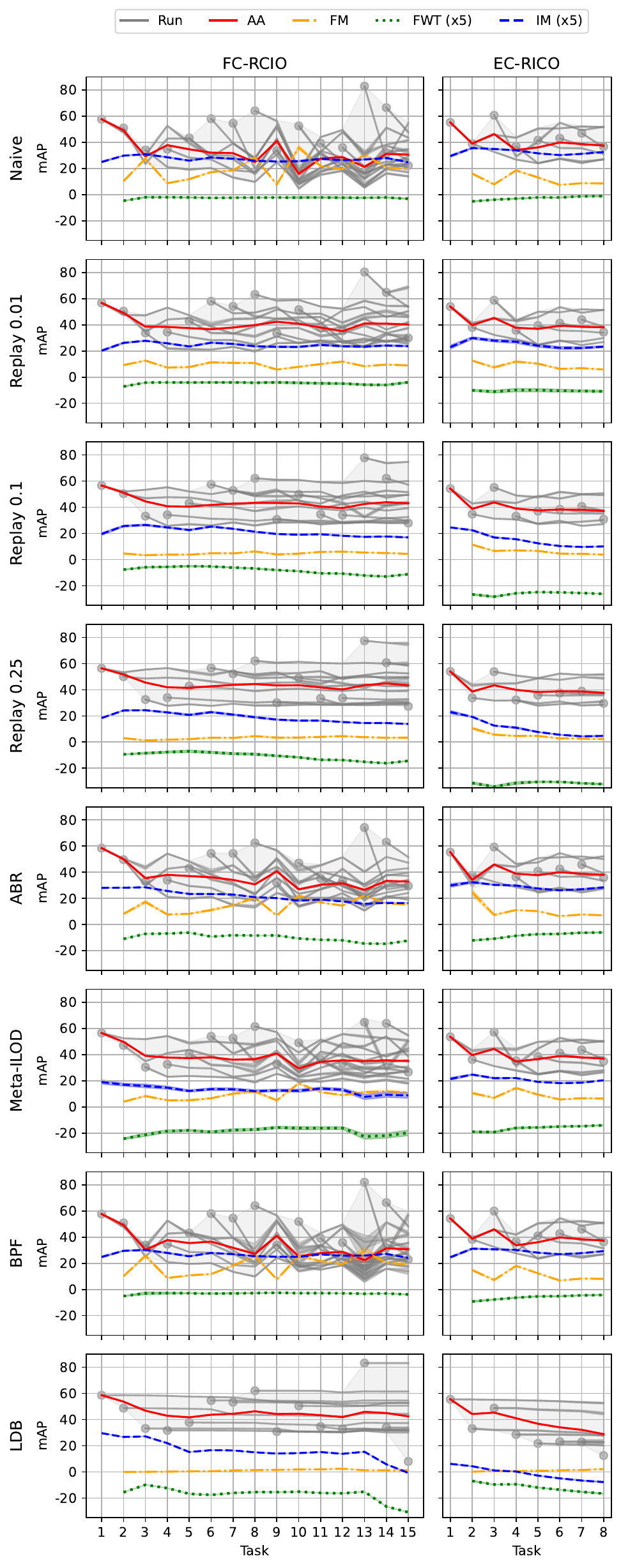}
    \caption{Test performance on D-RICO and EC-RICO for the tested methods. The value of  $\mathrm{FWT}$ and  $\mathrm{IM}$ (except for LDB for EC-RICO) are scaled by a factor five to enhance visibility.}
    \label{fig:results:runs}
\end{figure}

Table~3 presents the $\mathrm{AA}$, $\mathrm{FM}$, $\mathrm{FWT}$, and $\mathrm{IM}$ metrics, summarizing each method's performance. To provide a more detailed view, Figure~\ref{fig:results:runs} illustrates the progression of each task after its initial learning, along with the evolution of these metrics.

Performance generally declines after learning a new task. With increasing replay size, IL metrics become more parallel and stable. The ABR and BPF methods closely resemble naïve FT, highlighting their underperformance in IL settings. In contrast, Meta-ILOD and LDB exhibit greater stability.

Qualitatively, D-RICO and EC-RICO behave similarly, with weaker methods showing more unstable learning curves. Both benchmarks exhibit a decreasing  $\mathrm{AA}$ trend. The IM drops across all runs, while  $\mathrm{FWT}$ increases, indicating that at the start of training, all approaches outperform joint training but still lag behind individual training. Interestingly, $\mathrm{FWT}$ continues to rise, suggesting increased model plasticity, though this effect is minor and stabilizes after one or two tasks.

As expected,  $\mathrm{FM}$ increases for D-RICO, reflecting the growing challenge of knowledge retention with additional tasks. However, unexpectedly, $\mathrm{FM}$ decreases in EC-RICO. We hypothesize that this results from the increasing number of labels, reducing the relative contribution of early tasks.

Analyzing individual runs reveals that some tasks follow similar trends, suggesting they rely on overlapping features.

\subsection{Joint and Individual Training Results}\label{subsubsec:supp:discrepancy}
Joint and individual training represent two kinds of upper bounds, with joint training (D-RICO: 43.75, EC-RICO: 38.46) serving as a practical upper bound for single model configurations.
The individual training (D-RICO: 49.37, EC-RICO: 45.54) demonstrates what would be possible without the restriction of a single model.
To analyze this discrepancy further, we compare individual and joint training using the relative difference to individual training.

Table~\ref{tab:contradiction} quantifies the performance gap between joint and individual training. Smaller objects have the highest differences, with mAP$_{\text{small}}$ dropping most (D: 0.25, EC: 0.23), highlighting the challenge of learning fine-grained details in a joint setting.

For D-RICO, bicycles show the highest discrepancy (0.39) due to inconsistent labeling, often misclassified as motorcycles or treating bicycle racks differently. In EC-RICO, motorcycles (0.27) and bicycles (0.19) also exhibit high gaps due to annotation policies, where motorcycles are sometimes only labeled when ridden. Traffic lights (0.24) and street signs (0.35) show notable differences, driven by noisy annotations and small bounding boxes.

As street signs appear in only one task, this suggests that intra-class contradictions alone do not fully explain the discrepancy, making the benchmark inherently challenging for both IL methods and joint training. 
The reason could be that the street sign class is introduced during nighttime tasks, making it not always clearly visible and somewhat ambiguous.

Our further analysis in this regard yields the following concrete examples, which demonstrate contradictions and false negatives during both training and evaluation:

\begin{enumerate}
    \item \textit{Thermal+RGB:} The object labeled as a \textit{bicycle} in the ground truth is actually a motorbike, which is instead predicted as a \textit{vehicle} by the model.
    
    \item \textit{Drone:} In this case, the ground truth labels for bicycles are annotated as \textit{vehicle}, whereas the prediction correctly identifies them as \textit{bicycle}. Additionally, the amodal ground truth bounding boxes do not match with the predictions.
    
    \item \textit{Video game:} The ground truth annotations exclude the rider from the motorbike bounding boxes, in contrast to the predictions, which include the rider as part of the object.
    
    \item \textit{Fisheye car} and \textit{Fisheye Fixed:} In these settings, small and distant objects present in the ground truth are frequently missed by the model's predictions.
\end{enumerate}

\begin{table}[htbp]
\centering
\footnotesize
\caption{Results for D-RICO and EC-RICO.}
\label{tab:contradiction}
\begin{tabular}{llll}
\toprule
\multicolumn{2}{c}{\textbf{D-RICO}} & \multicolumn{2}{c}{\textbf{EC-RICO}} \\
\cmidrule(r){1-2} \cmidrule(l){3-4}
\textbf{Metric} & \textbf{Rel. Diff.} & \textbf{Metric} & \textbf{Rel. Diff.} \\
\midrule
mAP            & $0.13 {\scriptstyle\pm 0.05}$ & mAP            & $0.16 {\scriptstyle\pm 0.05}$ \\
mAP$_{\text{large}}$  & $0.08 {\scriptstyle\pm 0.06}$ & mAP$_{\text{large}}$  & $0.12 {\scriptstyle\pm 0.06}$ \\
mAP$_{\text{medium}}$ & $0.13 {\scriptstyle\pm 0.05}$ & mAP$_{\text{medium}}$ & $0.16 {\scriptstyle\pm 0.06}$ \\
mAP$_{\text{small}}$  & $0.25 {\scriptstyle\pm 0.19}$ & mAP$_{\text{small}}$  & $0.23 {\scriptstyle\pm 0.09}$ \\
\midrule
mAP$_{\text{person}}$       & $0.14 {\scriptstyle\pm 0.06}$ & mAP$_{\text{person}}$       & $0.16 {\scriptstyle\pm 0.06}$ \\
mAP$_{\text{bicycle}}$      & $0.39 {\scriptstyle\pm 0.48}$ & mAP$_{\text{car}}$          & $0.11 {\scriptstyle\pm 0.07}$ \\
mAP$_{\text{vehicle}}$      & $0.11 {\scriptstyle\pm 0.06}$ & mAP$_{\text{bicycle}}$      & $0.19 {\scriptstyle\pm 0.10}$ \\
                     &                               & mAP$_{\text{motorcycle}}$   & $0.27 {\scriptstyle\pm 0.22}$ \\
                     &                               & mAP$_{\text{truck}}$        & $0.17 {\scriptstyle\pm 0.09}$ \\
                     &                               & mAP$_{\text{bus}}$          & $0.14 {\scriptstyle\pm 0.02}$ \\
                     &                               & mAP$_{\text{traffic\ light}}$ & $0.24 {\scriptstyle\pm 0.02}$ \\
                     &                               & mAP$_{\text{street\ sign}}$  & $0.35 {\scriptstyle\pm 0.01}$ \\
\bottomrule
\end{tabular}
\end{table}

\begin{figure}[t]
    \centering
    \includegraphics[width=0.9\linewidth]{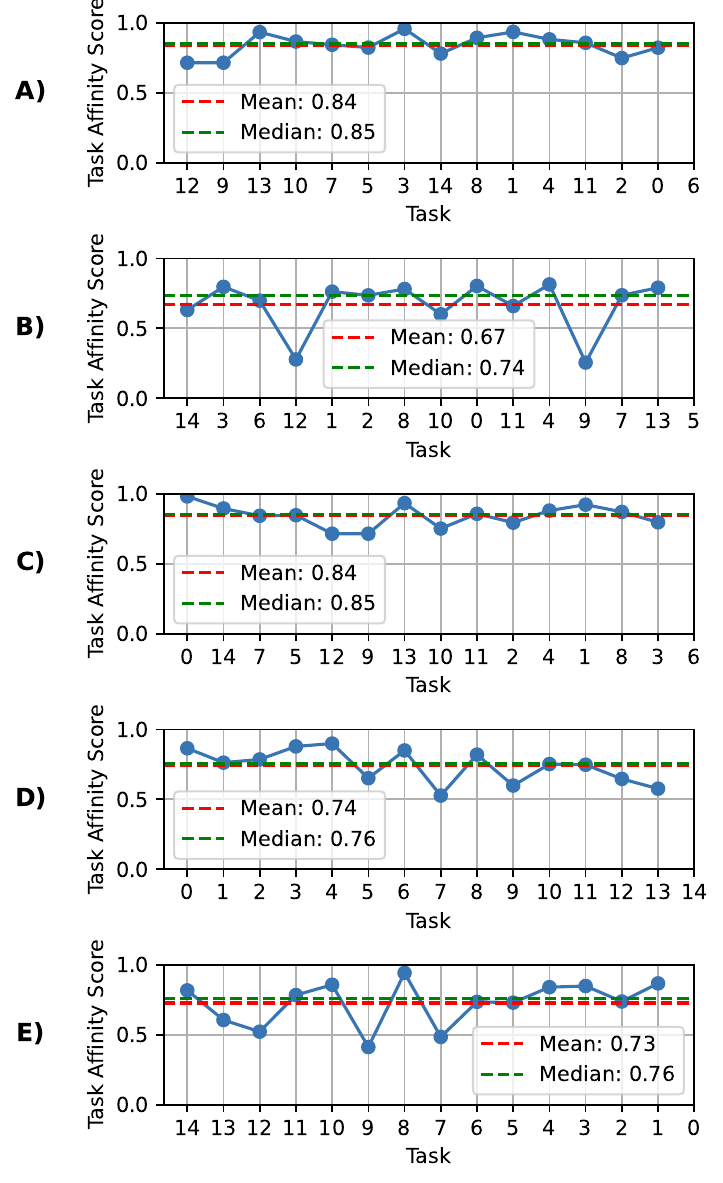}
    \caption{Task affinity to the next task for different task orders.}
    \label{fig:ta_by_task_order}
\end{figure}

\subsection{Task Order}
\label{supp:task_order}

In Section~4.7, we analyze the effect of different task orders on performance. The following task orders were evaluated:

\begin{enumerate}[label=\textbf{\Alph*)}]
    \item $1 \veryshortarrow 2 \veryshortarrow 3 \veryshortarrow 4 \veryshortarrow 5 \veryshortarrow 6 \veryshortarrow 7 \veryshortarrow 8 \veryshortarrow 9 \veryshortarrow 10 \veryshortarrow 11 \veryshortarrow 12 \veryshortarrow 13 \veryshortarrow 14 \veryshortarrow 15$
    \item $15 \veryshortarrow 14 \veryshortarrow 13 \veryshortarrow 12 \veryshortarrow 11 \veryshortarrow 10 \veryshortarrow 9 \veryshortarrow 8 \veryshortarrow 7 \veryshortarrow 6 \veryshortarrow 5 \veryshortarrow 4 \veryshortarrow 3 \veryshortarrow 2 \veryshortarrow 1$
    \item $1 \veryshortarrow 15 \veryshortarrow 8 \veryshortarrow 6 \veryshortarrow 13 \veryshortarrow 10 \veryshortarrow 14 \veryshortarrow 11 \veryshortarrow 12 \veryshortarrow 3 \veryshortarrow 5 \veryshortarrow 2 \veryshortarrow 9 \veryshortarrow 4 \veryshortarrow 7$
    \item $13 \veryshortarrow 10 \veryshortarrow 14 \veryshortarrow 11 \veryshortarrow 8 \veryshortarrow 6 \veryshortarrow 4 \veryshortarrow 15 \veryshortarrow 9 \veryshortarrow 2 \veryshortarrow 5 \veryshortarrow 12 \veryshortarrow 3 \veryshortarrow 1 \veryshortarrow 7$
    \item $15 \veryshortarrow 4 \veryshortarrow 7 \veryshortarrow 13 \veryshortarrow 2 \veryshortarrow 3 \veryshortarrow 9 \veryshortarrow 11 \veryshortarrow 1 \veryshortarrow 12 \veryshortarrow 5 \veryshortarrow 10 \veryshortarrow 8 \veryshortarrow 14 \veryshortarrow 6$
\end{enumerate}

To evaluate robustness to task ordering, we extend this with five additional randomly sampled orders (ten total). These experiments confirm that performance is largely insensitive to task order for replay-based methods, with average performance $\overline{\bm{\mAP}} = 42.57 {\scriptstyle\pm 0.46}$ and forgetting measure $\mathrm{FM}= 4.41 {\scriptstyle\pm 0.65}$. In contrast, naïve fine-tuning (FT) exhibits moderate sensitivity, with $\overline{\bm{\mAP}} = 32.33 {\scriptstyle\pm 3.39}$ and $\mathrm{FM}= 17.85 {\scriptstyle\pm 3.75}$.

The TA score, as defined in Section~3.5, is visualized for each order in Figure~\ref{fig:ta_by_task_order}. Interestingly, task order \textbf{B)} results in the best outcome for naïve FT, despite having the lowest average TA and exhibiting several sharp drops in affinity. This reinforces our finding that TA does not correlate with IL performance for either FT or replay. However TA remains a valuable unified metric for quantifying task (dis)similarity—even if it does not predict continual learning performance directly or capture all influencing factors.

\subsection{Class imbalance}
We evaluated \textit{federated loss} (EVA-02 implementation) for EC-RICO, but see no gains (Tab. \ref{tab:fed-loss}). Improving balancing further could boost future methods, but that lies beyond this benchmark’s scope.

\begin{table}[htbp]
\centering
\setlength{\tabcolsep}{3pt}
\scriptsize
\caption{Results on EC-RICO with federated loss.}
\label{tab:fed-loss}
\begin{tabular}{ccccc}
\toprule
                \textbf{Method}                     & \textbf{$\overline{\bm{\mAP}}$ \textuparrow} & \textbf{$\mathbf{FM}$  \textdownarrow} & \textbf{$\mathbf{FWT}$ \textuparrow} & \textbf{$\mathbf{IM}$ \textuparrow} \\ \midrule

 joint training              & $37.48$ &  &  &   \\
  individual training            & $45.40$ &  &  &   \\
  naive finetuning            & $37.86$ & $7.51$ & $-0.67$ & $6.96$  \\
  replay 10\%            & $37.69$ & $3.39$ & $-4.90$ & $3.18$  \\
            \bottomrule
\end{tabular}
\end{table}

\section{Detectron2 for IL}\label{supp:sec:il-detectron2}
Incremental Learning (IL) for classification typically relies on the Avalanche library~\cite{lomonaco_avalanche_2021}.
However, the model architecture and evaluation protocols for object detection differ substantially from classification, necessitating extensive modifications to the Avalanche framework.
Most prior research on IL for object detection utilizes established object detection frameworks such as Detectron2~\cite{wu_detectron2_2019}, restarting the entire codebase for each task while preserving and transferring essential information between tasks~\cite{liu_augmented_2023, leonardis_bridge_2025, song_non-exemplar_2024, joseph_incremental_2022}.

While this approach is functional and existing templates can facilitate future work, we opted to refactor Detectron2 to better accommodate IL requirements.
The primary enhancement is the implementation of task-iterative processing, allowing seamless storage of data, models, and images as variables between consecutive tasks.
This refactoring required the following modifications:

\begin{itemize}
    \item \textbf{Data Sampler:} We developed a specialized data sampler that processes the complete training, validation, and testing annotation files while allowing selective loading of task-specific data.
    \item \textbf{Evaluator:} IL frameworks necessitate evaluation across both current and previous tasks. We modified the evaluator to systematically iterate through and assess all tasks.
    \item \textbf{Event Storage:} When training and evaluating tasks within iterative loops, metrics must be stored in task-specific contexts. Our custom event storage mechanism preserves metrics separately for each task.
    \item \textbf{IL Metrics:} After each task, all test sets are evaluated and key IL metrics are calculated, including \textit{Average Accuracy}, \textit{Average Incremental Accuracy}, \textit{Forgetting Measure}, and \textit{Backwards Transfer}. Plasticity metrics require subsequent computation using results from both joint and individual training paradigms.
    \item \textbf{Joint and Individual Training:} Both joint and individual training runs must be conducted to calculate plasticity metrics. These results can be generated using a designated flag in our framework.
    \item \textbf{Replay:} Using a configurable percentage parameter, historical data can be incorporated into a replay buffer and utilized in subsequent tasks.
\end{itemize}

Beyond these core modifications, numerous additional changes were implemented regarding hooks, model persistence mechanisms, task ordering protocols, and other system components.

The advantages of our customized Detectron2 version include:
\begin{itemize}
    \item Accelerated integration of novel methodologies
    \item More coherent experimental configuration
    \item Enhanced training efficiency
\end{itemize}

Detailed installation instructions for \texttt{Detectron2-IL} are provided in the corresponding GitHub Repository.

\end{document}